\pdfoutput=1
\PassOptionsToPackage{sort}{natbib}
\documentclass[preprint,11pt,authoryear]{elsarticle}

\usepackage[a4paper,top=25mm,bottom=25mm,left=25mm,right=25mm,footskip=10mm]{geometry}
\usepackage[T1]{fontenc}
\usepackage[nohelv]{newtxtext}
\usepackage[scaled=0.92]{helvet}
\usepackage[varvw]{newtxmath}
\DeclareMathAlphabet{\mathsf}{T1}{phv}{m}{n}
\SetMathAlphabet{\mathsf}{bold}{T1}{phv}{b}{n}
\usepackage{amsmath}
\usepackage{mathtools}

\usepackage{amsthm}
\usepackage{bm}
\usepackage{dsfont}
\usepackage{graphicx}
\graphicspath{{figures/}{./}}
\usepackage[dvipsnames,svgnames,table]{xcolor}
\usepackage{booktabs}
\usepackage{array}
\usepackage{tabularx}
\usepackage{multirow}
\usepackage[shortlabels]{enumitem}
\setlist[itemize,1]{label=$\bullet$}
\usepackage{algorithm}
\usepackage[noend]{algpseudocode}
\usepackage{setspace}
\usepackage{placeins}
\usepackage{xurl}
\usepackage{xr-hyper}
\usepackage[hidelinks]{hyperref}
\hypersetup{colorlinks=true,linkcolor=NavyBlue,citecolor=NavyBlue,urlcolor=NavyBlue}
\usepackage{orcidlink}

\makeatletter
\renewcommand{\emailauthor}[2]{%
  \stepcounter{ead}%
  \g@addto@macro\@elseads{%
    \raggedright
    \let\corref\@gobble
    \def\@@tmp{#1}%
    \eadsep{\ttfamily\expandafter\strip@prefix\meaning\@@tmp}%
    \def\eadsep{\unskip,\space}%
  }%
}
\makeatother

\theoremstyle{plain}
\newtheorem{theorem}{Theorem}
\newtheorem{proposition}{Proposition}
\newtheorem{lemma}{Lemma}
\newtheorem{corollary}{Corollary}
\theoremstyle{definition}
\newtheorem{definition}{Definition}
\theoremstyle{remark}

\algrenewcommand\algorithmicrequire{\textbf{Input:}}
\algrenewcommand\algorithmicensure{\textbf{Output:}}
\algrenewcommand\algorithmiccomment[1]{\hfill{\footnotesize\itshape #1}}

\newcommand{\1}{\mathds{1}}
\allowdisplaybreaks[2]

\makeatletter
\def\ps@pprintTitle{%
  \let\@oddhead\@empty
  \let\@evenhead\@empty
  \def\@oddfoot{\reset@font\footnotesize\itshape Preprint submitted to \@journal\hfill}%
  \let\@evenfoot\@oddfoot}
\long\def\@makecaption#1#2{%
  \vskip\abovecaptionskip\footnotesize
  \sbox\@tempboxa{#1. #2}%
  \ifdim \wd\@tempboxa >\hsize
    #1. #2\par
  \else
    \global \@minipagefalse
    \hb@xt@\hsize{\hfil\box\@tempboxa\hfil}%
  \fi
  \vskip\belowcaptionskip}
\keywordtitlesep{.\ }
\providecommand{\doi}[1]{}
\renewcommand{\doi}[1]{\href{https://doi.org/#1}{https://doi.org/\nolinkurl{#1}}}

\makeatother

\makeatletter
\def\ps@pprintTitle{%
  \let\@oddhead\@empty
  \let\@evenhead\@empty
  \def\@oddfoot{\reset@font\footnotesize\itshape Preprint\hfill}%
  \let\@evenfoot\@oddfoot}
\makeatother
\hypersetup{pdfauthor={Ali Nikseresht}}
\usepackage[final]{microtype}
\begin{document}
\begin{frontmatter}

\title{Tourist arrival forecasting with game-theoretic social influence weighting and deep learning}

\author{Ali Nikseresht\,\orcidlink{0000-0002-6107-7699}}

\ead{ali.nikseresht@gatech.edu}

\affiliation{
    organization={School of Computer Science, Georgia Institute of Technology},
    city={Atlanta},
    postcode={GA 30332},
    country={United States}
}
\begin{abstract}
In the era of Industry 5.0, data-driven decision-making has become indispensable for planning in service systems such as tourism. This paper proposes a non-linear hybrid approach for forecasting tourist arrivals to Hong Kong from five major source markets before COVID-19 and to Sanya, China, through the COVID-19 shock. The method integrates multiple sources of Internet big data and employs a cooperative game-theoretic algorithm to identify coalitions of opinion leaders on social media, whose content then receives larger weights. The Empirical Wavelet Transform (EWT), applied causally, manages nonstationary attributes, and a Stacked Bi-directional Long Short-Term Memory (Stacked BiLSTM) network tuned by Bayesian optimization generates the forecasts. We prove that the Shapley values of the underlying trust-weighted coverage game have a closed form and that the coalition of maximal synergy can be found exactly in polynomial time. Relative to the same network fed only with past arrivals, Internet big data reduce the one-step-ahead root mean squared error by 4.1--41.7\% across the Hong Kong markets and opinion-leader weighting by a further 10.1--49.4\%, against 5.3\% and 8.7\% for Sanya. The approach compares favorably with published benchmarks and remains robust under volatile conditions, which suggests uses in logistics, supply chains and production planning.
\end{abstract}


\end{frontmatter}
\section{Introduction}\label{sec:intro}

In the era of Industry 5.0, the ability to harness vast amounts of data for optimization, efficiency enhancement and intelligent decision-making has become a key driver of innovation across industries \citep{nikseresht2024blockchain}. Big data analytics, Artificial Intelligence (AI) and game-theoretical modeling are revolutionizing forecasting and resource management by enabling organizations to anticipate demand patterns, mitigate uncertainties and optimize operations. Firms that adopt big data analytics report better forecasting and decision-making \citep{chatterjee2023bda}, and AI has become a driver of efficiency and sustainable growth in tourism \citep{siddik2025ai}. In this context, tourism demand forecasting presents a complex, dynamic challenge where traditional econometric models struggle to handle non-linear dependencies, volatility and seasonality \citep{song2019review}. Effective Tourism Management (TM) is crucial as it directly impacts tourism enterprises, visitor satisfaction and destination competitiveness. Reliable tourism demand predictions enable businesses such as travel agents and hotels to adjust prices and promotions, plan their staff better and increase revenue. Moreover, investors and tourism supporters can identify suitable financing opportunities for lodging facilities, amusement parks or airports to ensure profitability. However, timely demand forecasts are hindered by delayed publishing of tourism demand predictors \citep{choi2012predicting}. To address this, big data from the Internet and information technology offer a potential solution \citep{guizzardi2021dynamic}. Internet big data are available almost in real time and at little or no cost, which gives them an advantage in timeliness over conventional economic data, and even the numbers of likes on the Facebook pages of destination management organizations have improved forecasts of tourist arrivals for some cities \citep{gunter2019facebook}. Data generated through online and social media platforms such as search engines, Facebook and Twitter provide valuable insights into user behavior and interests, and online sources supply a large share of the big data used in tourism research \citep{li2018bigdatareview}. Internet data play a significant role in forecasting tourism trends, as they closely reflect tourist behavior and preferences, making them sensitive to changes in tourist conduct \citep{song2019review}. The Internet serves as a vital source of information for tourists, impacting tourism demand \citep{li2021internetreview}. Internet big data analytics can reveal travelers' destination preferences and effectively forecast tourism demand \citep{li2020multisource}. These Internet big data in TM are sourced from three primary resources, namely (1) people, such as Internet reviews, (2) devices, including cellphone and geolocation data, and (3) operations, such as Google Trends, website traffic and internet-based booking data \citep{li2018bigdatareview}.

Academics have recently started utilizing Internet big data, such as Google Trends, to predict tourism demand and behavior \citep{gunter2019facebook,miah2017bigdata}. Search queries and website traffic data are essential and cost-effective for forecasting. In sales prediction, search data perform at least as well as social media data and improve the accuracy of models built on them \citep{geva2017forum}, but they record what users look for rather than what users think of it. Social media data are more information-rich in this respect, reflecting tourists' preferences and emotions more precisely, so the two kinds of data complement each other. Real-time social media data are valuable for gauging events and attitudes \citep{chen2017smog}, making them beneficial for predicting commercial sales \citep{cui2018operational}, and the conversion of social media communication into demand signals has recently been shown to improve tourism forecasts \citep{qin2025socialmedia}.

In TM, social networking data serve a crucial role in tourists' information-seeking behavior, knowledge-sharing and decision-making. When planning a vacation, travelers often search for destination-specific information online. In comparison to information from the website of an organization or from marketing, customers are becoming more discriminating and more willing to trust the views of their peers \citep{fronzetti2019forums}. These characteristics are especially valuable during uncertain times such as COVID-19, which has impacted every industry, especially tourism and hospitality \citep{filieri2023covid,skare2021covid}. Additionally, the prominence of user-generated internet evaluations has increased the accessibility of digital word of mouth (WOM) \citep{geva2017forum}. Therefore, the significance of online reviews has increased, boosting the appeal of third-party review platforms \citep{fang2016perceived}. Social media user-generated online review data may reflect travelers' tastes, sentiments and the overall reputation of attractions and destinations \citep{geetha2017sentiment,mariani2023devices}. \citet{jacobsen2012search} investigated how electronic and other information sources influence tourists' choice of destination, and found that direct word of mouth, conventional websites and own experience remain highly influential and that conventional websites and social media complement each other. Additionally, some researchers have collected and analyzed data from social media platforms to gain insights from online reviews. Research on tourism demand prediction employing Internet big data is still in its early stages due to the absence of a standardized best practice and the use of diverse approaches. A thorough literature study uncovered a number of research gaps in Internet-based tourism demand forecasting based on big data. There have been only a few efforts to use social media data based on volume and valence for tourism demand forecasts \citep{hu2022review}. More importantly, even fewer papers have considered a mixture of multi-Internet big data in tourism demand forecasting \citep{li2020multisource,gao2025timefeature}. This paper introduces a data-driven optimization framework that integrates game-theoretic big data analytics, time-frequency decomposition and Deep Learning (DL) to improve the accuracy and reliability of forecasting models. By leveraging multi-source Internet big data, this approach enhances decision-making efficiency, enabling businesses and policymakers to allocate resources more effectively. The study aligns with the growing need to integrate data-driven methodologies into industrial decision-making to boost performance and adaptability in an increasingly uncertain economic environment. As a result, the \textbf{first contribution} of this paper is to utilize a mixture of multi-Internet big data sources, considering both volume and valence data features,\footnote{The data pertaining to volume can serve as an indicator of the level of attention that tourists have towards a particular destination. On the other hand, the data related to endorsement or valence can provide insights into the preferences and sentiments of tourists towards the destination \citep{geetha2017sentiment}. Valence data serve as a comprehensive evaluation of the quality of consumers' experiences and can be used to promptly discern their sentiments and attitudes, because the findings obtained by \citet{geetha2017sentiment} indicate that there is a congruity between the valence data provided by customers and their corresponding sentiments towards budget and premium hotels, with the sentiment polarity of the reviews explaining significant variation in the ratings of both hotel categories. The present investigation incorporates variables based on volume and valence into a forecasting system for destination tourism demand. This is achieved through the utilization of a combination of multiple sources of big data obtained from the Internet. Furthermore, the results section includes an analysis that separates the contributions of the volume and valence variables (Section~\ref{sec:hk}).} along with the tourism demand data, to predict a destination's demand. These Internet big data sources are TripAdvisor, Google, Baidu, Facebook and Twitter (now X).

Furthermore, the exchanges between users can be valuable in determining the collective viewpoint of a cohort, and promotional activities that build on such exchanges have occasionally proved advantageous for the industry. The interplay between individuals has an impact on the conduct and cognitive processes of others \citep{jain2020coalitions}. In contemporary times, social networking platforms offer prospects for users to collaborate with one another in order to exchange information. As a result of this collaborative effort, a distinct subset of individuals, commonly referred to as opinion leaders \citep{katz1955personal}, possess the ability to exert influence and alter the conduct and evaluations of others \citep{bamakan2019ol}, and they play distinct roles in how information diffuses on platforms such as Weibo \citep{zhao2024ol}. The process of identifying individuals who hold significant influence and deviate from the norm is commonly referred to as opinion leader detection in social networks \citep{li2019opinioncommunity,parau2017ol}. Opinion leaders are also called ``trendsetters'', ``influencers'', ``opinion makers'' or ``thought leaders'', and they can be categorized in several ways \citep{bamakan2019ol}. By the extent of their influence, global opinion leaders influence large and densely connected communities, whereas local leaders influence a limited set of communities that may be homogeneous or diverse. By domain, monomorphic leaders are influential in a single field, while polymorphic leaders are influential in several fields at once and therefore tend to combine broad knowledge with wide reach. By the character of their influence, leaders can be beneficial or detrimental, since some influential users try to damage the standing of a product by spreading unfavorable information as part of a marketing tactic. Finally, by the duration of their influence, leaders can be short-term or long-term. The influence of opinion leaders has a significant impact on the decision-making process of their followers, as the preferences of the latter are often shaped by the trust and belief dynamics established with their respective opinion leaders \citep{bamakan2019ol}. Opinion leaders are active in a broad range of fields such as political science, healthcare, financial business, marketing, dissemination of information, education and economics, and the methods proposed to detect them are correspondingly diverse \citep{bamakan2019ol}. In contrast to the present investigation, previous research in the field of TM has not explored identifying opinion leaders through game theory. This approach involves the identification of a group of users who exhibit the highest level of synergy, which is then designated as the coalition of opinion leaders \citep{jain2020coalitions}. Subsequently, these opinion leaders and their associated network (coalition) are assigned a greater positive or negative weight, reflecting their heightened impact and significance in relation to the content they generate and endorse. As a result, our \textbf{second contribution} posits that game theory plays a crucial role in identifying opinion leaders within social networks, and it adapts to tourism the coalition approach of \citet{jain2020coalitions}, which combines trust, centrality and cooperative and non-cooperative games. In a social network, the actions undertaken by a user have an impact on the outcomes of the choices and decisions made by another user, and although individuals may lack knowledge regarding the impact of their actions on the outcomes of other users, they are aware that other users observe and react to what they do. Studies of social networks have used the Shapley value to identify influential individuals who can effectively promote and drive the growth of products in the practical business sector \citep{narayanam2011shapley}. The application of game theory can yield valuable insights into the identification of various forms of power centrality and the dynamics of relationships among users within a social network \citep{michalak2013shapley}. Game theory approaches have also been found to address various concerns about network privacy and security \citep{manshaei2013security}. In line with our second contribution, we refrain from indiscriminately extracting and examining all cases using social media data. Instead, we first detect the most impactful communities and opinion leaders through a game-theoretic approach and assign significance weights to them. In this way, the social media analytics that feed the forecasting stage become goal-oriented, and the procedure is consistent with the Pareto principle (80--20 rule), under which a small share of users accounts for a large share of the influence.

In addition, the time series pertaining to tourism demand encompasses four latent sub-components, namely the seasonal element, cycles, trends and an irregular term. Seasonality is the dominant factor in tourism demand time series. Recurring seasonal patterns can be observed in tourism demand time series, such as the number of international tourist arrivals, even upon a cursory examination. Tourism demand cycles, which can endure for multiple years and can be impacted by non-seasonal factors, including economic cycles, exchange rates for currencies, money supply, promotional efforts and terrorism, are concealed behind the prominent seasonal fluctuations \citep{botha2022cycles}. The long-term trend, being one of the four components of the tourism demand time series, can be readily identified, as the rise of global tourism demand over the past four decades shows. Despite the relevance of dynamic behaviors in the tourism demand forecasting literature, there is a dearth of prior research in TM that has taken into account these dynamic statistical behaviors to predict the tourism demand of a destination. Accordingly, as the \textbf{third contribution}, the present study utilizes a capable wavelet-based time-frequency analysis, the Empirical Wavelet Transform (EWT) \citep{gilles2013ewt}, to capture these dynamic statistical data features. It applies the transform causally, so that the decomposition never uses observations beyond the forecast origin \citep{zhang2015hybrid}, and it forecasts the tourism demand with a memory-aware hybrid method based on Long Short-Term Memory (LSTM), namely a robust Stacked BiLSTM model whose hyperparameters are tuned by Bayesian optimization (BO). Decomposition--ensemble designs are now common in technology-enabled forecasting and handle nonstationary and noisy series in other domains as well \citep{mohammadi2023ewt,wang2023aqi}, and the EWT suits tourism data because its filter bank adapts to the spectrum of each series. 

Beyond the empirical application, the paper makes the game-theoretic component analytically tractable, which is our \textbf{fourth contribution}. We formalize trust-weighted influence as a coverage game and prove that its Shapley values have a closed form that can be computed in time linear in the size of the network (Theorem~\ref{thm:shapley}). We show that the coalition of maximal synergy, which enumeration-based procedures search for over all subsets of users, can be found exactly in polynomial time and that the eigenvector-centrality tie-break used in practice is resolved by a unique largest optimal coalition (Theorem~\ref{thm:coalition}). We also characterize the equilibria of the pairwise persuasion games that decide whether two users cooperate (Proposition~\ref{prop:sol4}), show that the associated opinion dynamics converge geometrically to a consensus (Proposition~\ref{prop:dyn}) and bound how errors in the estimated influence probabilities and synergies can change the selected coalition (Proposition~\ref{prop:stab}). Finally, we prove that a full-sample EWT leaks future information and state the conditions under which the forecasting pipeline is free of look-ahead bias (Lemma~\ref{lem:noncausal} and Proposition~\ref{prop:causal}).

In summary, the present study's primary goal is to predict tourists visiting (i) Hong Kong from five major source markets before COVID-19 and (ii) Sanya, located in the Hainan province of China, over a period that includes the COVID-19 shock. This will be achieved by incorporating tourist-related online big data into the destination-predicting system. The present study addresses previously identified research gaps by introducing a new and resilient hybrid methodology that utilizes online data to predict a destination's tourism demand, and its integrated algorithmic components, to the best of our knowledge, have not previously been deployed together in tourism demand prediction. Ablation experiments on the same series isolate the contributions of Internet data, opinion leader detection and the EWT, and the framework is compared with benchmark models reported for the same destinations. Empirical findings indicate that the proposed technique achieves the lowest errors among the configurations and benchmarks considered.

The remainder of this paper is organized as follows. Section~\ref{sec:lit} surveys the relevant literature, pinpointing knowledge gaps in tourism demand forecasting. Section~\ref{sec:method} describes our methodology in detail, encompassing the data preprocessing steps, the hybrid algorithm's architecture, its theoretical properties and the developmental processes involved. Section~\ref{sec:results} presents and interprets the results, including comparisons between our method and various advanced forecasting techniques. Section~\ref{sec:discussion} discusses the theoretical foundations and managerial implications of the proposed approach in the context of tourism demand forecasting. Finally, Section~\ref{sec:conclusion} provides concluding remarks and outlines prospective avenues for future research. Appendices~A--E are placed after Section~\ref{sec:conclusion}. Appendices~A, B and C follow the order of Section~\ref{sec:method}, listing the variables and keywords, detailing the opinion-leader module and describing the stacked architecture, Appendix~D proves the formal results in the order in which they appear and Appendix~E lists the notation.
\section{Literature review}\label{sec:lit}

In this section, a background on tourism demand forecasting, social media analytics and the detection of opinion leaders in social media using game theory is provided.

\subsection{Tourism demand forecasting}\label{sec:lit:tdf}

Various techniques for forecasting have been utilized in the analysis of tourism demand, encompassing both quantitative and qualitative methodologies \citep{law2019deep}. Qualitative techniques such as Delphi and consensus methodologies are frequently utilized to predict tourism demand. These techniques are based on the subjective knowledge of specialists in particular tourism sectors. The limitations that are inherent in qualitative methods, such as their restricted generalizability, have resulted in a growing interest among researchers in utilizing quantitative approaches to estimate relationships among various factors in tourism data. As per the findings of \citet{law2019deep}, two primary approaches are commonly employed to improve the efficacy of quantitative methods. These include adding supplementary variables that may be pertinent and creating more intricate models to enhance the generalizability of outcomes, and the present study follows both. The primary quantitative methodologies employed in the analysis of tourism demand comprise time series analysis, econometric analysis and AI techniques \citep{song2019review}. Time series models are commonly employed, particularly those encompassing Autoregressive Moving Average (ARMA) and its derivatives, such as Autoregressive Integrated Moving Average (ARIMA) and seasonal ARIMA. The models above are utilized to predict forthcoming tourism demand based on past trends and patterns and have been extensively embraced as standard models in associated research \citep{song2019review}. The conventional models encounter challenges in handling intricate and non-linear tourism demand data due to their dependence on the stability of historical patterns and economic structure in real-world scenarios \citep{zhang2021decomposed}. The utilization of AI-based models has garnered significant attention in the pursuit of enhanced non-linear modeling, exhibiting notable predictive precision \citep{zhang2021decomposed,nikseresht2026CIE}. AI models exhibit strong adaptability and powerful processing capacity for non-linear time series, without requiring assumptions of data stationarity and distribution. Based on their layer depth, AI models can be categorized into shallow and DL models \citep{law2019deep}. Radial basis function networks \citep{claveria2015nn}, multi-layer perceptron and support vector regression are commonly utilized shallow learning models. Compared to other models, DL models have demonstrated the ability to overcome such limitations and garnered significant attention in recent research endeavors. LSTM-based methodologies are prevalent among these models. \citet{law2019deep} utilized a DL framework to predict Macau tourism arrivals from a large set of search intensity indices, and their model outperformed several benchmark models, among them support vector regression and an artificial neural network. \citet{zhang2020dailyflow} and \citet{bi2020daily} integrated multiple explanatory variables, such as search engine and weather data, into their LSTM models of daily tourist flows, and \citet{bi2020daily} showed that these variables improved forecasting accuracy. In addition to the utilization of a singular LSTM model, several studies have sought to expand upon the LSTM model in order to mitigate its inherent limitations, including but not limited to model overfitting \citep{gal2016dropout}, a high number of parameters \citep{greff2017lstm} and a sluggish rate of convergence, for example by replacing the LSTM cell with a lighter gated recurrent unit inside a decomposition-based hybrid \citep{wang2021gas}. \citet{kulshrestha2020bilstm} utilized a Bayesian Bidirectional LSTM model to predict the tourism demand in Singapore originating from various prominent source countries, and \citet{tan2026ensemble} combined two recurrent networks in an ensemble to forecast outbound tourism before and during the pandemic. Despite the significant enhancement in forecasting accuracy brought about by DL models, such as LSTM, the pursuit of further improvement in forecasting remains ongoing.

Furthermore, prior research has presented various approaches aimed at enhancing the accuracy of forecasting outcomes, such as the development of combination models to mitigate the risk of forecasting failure \citep{gao2025timefeature}, the construction of decomposition-ensemble models that divide complex forecasting tasks into simpler subtasks to generate more precise forecasts \citep{nikseresht2024ppc,wang2023aqi,zhang2021decomposed}, and the inclusion of additional influencing factors (e.g., data from social media platforms) to better explain and model the non-linear stochastic statistical behaviors in tourism demand time series \citep{hu2022review,li2021internetreview,park2021news}. Outside tourism, news-sentiment indices fed to interpretable machine learning models have improved forecasts of economic cycles under uncertainty shocks \citep{sun2025cycle}, and such cycles are among the drivers of tourism demand cycles. Decomposition-based hybrids must, however, be applied with care, because decomposing a whole series before splitting it into training and test samples lets future values leak into the inputs and makes reported accuracy optimistic \citep{zhang2015hybrid}, and in real-time use the decomposed sub-series change whenever new observations arrive, so that models trained on earlier sub-series may no longer fit them \citep{wang2016practical}. Search data have been combined with mixed-frequency regressions \citep{bangwayo2015midas}, denoising them before modeling improves tourism forecasts \citep{cao2025denoising}, wavelet analysis of a Google search index has dated the phases of the COVID-19 crisis \citep{szczygielski2024crisis}, and \citet{liu2018bigdata} showed with a vector autoregression that web search volumes and visitor arrivals at a cultural destination Granger-cause each other. Turning to multisource designs, \citet{li2020multisource} utilized Internet big data from various sources, such as search engines and online reviews, to predict the number of tourists that would visit a national park, and found that utilizing multiple sources of big data can lead to improved performance in predicting tourist arrivals compared to relying on a single source of big data, such as online reviews or search engines. More recently, \citet{gao2025timefeature} let the weights of a forecast combination vary over time and across the features of economic and search engine data for Hong Kong and Sanya, the two destinations studied here, and \citet{qin2025socialmedia} built platform-specific indices of how online content converts into later visits.

\subsection{Roles of social media analytics}\label{sec:lit:sma}

There is increasing attention towards social media usage in various applications, such as the forecasting of sales and of warranty claims \citep{cui2018operational,warrantyforecast2024}. The impacts of social media information on consumer co-creation \citep{zhang2023cocreation} and purchasing behavior \citep{cui2018operational} have been explored in prior research. The literature distinguishes two distinct effects of such information, the attention effect and the endorsement effect. The effects have been confirmed by pertinent literature, for example in the finding that social media word of mouth raises product awareness while displayed sales act as a signal of quality \citep{li2018herding}, and social media signals have also improved operational forecasts such as those of consumer returns \citep{nikseresht2025returns}. Scholars have argued that the awareness of a product is necessary for consumers to engage in purchase behavior. Consumers' purchasing decisions are ultimately influenced by the purchases and endorsements of their peers. The attention effect pertains to the level of consumer product awareness. Scholars have posited that the allocation of attention can exert an impact on the purchasing behavior of individuals, as well as serve as a tool for predicting economic outcomes. As demonstrated by prior studies \citep{cui2018operational,duan2008reviews}, the attention garnered by social media platforms has the potential to increase the sales of products. The impact of attention is typically quantified through metrics such as count or popularity data. \citet{dellarocas2007reviews} posited that the utilization of volume data is rooted in the notion that heightened customer discourse regarding a product is positively correlated with increased visibility and awareness of the said product among potential consumers, and \citet{duan2008reviews} showed that the volume of reviews indicates the strength of electronic word-of-mouth (eWOM), a form of awareness effect that significantly influences product sales. The phenomenon of the endorsement effect is indicative of the product's perceived quality. The distinction between the endorsement and attention effect lies in the former's ability to convey a consumer's inclination or sentiment orientation, which is indicative of their level of satisfaction and attitude towards a given product. \citet{dellarocas2007reviews} and \citet{duan2008reviews} employed valence data to examine the impact of product endorsement, whereas alternative studies utilized sentiment analysis for the same purpose. The evidence on valence is less clear-cut than that on volume. Once the endogeneity of ratings was taken into account, \citet{duan2008reviews} found no significant effect of the average rating on box office revenues, whereas the volume of online posts remained a significant driver. The sentiment categorization can be divided into three distinct classifications, namely negative, neutral or positive \citep{geetha2017sentiment}. Sentiment has been characterized as displaying diverse levels of intensity or potency \citep{thelwall2010strength}. Individuals have the ability to express their opinions on various social media platforms. Research has demonstrated that sentiment analysis can offer supplementary insights and advantages when compared to data based solely on counts or volumes \citep{lau2018aspect}. \citet{dellarocas2007reviews} expounded upon the rationale for utilizing valence data, positing that unfavorable reviews can dissuade consumers from making a purchase, whereas favorable reviews may stimulate consumer interest and promote sales. Furthermore, a number of scholars have confirmed the impact of the endorsement effect on consumers' buying patterns. The impact of a customer's sentiment on an item's page has been observed to influence the perceptions of others regarding the quality of the product, subsequently affecting the sales of the product \citep{seeto2018reviews}. Big data on millions of hotel reviews further show that the ratings themselves depend on the platform and on the device used to submit them \citep{mariani2023devices}, and user-generated content media differ in how much semantic, sentiment and trend information they carry \citep{park2025ugc}. Both findings are reasons to combine several platforms rather than rely on one.

\subsection{Identification of opinion leaders within social networks}\label{sec:lit:ol}

In contemporary times, there has been a significant shift in human decision-making and attitudes, which has been influenced by the vast amount of data available on the Internet, mainly through social media. This has led to a transformation in individuals' perspectives toward firmly held values. The crucial role of opinion leaders in the dissemination of information and promotional activities is paramount within online social networks. Numerous methodologies have been suggested to ascertain the opinion leader within the realm of online social networks, owing to the progression of technological advancements \citep{bamakan2019ol}. Trust is a crucial factor in the establishment of relationships within social networks \citep{kwan2009trust,mayer1995trust}, and it also shapes how node centrality and link strength are assessed \citep{ruan2016trust}. The degree of connection in a network is influenced by a range of factors, including but not limited to indegree--outdegree covariance, clustering coefficient, network size, density and transitivity \citep{jain2020coalitions}.

Our methodology makes the identification of opinion leaders a constituent element of tourism demand forecasting and performs it with an innovative game-theoretic technique. The approach combines trust statistics, network measures and cooperative and non-cooperative solution concepts, and it addresses three weaknesses of earlier methods, namely their reliance on a single centrality measure, their neglect of how users jointly create value, and the exponential cost of searching over all coalitions.

\subsection{Research gaps}\label{sec:lit:gaps}

Three gaps emerge from this review. First, most Internet-based tourism forecasting studies use one or two kinds of online data, for example search queries \citep{liu2018bigdata,hu2023sanya}, news \citep{park2021news} or reviews \citep{hu2022review}. Studies that combine several platforms are rare \citep{li2020multisource,gao2025timefeature}, and none, to our knowledge, combines search engines, two social networks and a review platform while separating volume from valence. Second, the content entering these models is seldom linked to the influence of the users who produce it. \citet{fronzetti2019forums} showed that the centralization of forum communication networks, which reflects the presence of eminent contributors, helps to forecast arrivals, and \citet{qin2025socialmedia} weighted platforms by how their content converts into visits, but no study weighs individual items of online content by the game-theoretic influence of the users who produce or endorse them. Third, decomposition and deep learning have each improved tourism forecasts \citep{zhang2021decomposed,kulshrestha2020bilstm}, yet their combination with multisource Internet data is scarce, and the causal application of the decomposition, which is necessary for out-of-sample evaluation \citep{zhang2015hybrid}, is seldom discussed. The framework developed in Section~\ref{sec:method} addresses all three gaps.
\section{Methodology}\label{sec:method}

Developing reliable forecasts for business-related time series is instrumental for practical applications in areas such as cost control, profit maximization and resource allocation. In this context, enhancing accuracy directly translates into more effective decision-making and improved planning outcomes. The present study contributes to the business literature, particularly in the domain of tourism demand forecasting, by proposing an advanced forecasting algorithm that bridges the domains of engineering and computer science and offers a robust and innovative prediction framework. The subsequent sections detail the data utilized in this study, outline the data preprocessing steps and describe the core components of our hybrid algorithm. We also explain the evaluation metrics used to assess forecasting performance and establish the theoretical properties of the game-theoretic and decomposition components on which the framework relies.

\subsection{Data and evaluation metrics}\label{sec:data}

\subsubsection{Brief data description}\label{sec:data:desc}

\begin{figure}[!t]
\centering
\includegraphics[width=\textwidth]{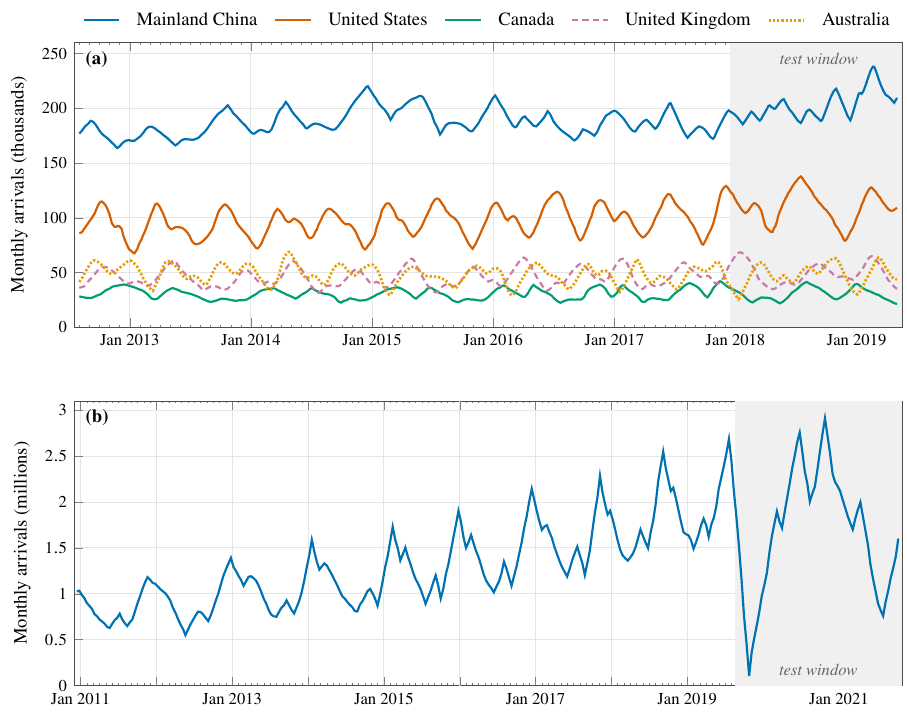}
\caption{\textbf{(a)} Historical monthly time series of the tourist arrivals to Hong Kong from the five chosen source markets, August 2012 to May 2019, and \textbf{(b)} historical monthly tourism demand data in Sanya, January 2011 to October 2021. Shaded areas mark the test windows (last 20\% of each series).}\label{fig:series}
\end{figure}

In this study, we utilize two tourism demand cases concerning the arrivals into two different tourist destinations. We chose two different cases for two reasons. The first is to test the robustness of the proposed hybrid algorithm in versatile settings, and the second is that COVID-19 has had devastating impacts on the tourism industry, so we wanted to see how the method proposed in this paper performs during the pandemic. The first case is Hong Kong. Hong Kong has earned the ``Asia's World City'' moniker due to its distinct cultural amalgamation of Western lifestyle and Chinese traditions \citep{park2021news}. For this case, Internet big data sourced from Baidu, Google, Twitter, Facebook and TripAdvisor are combined with monthly tourist arrival data from five selected source markets. These markets include the most significant short-haul market, \textit{Mainland China}, which accounted for 78.29\% of all tourist arrivals in 2019, and the most substantial long-haul market, \textit{the U.S.}, which constituted 28.09\% of the long-haul market in the same year. The study also considers Canada, the U.K. and Australia as significant source markets. The data on monthly tourist arrivals from the aforementioned countries spanning from August 2012 to May 2019 (82 months per market) were collected from the Hong Kong Tourism Board\footnote{\url{https://www.discoverhongkong.com}} (see Fig.~\ref{fig:series}(a)). As the second case of our study, we take Sanya as the research destination. Sanya is a well-known coastal resort with distinctive tourism attractions and serves as a significant hub for tourism development in Hainan, China. To gather data, we obtained monthly records of tourism demand in Sanya from the official website of the Hainan Provincial Government Tourism Authority (refer to Fig.~\ref{fig:series}(b)). The dataset spanned from January 2011 to October 2021, comprising a total of 130 data points. Following the framework illustrated in Fig.~\ref{fig:framework}, we created a keyword database specific to Sanya tourism and collected search query data. For this case, due to access limitations, we only use Twitter and Baidu. We followed the exact steps of \citet{hu2023sanya} for the Baidu part of the second case. In both cases the first 80\% of the months, in chronological order, are used for training and validation, and the remaining 20\% form the out-of-sample test window. For Hong Kong the test window runs from January 2018 to May 2019 (17 months), and for Sanya it runs from September 2019 to October 2021 (26 months), so it contains the collapse and partial recovery of arrivals during the pandemic.

\subsubsection{Data mining and preprocessing procedure}\label{sec:data:prep}

\begin{figure}[!p]
\centering
\includegraphics[width=\textwidth]{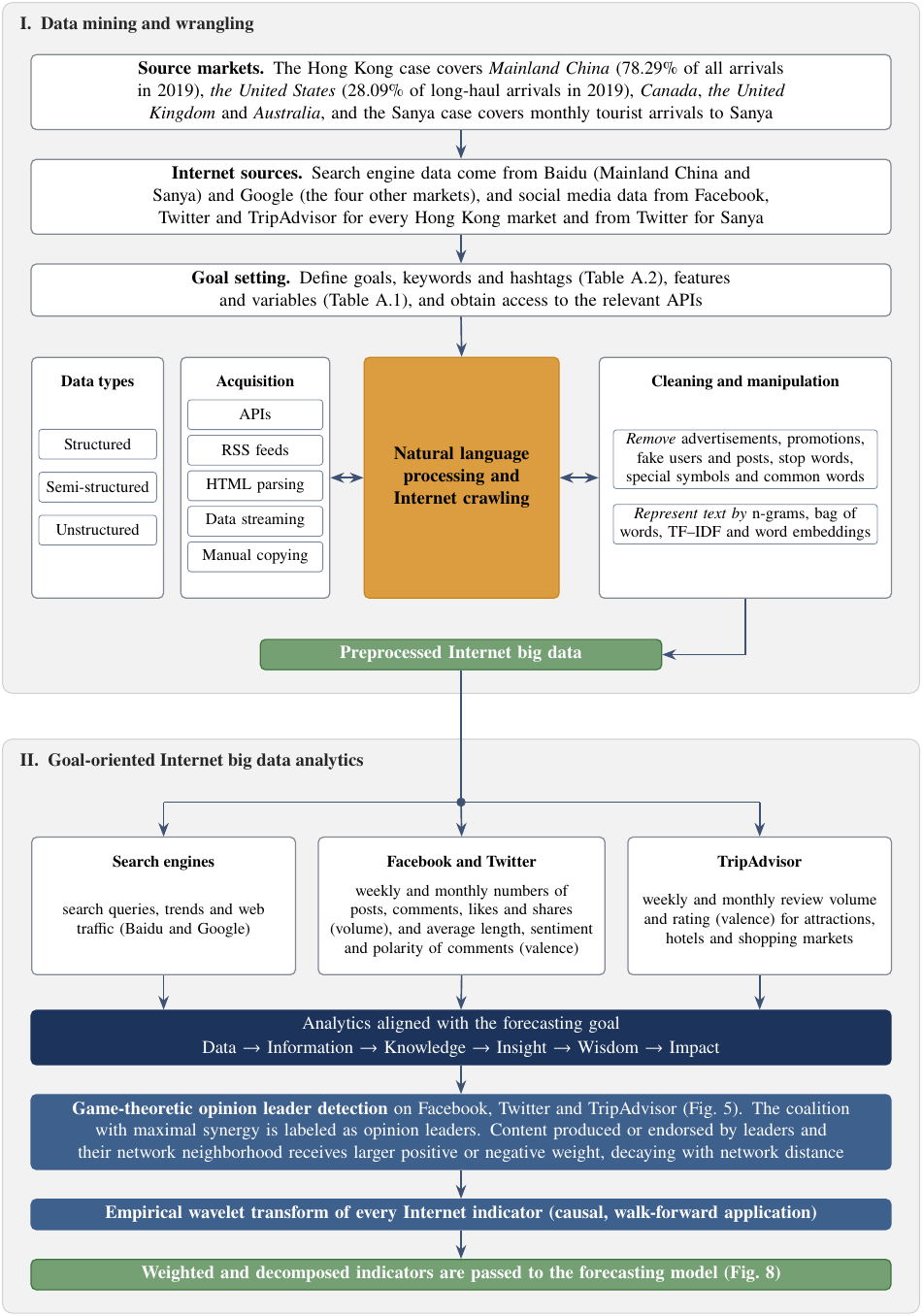}
\caption{Holistic data mining and goal-oriented Internet big data analytics framework.}\label{fig:framework}
\end{figure}

A holistic data mining and goal-oriented Internet big data analytics framework (see Fig.~\ref{fig:framework}) is provided to encapsulate weekly and monthly Internet big data with monthly tourist arrival data in the prediction system. This framework consists of two main phases, namely (\textit{i}) data mining and wrangling and (\textit{ii}) goal-oriented Internet big data analytics. These two phases' most essential parts are summarized in five steps, namely (1) defining the goals and collecting the online big data, (2) processing the data, (3) specifying the required features, performing the opinion leaders' detection and assigning the significance weights (positive or negative) to the data, (4) performing the wavelet-based time-frequency analysis (EWT), and (5) injecting the finalized data and generated variables into the proposed hybrid forecasting approach.

The first stage includes identifying and gathering the online Internet big data. Regarding the search engines, Baidu data (i.e., web traffic, search query and trend data) are only utilized for Mainland China in case one and for case two. For the other four countries of case one, Google information is used. Concerning Facebook, Twitter and TripAdvisor, both volume (attention) and valence (endorsement) types of data are considered for case one. However, for the second case, only Twitter is considered. In order to create user-generated online review variables pertaining to TripAdvisor for a particular destination, namely Hong Kong, data pertaining to online reviews were gathered for the primary tourism-related sectors, including shopping markets, tourist attractions and hotels, following exactly the steps taken by \citet{hu2022review}. The online data cover the same months as the arrival series of each case. \ref{app:features} lists all the variables and features (Table~\ref{tab:features}), the main keywords and hashtags whose combinations were used for Facebook and Twitter data extraction (Table~\ref{tab:keywords}), and the text preprocessing steps, which use the Python NLTK library.

The second step was data processing (shown in Fig.~\ref{fig:framework}), and the TripAdvisor variables were again computed following \citet{hu2022review}. Furthermore, to streamline subsequent analytical steps, the dataset undergoes min--max normalization. After deriving the results, a reverse transformation (de-normalization) is applied to restore the data to its original scale for final evaluation. Formally, let $x_{\min}$ and $x_{\max}$ denote the minimum and maximum values of the raw time series over the training sample, while $x_{\mathrm{low}}$ and $x_{\mathrm{high}}$ represent the corresponding minimum and maximum values of the normalized data. The normalization procedure of the time series from $[x_{\min},x_{\max}]$ into $[x_{\mathrm{low}},x_{\mathrm{high}}]$ is as follows,
\begin{equation}\label{eq:minmax}
\widetilde{X}=\frac{X-x_{\min}}{x_{\max}-x_{\min}}\left(x_{\mathrm{high}}-x_{\mathrm{low}}\right)+x_{\mathrm{low}},
\end{equation}
where $X$ is the observed time series and $\widetilde{X}$ stands for the normalized time series, and the de-normalization is its inverse,
\begin{equation}\label{eq:denorm}
X=x_{\min}+\frac{\widetilde{X}-x_{\mathrm{low}}}{x_{\mathrm{high}}-x_{\mathrm{low}}}\left(x_{\max}-x_{\min}\right).
\end{equation}
Because $x_{\min}$ and $x_{\max}$ are fixed on the training sample, test observations may fall slightly outside $[x_{\mathrm{low}},x_{\mathrm{high}}]$. This is harmless because the output layer of the network is linear, and it keeps the scaling free of information from the test window. Monthly arrivals are also interpolated onto the higher-frequency grid of the Internet indicators (data augmentation), and the forecasts are aggregated back to months (down-sampling) before evaluation (Fig.~\ref{fig:pipeline}). The interpolation at a forecast origin uses only observations up to that origin (Section~\ref{sec:protocol}), and all errors are computed on the original monthly observations, so interpolated values never count as test data.

\begin{figure}[!t]
\centering
\includegraphics[width=0.86\textwidth]{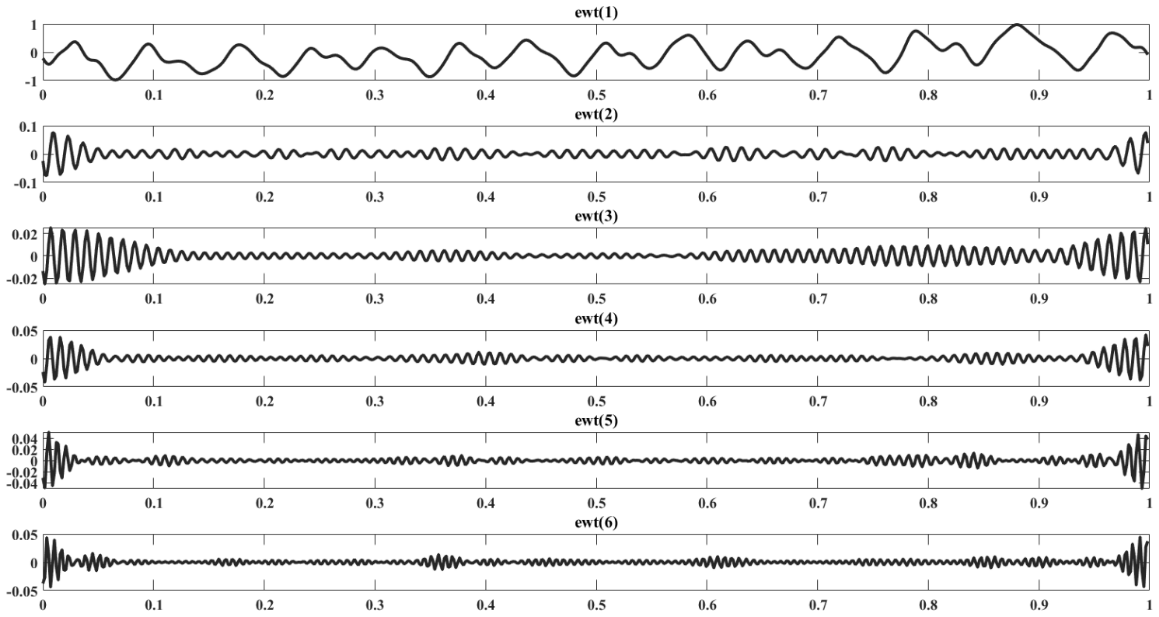}\\[2pt]
{\small (a)}\\[6pt]
\includegraphics[width=\textwidth]{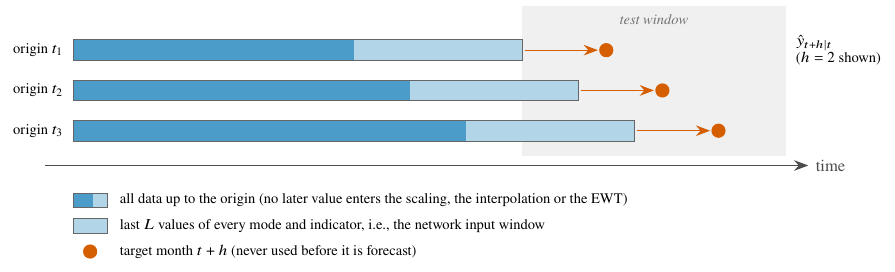}\\[2pt]
{\small (b)}
\caption{\textbf{(a)} A sample of EWT executed on the tourism demand data (the arrivals from the U.S. for case one, six modes on a normalized time axis). \textbf{(b)} Walk-forward (rolling-origin) evaluation with causal decomposition. At every origin the scaling, the interpolation and the EWT use only the data available at that origin.}\label{fig:ewt}
\end{figure}

Afterward, the third step was specifying the opinion leaders via a game-theoretical approach and assigning the significance weights to the data. In the fourth step the series are decomposed with the EWT, of which Fig.~\ref{fig:ewt}(a) shows an actual sample, and in the fifth step the finalized data and generated variables are injected into the proposed hybrid forecasting approach. The whole framework regarding the data manipulation and flow of the data mining, whose text preprocessing steps are listed in \ref{app:features}, and those related to goal-oriented big data analytics are shown in Fig.~\ref{fig:framework}.

\subsubsection{Evaluation metrics}\label{sec:metrics}

Root Mean Square Error (RMSE), Mean Absolute Percentage Error (MAPE) and Root Mean Squared Relative Error (RMSRE) are utilized to identify forecasting errors. A lesser numerical value of said indices demonstrates superior performance. For a forecast horizon $h$, let $\mathcal{T}_h$ be the set of test months whose $h$-step-ahead forecasts are evaluated, $y_t>0$ the observed arrivals and $\hat{y}_t$ the forecast made $h$ months earlier. The three measures are computed as follows,
\begin{align}
\mathrm{MAPE}&=\frac{1}{|\mathcal{T}_h|}\sum_{t\in\mathcal{T}_h}\left|\frac{y_t-\hat{y}_t}{y_t}\right|, \label{eq:mape}\\
\mathrm{RMSE}&=\sqrt{\frac{1}{|\mathcal{T}_h|}\sum_{t\in\mathcal{T}_h}\left(y_t-\hat{y}_t\right)^2}, \label{eq:rmse}\\
\mathrm{RMSRE}&=\sqrt{\frac{1}{|\mathcal{T}_h|}\sum_{t\in\mathcal{T}_h}\left(\frac{y_t-\hat{y}_t}{y_t}\right)^2}, \label{eq:rmsre}
\end{align}
where MAPE and RMSRE are reported as proportions, so that a MAPE of 0.0026 corresponds to 0.26\% \citep{hu2022review,park2021news}. The RMSE is scale-dependent, whereas MAPE and RMSRE are scale free \citep{hyndman2006accuracy}, which is why the three measures are reported together.

\subsection{Proposed hybrid approach}\label{sec:hybrid}

This section develops a multi-staged algorithmic framework as our hybrid solution for tourism demand forecasting comprising game-theoretical detection of opinion leaders in social networks, the EWT and the Stacked BiLSTM model as its backbone. As for the implementation steps, since social media data are mainly unstructured and vast, to make the social media analytics organized and goal-oriented, at the first stage we implement a game-theoretical approach to detect the opinion leaders in tourism and assign them and their vicinity network (coalition) a higher positive or negative weight (showing and considering their more positive or negative impact and significance) with regard to the content that they create and promote. This weighting is the highest for the leaders themselves, and it decreases with the network distance from the leaders. The holistic view regarding the whole process of the utilized Internet big data is shown in Fig.~\ref{fig:framework}. After detecting the leaders and assigning the weights, we go to the second stage and compensate for the cyclical, seasonal and trend features, among others, utilizing the EWT and the stacked version of the BiLSTM. We execute the EWT on all our data, i.e., the Internet big data and tourism demand time series, within the walk-forward protocol of Section~\ref{sec:protocol}. Finally, we do the ultimate forecasting by implementing the stacked version of the BiLSTM that not only considers the memory features of our data but also better deals with the nonstationary features of the data.

Concerning both cases of the current study, the proposed hybrid approach and its ablation variants were estimated on training data, and the main comparisons were reported with regard to the out-of-sample data (test data) forecasts, whereas the results of benchmark models from earlier studies are reported as published. The forecasting outcomes were inspected on the basis of the RMSRE, MAPE and RMSE. In the rest of this section, every component of the proposed hybrid prediction algorithm is detailed. \ref{app:games} details the opinion-leader module, \ref{app:lstm} the stacked architecture and its dropout layers, and \ref{app:proofs} proves all formal results.

\subsubsection{Game-theoretical detection of opinion leaders in social networks}\label{sec:ol}

\begin{figure}[!t]
\centering
\includegraphics[width=0.86\textwidth]{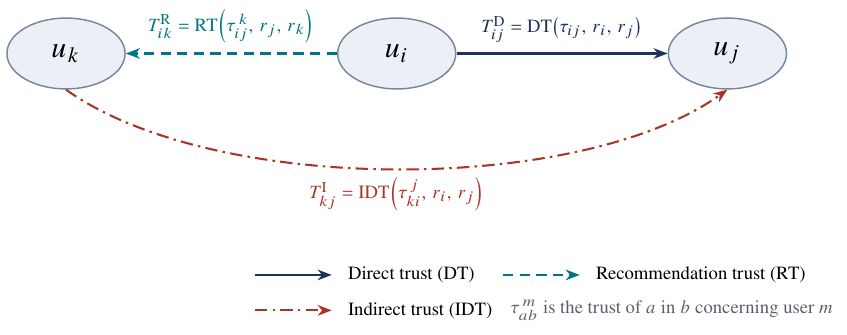}
\caption{Categorization of the trust modes among three users $u_i$, $u_j$ and $u_k$.}\label{fig:trust}
\end{figure}

Detection of opinion leaders in social networking platforms is a complex undertaking. Nonetheless, we aim to identify the users who hold the most influential perspectives within the given social network, and \ref{app:games} begins with a review of the criteria that earlier studies used to select them. The approach combines non-cooperative games between pairs of users, represented in normal or extensive form, with a cooperative game in which every coalition of users has a value \citep{osborne1994course}. Trust relations are represented by an extensive-form structure, namely a trustor--trustee tree generated on the basis of trust. Trust and centrality metrics serve as the primary means of characterizing a user's attributes within a given network \citep{jain2020coalitions}. Our hypothesis posits that users can be likened to players within a network, wherein trust and other centrality measures serve as attributes that aid in determining the marginal contribution of each user in the game. The degree of trust was initially assessed by analyzing the user's frequent interaction patterns. The nature of trust can vary based on the actions of the user, resulting in either unidirectional or bidirectional trust, which is why the trust of $i$ towards $j$ and that of $j$ towards $i$ are distinguished below. Having access to prior information or recommendations regarding an individual's past experiences and performances can heighten the level of trust placed in said individual \citep{jain2020coalitions}. Trust is categorized into three modes, namely direct, indirect and recommendation trust, as illustrated in Fig.~\ref{fig:trust}. Direct trust (DT) refers to a situation where a user trusts another individual without any intervening factors or external influences. In contrast, indirect trust (IDT) refers to a scenario where a user places trust in another individual through the effects they experience as a result of the actions of a third party. Recommendation trust (RT) is a trust mechanism established through the endorsement of other individuals based on their personal experiences.

\textit{Network and trust model.} Let $G=(V,E)$ be the directed interaction network of a platform in the observation window, with $n=|V|$ users. An arc $(i,j)\in E$ records that user $i$ interacted with user $j$ (a reply, mention, comment, share, like or review response), and $n_{ij}\ge 1$ counts these interactions. We write $N(i)$ for the neighbors of $i$ in the underlying undirected graph, $d_i=|N(i)|$ for its degree, $N^{-}(i)=\{j\mid (j,i)\in E\}$ for the users who point to $i$, $d_j^{+}$ for the out-degree of $j$ and $\mathrm{CL}_i$ for the local clustering coefficient of $i$. Interaction intensities are normalized as $\bar{n}_{ij}=(n_{ij}+n_{ji})/\max_{(k,l)\in E}(n_{kl}+n_{lk})\in(0,1]$. The trust degree $T_{ij}$ of user $i$ towards user $j$ is defined as
\begin{equation}\label{eq:trustdegree}
T_{ij}=f\!\left(\tau_{ij},\,\tau_{ji},\,\tau^{\,k}_{ij},\,\tau^{\,k}_{ji},\,r_{i},\,r_{j}\right),
\end{equation}
where $\tau_{ij}$ and $\tau_{ji}$ represent the level of mutual direct trust between users $i$ and $j$, measured by the normalized numbers of interactions from $i$ to $j$ and from $j$ to $i$. On the other hand, the variables $\tau^{\,k}_{ij}$ and $\tau^{\,k}_{ji}$ denote the degree of trust between $i$ and $j$ that pertains to a third user $k$. Additionally, the variables $r_{i}$ and $r_{j}$ signify the reputation of the respective users, which we measure by their normalized recommendation trust defined below. The aggregator $f$ takes values in $[0,1]$ and is nondecreasing in each argument.

The present study examines trust as a crucial element in fostering a connection among users. The trustworthiness of a user reflects three components, namely ability, benevolence and integrity \citep{mayer1995trust}. Together they indicate the cumulative reliability of a given user and the inclination of fellow users to cultivate a social bond with that user. We capture them with the following three user-level metrics, computed on $G$,
\begin{align}
\mathrm{DT}_i&=\sum_{j\in N(i)}\bar{n}_{ij}, \label{eq:dt}\\
\mathrm{IDT}_i&=\sum_{j\in N(i)}\Big[\alpha\,\bar{n}_{ij}+(1-\alpha)\sum_{k\in N(j)\setminus\{i\}}\mathrm{CL}_k\Big], \label{eq:idt}\\
\mathrm{RT}_i&=\alpha\,b_i+(1-\alpha)\sum_{j\in N^{-}(i)}\frac{\chi_{ji}}{d_j^{+}}\,\mathrm{RT}_j, \label{eq:rt}
\end{align}
where $\alpha\in(0,1)$ is the weight of direct evidence relative to second-order evidence in Eq.~\eqref{eq:idt} and to propagated evidence in Eq.~\eqref{eq:rt}, $b_i\in[-1,1]$ is the share of positive minus the share of negative recommendations that user $i$ has received ($b_i=0$ if there are none), and $\chi_{ji}\in\{-1,0,+1\}$ is the polarity of $j$'s recommendation of $i$, with $\chi_{ji}=0$ if $j$ has not recommended $i$. Eq.~\eqref{eq:dt} rewards frequent interaction, Eq.~\eqref{eq:idt} adds the cohesion of the neighbors' neighborhoods, and Eq.~\eqref{eq:rt} lets trust flow along recommendations in the manner of PageRank \citep{page1999pagerank}. Because Eq.~\eqref{eq:rt} is recursive, its well-posedness has to be established.

\begin{proposition}[Recommendation trust is well defined]\label{prop:rt}
Let $\alpha\in(0,1)$, let $\mathbf{R}=(\mathrm{RT}_i)_{i\in V}$ and $\mathbf{b}=(b_i)_{i\in V}$, and let $P_{ji}=\chi_{ji}/d_j^{+}$ if $(j,i)\in E$ and $P_{ji}=0$ otherwise. The system \eqref{eq:rt} has the unique solution $\mathbf{R}^{\ast}=\alpha\big(\mathbf{I}-(1-\alpha)\mathbf{P}^{\top}\big)^{-1}\mathbf{b}$. For any starting vector, the iteration $\mathbf{R}^{(\ell+1)}=\alpha\mathbf{b}+(1-\alpha)\mathbf{P}^{\top}\mathbf{R}^{(\ell)}$ satisfies $\|\mathbf{R}^{(\ell)}-\mathbf{R}^{\ast}\|_1\le(1-\alpha)^{\ell}\|\mathbf{R}^{(0)}-\mathbf{R}^{\ast}\|_1$. Started at $\mathbf{R}^{(0)}=\mathbf{0}$ and stopped after $T_{\max}$ iterations, its error is at most $(1-\alpha)^{T_{\max}}\|\mathbf{b}\|_1$.
\end{proposition}

The proofs of all formal results are collected in \ref{app:proofs}, and \ref{app:notation} lists the notation (Tables~\ref{tab:notation} and \ref{tab:notation2}). Throughout the paper the term ``forecast lead time'' and the symbol $h$ refer to the same quantity. Also, this study employs three constant threshold variables, $\theta_{\mathrm{D}}$, $\theta_{\mathrm{I}}$ and $\theta_{\mathrm{R}}$, which vary within the range $[0.5,1)$ for the three trust types. Each metric is first rescaled to $[0,1]$ across users, $\widetilde{\mathrm{X}}_i=(\mathrm{X}_i-\min_k\mathrm{X}_k)/(\max_k\mathrm{X}_k-\min_k\mathrm{X}_k)$ for $\mathrm{X}\in\{\mathrm{DT},\mathrm{IDT},\mathrm{RT}\}$, and the trust score of user $i$ is
\begin{equation}\label{eq:ts}
\mathrm{TS}_i=\widetilde{\mathrm{DT}}_i\,\1\{\widetilde{\mathrm{DT}}_i>\theta_{\mathrm{D}}\}+\widetilde{\mathrm{IDT}}_i\,\1\{\widetilde{\mathrm{IDT}}_i>\theta_{\mathrm{I}}\}+\widetilde{\mathrm{RT}}_i\,\1\{\widetilde{\mathrm{RT}}_i>\theta_{\mathrm{R}}\}\in[0,3].
\end{equation}
Determining the threshold value for all three variables is contingent upon the specific network and is influenced by its structure and dynamics. The trust score is the user-level counterpart of the trust degree of Eq.~\eqref{eq:trustdegree}, because DT and IDT aggregate the direct and third-party trust levels over the neighbors of $i$ and RT measures the reputation. Users with $\mathrm{TS}_i>0$ form the trusted set $V_T$, of size $n_T$. The trusted users and the links among them form the trust graph $G_T=(V_T,E_T)$, an undirected graph in which $\{i,j\}\in E_T$ whenever $i$ and $j$ interacted in either direction. Its neighborhoods are denoted by $N_T(i)$, and from here on $d_i$ denotes the degree of $i$ in $G_T$. Algorithm~\ref{alg:trust} in \ref{app:games} summarizes the computation.

\textit{Influence probabilities.} The trustor--trustee tree is completed with conditional probabilities of influence. Let $\mathcal{A}_i$ denote the event that user $i$ endorses (likes, shares or positively comments on) a content item, and let $m_i$ be the number of items in the observation window that $i$ endorsed. The probability that user $j$ follows an endorsement by user $i$, in which case $j$ acts as trustor and $i$ as trustee, is obtained from Bayes' theorem,
\begin{equation}\label{eq:bayes}
p_{ij}=\Pr(\mathcal{A}_j\mid\mathcal{A}_i)=\frac{\Pr(\mathcal{A}_i\mid\mathcal{A}_j)\,\Pr(\mathcal{A}_j)}{\Pr(\mathcal{A}_i)},
\end{equation}
and it is estimated by $\hat{p}_{ij}=m_{ij}/m_i$, where $m_{ij}$ counts the items that $j$ endorsed after $i$ had endorsed them \citep{goyal2010influence}. The reliability of these estimates improves with the volume of observed activity, as the next result makes precise.

\begin{proposition}[Accuracy of the influence probabilities]\label{prop:hoeffding}
Suppose that, conditionally on $m_i$, the indicators of $j$ endorsing each of the $m_i$ items after $i$ are independent Bernoulli variables with mean $p_{ij}$. Then $\Pr\big(|\hat{p}_{ij}-p_{ij}|\ge\varepsilon\mid m_i\big)\le2\exp(-2m_i\varepsilon^{2})$ for every $\varepsilon>0$. Consequently, if every trusted user with at least one neighbor in $G_T$ has endorsed at least $m_{\min}$ items, the estimates for all $2|E_T|$ ordered pairs of linked users are simultaneously $\varepsilon$-accurate with probability at least $1-\nu$, for a given $\nu\in(0,1)$, whenever $m_{\min}\ge\ln(4|E_T|/\nu)/(2\varepsilon^{2})$.
\end{proposition}

\textit{Coalitional game and Shapley value.} Upon computation of the trust metric, the subsequent objective is determining the Shapley value \citep{shapley1953value} within a given coalition. Within a network, individuals form a coalition predicated upon their mutual interests, as well as factors such as dyadic and triadic closure, strong and weak ties, and other network-specific considerations \citep{granovetter1973weak}. The present study posits that every player either belongs to the coalition or is in close proximity to at least $\kappa$ neighbors who belong to it. The primary basis for this assertion is that, in actuality, individuals tend to engage with others who exhibit similar conduct and are predisposed to form a collective with them. Accordingly, a user is reached by a coalition if she belongs to it or if at least $\kappa$ of her trusted neighbors do, and the value of a coalition is the trust mass that it reaches. With unit trust scores this is the coverage game of \citet{michalak2013shapley}.

\begin{definition}[Trust-weighted $\kappa$-coverage game]\label{def:game}
For an integer threshold $\kappa\ge1$, the characteristic function on coalitions $C\subseteq V_T$ is
\begin{equation}\label{eq:game}
v(C)=\sum_{u\in V_T}\mathrm{TS}_u\,\1\big\{u\in C\ \text{or}\ |N_T(u)\cap C|\ge\kappa\big\},\qquad v(\emptyset)=0,
\end{equation}
and $\varphi_i(v)$ denotes the Shapley value of user $i$ in this game \citep{shapley1953value}, i.e., the marginal contribution $v(C\cup\{i\})-v(C)$ of $i$ averaged over all $n_T!$ orders in which the trusted users can join the coalition.
\end{definition}

The procedure of \citet{jain2020coalitions} evaluates this average by enumerating permutations, which is impossible for networks with thousands of users. The next theorem, which extends the unweighted coverage result of \citet{michalak2013shapley} to trust weights, removes the difficulty.

\begin{theorem}[Closed-form Shapley value]\label{thm:shapley}
For every trusted user $i$,
\begin{equation}\label{eq:shapleyclosed}
\varphi_i(v)=\mathrm{TS}_i\min\Big\{1,\frac{\kappa}{1+d_i}\Big\}+\sum_{u\in N_T(i)}\mathrm{TS}_u\,\frac{\big(d_u-\kappa+1\big)^{+}}{d_u\,(1+d_u)},
\end{equation}
where $(x)^{+}=\max\{x,0\}$. All Shapley values can therefore be computed in $O(n_T+|E_T|)$ time.
\end{theorem}

\begin{corollary}\label{cor:shapley}
\begin{enumerate}[label=(\roman*),leftmargin=*,itemsep=1pt]
\item Efficiency holds, $\sum_{i\in V_T}\varphi_i(v)=\sum_{u\in V_T}\mathrm{TS}_u$.
\item For $\kappa=1$, $\varphi_i(v)=\sum_{u\in\{i\}\cup N_T(i)}\mathrm{TS}_u/(1+d_u)$.
\item If $\kappa$ exceeds the maximum degree of $G_T$, then $v(C)=\sum_{u\in C}\mathrm{TS}_u$ is additive, $\varphi_i(v)=\mathrm{TS}_i$ for every $i$, and no coalition creates synergy. A meaningful game therefore requires users whose degree is at least $\kappa$.
\end{enumerate}
\end{corollary}

Part (iii) of Corollary~\ref{cor:shapley} shows that node degrees must reach the coverage threshold for the calculation to discriminate between users, because a coalition whose members have fewer than $\kappa$ trusted neighbors each reaches no one outside itself. When the combined marginal contributions of all users within a coalition surpass the sum of marginal contributions of users in any other group, the users within that coalition are deemed to be the opinion leaders within the network. When several coalitions share the same synergy, the coalition whose members possess greater eigenvector centrality is preferred. Theorem~\ref{thm:coalition} below turns this rule into a well-posed optimization problem.

\textit{Pairwise persuasion games.} The employed methodology is distinctive and can produce a wide range of outcomes depending on the parameters of each platform. In practical scenarios, it is not uncommon for certain participants in a game to lack sufficient expertise, resulting in an inaccurate reading of the game. In such instances, it is common for individuals to experience betrayal from their associates and fellow professionals. Novice players may base their decisions on the shrewd maneuvers executed by their counterparts. Under certain circumstances, it is possible that one player's strategy may have an impact on the other player's decisions to a certain degree, leading them to adjust their actions accordingly. Therefore, taking into account the aforementioned facts, four options are evaluated and elaborated upon in \ref{app:games}, along with a detailed analysis of their synergistic effects. In each option, two linked users $A=i$ and $B=j$ choose among keeping their original plan ($\mathsf{o}$), adopting the other's plan ($\mathsf{a}$) and moving to an intermediate agreement ($\mathsf{g}$), where a plan stands for the opinion or content that a user promotes. \ref{app:games} shows that the first two options, whose payoff matrices are given in Eqs.~\eqref{eq:C1sol1} and \eqref{eq:C2sol2}, end in deadlock (Proposition~\ref{prop:sol12}). The third, Eq.~\eqref{eq:C3sol3}, reaches agreement only when the closeness bonus outweighs the cost of compromise and otherwise ends in a social dilemma (Proposition~\ref{prop:sol3}). \ref{app:games} also derives the payoff matrices of the fourth option (Eqs.~\eqref{eq:C4MA} and \eqref{eq:C5MB}) and summarizes the four options in Table~\ref{tab:games}. In the fourth option, the payoff of $A$ is additively separable, $M_A(s_A,s_B)=a_A(s_A)+b_A(s_B)$, with
\begin{equation}\label{eq:sep}
\begin{aligned}
a_A(\mathsf{o})&=\pi_{\mathsf{o}}(1-p_B), & a_A(\mathsf{a})&=-\pi_{\mathsf{a}}p_B, & a_A(\mathsf{g})&=\frac{1}{d_{ij}}-\pi_{\mathsf{g}},\\
b_A(\mathsf{o})&=-\pi_{\mathsf{o}}(1-p_A), & b_A(\mathsf{a})&=\pi_{\mathsf{a}}p_A, & b_A(\mathsf{g})&=\frac{1}{d_{ij}}+\pi_{\mathsf{g}},
\end{aligned}
\end{equation}
and the payoff of $B$ follows by exchanging the roles of $A$ and $B$. Here $\pi_{\mathsf{o}},\pi_{\mathsf{a}},\pi_{\mathsf{g}}>0$ are the values attached to holding the original plan, to altering it and to compromising, $p_A=p_{ij}$ and $p_B=p_{ji}$ are the influence probabilities of Eq.~\eqref{eq:bayes}, and $d_{ij}=d_{ji}>0$ is a symmetric, centrality-based distance whose inverse rewards agreement between close users (Eq.~\eqref{eq:dist} in \ref{app:games}).

\begin{proposition}[Equilibrium of the persuasion game]\label{prop:sol4}
In the fourth option with $p_A,p_B\in(0,1)$, \textup{(i)} altering is strictly dominated by keeping the original plan for both players. \textup{(ii)} Agreement is a strictly dominant strategy for $A$ if $\pi_{\mathsf{o}}(1-p_B)<1/d_{ij}-\pi_{\mathsf{g}}$, and keeping the original plan is strictly dominant if the inequality is reversed, and the same holds for $B$ with $p_A$ in place of $p_B$. \textup{(iii)} Unless one of these inequalities holds with equality, the game has a unique Nash equilibrium \citep{nash1950equilibrium}, and it is mutual agreement if and only if
\begin{equation}\label{eq:agree}
\pi_{\mathsf{o}}\big(1-\min\{p_{ij},p_{ji}\}\big)<\frac{1}{d_{ij}}-\pi_{\mathsf{g}}.
\end{equation}
\end{proposition}

Linked trusted users who satisfy Eq.~\eqref{eq:agree} cooperate. They form the agreement graph $G^{\mathrm{ag}}=(V_T,E^{\mathrm{ag}})$, the edges along which synergy can arise. Two users therefore cooperate when each is sufficiently persuasive to the other and when the distance $d_{ij}$ between them is small. \ref{app:games} also models how the opinions of two such users evolve over rounds of persuasion (Eqs.~\eqref{eq:C6a} and \eqref{eq:C6b}) and shows that the gap between them shrinks geometrically to a consensus that lies between the initial opinions (Proposition~\ref{prop:dyn}).

\textit{Synergy and coalition selection.} The synergy of two cooperating users combines their eigenvector centrality $\mathrm{EC}$ in $G_T$ \citep{bonacich1972factoring} and their Shapley values, and the synergy of a coalition aggregates the pairwise synergies of its members. \ref{app:games} motivates both definitions and illustrates them on a 16-node network (Fig.~\ref{fig:netC1} and Table~\ref{tab:toy}) and on two merging networks (Fig.~\ref{fig:netC2}).
\begin{definition}[Synergy]\label{def:synergy}
For $i,j\in V_T$, let $\Omega_{ij}=\vartheta\cdot\tfrac{1}{2}(\mathrm{EC}_i+\mathrm{EC}_j)\cdot\tfrac{1}{2}(\varphi_i+\varphi_j)\cdot\1\{\{i,j\}\in E^{\mathrm{ag}}\}$ with coordination rate $\vartheta\in(0,1]$, and for a nonempty coalition $C\subseteq V_T$ let
\begin{equation}\label{eq:Phi}
\Phi(C)=\delta\sum_{\substack{i,j\in C\\ i<j}}\left(\frac{\Omega_{ij}}{|C|}\right)^{c},\qquad \delta\in(0,1],\ c\in(0,1].
\end{equation}
\end{definition}

The exponent $c$ controls how strongly size is penalized. The following result shows which members raise the synergy of a coalition, and why the definition favors cohesive groups over large but loosely connected ones.

\begin{proposition}[Admission rule]\label{prop:admission}
For a coalition $C\subseteq V_T$ and a user $k\notin C$, let $S(C)=\sum_{i<j\in C}\Omega_{ij}^{c}$ and $S(k,C)=\sum_{j\in C}\Omega_{kj}^{c}$, so that $\Phi(C)=\delta S(C)/|C|^{c}$, and let $s=|C|$. Then $\Phi(C\cup\{k\})>\Phi(C)$ if and only if $S(k,C)>\big[(1+1/s)^{c}-1\big]S(C)$. In particular, $S(k,C)>cS(C)/s$ is sufficient for admission and $S(k,C)\le cS(C)/(s+1)$ is sufficient for rejection.
\end{proposition}

A newcomer is therefore admitted only if the synergy it creates with the current members exceeds roughly $c$ times the average synergy per member. A user who cooperates with every member of a clique and has with each of them the synergy that the members have with one another is therefore always admitted, whereas a user attached to a single member is admitted only if that one link is strong. The coalition of opinion leaders is defined as a maximizer of $\Phi$, and the next theorem shows that this maximizer can be computed exactly without enumerating the $2^{n_T}-1$ coalitions.

\begin{theorem}[Exact coalition selection]\label{thm:coalition}
Let $S(C)$ be as in Proposition~\ref{prop:admission}, and let $q^{\ast}$ be the maximum of $S(C)/|C|^{c}$ over the nonempty coalitions $C\subseteq V_T$, so that the maximal synergy is $\delta q^{\ast}$. The following statements hold.
\begin{enumerate}[label=(\roman*),leftmargin=*,itemsep=1pt]
\item For every $q\ge0$, the function $F_q(C)=S(C)-q|C|^{c}$ is supermodular on the subsets of $V_T$.
\item The Dinkelbach iteration, which starts from $C^{(0)}=V_T$ and sets $C^{(\ell+1)}\in\arg\max_{C}F_{q^{(\ell)}}(C)$ and $q^{(\ell+1)}=S(C^{(\ell+1)})/|C^{(\ell+1)}|^{c}$, strictly decreases $|C^{(\ell)}|$ for $\ell\ge1$ until $\max_C F_{q^{(\ell)}}(C)=0$, at which point $C^{(\ell)}$ is optimal. It therefore needs at most $n_T+1$ maximizations, each of which is a submodular minimization, so the whole procedure runs in polynomial time, and for $c=1$ each maximization reduces to a minimum $s$--$t$ cut \citep{goldberg1984density}.
\item The optimal coalitions, together with the empty set, are closed under union and intersection. Hence there is a unique inclusion-maximal optimal coalition $C^{\ast}$, and because $\mathrm{EC}_i\ge0$ it has the largest total eigenvector centrality among the optimal coalitions, strictly so if all $\mathrm{EC}_i>0$, as in a connected $G_T$.
\end{enumerate}
\end{theorem}

\begin{figure}[!p]
\centering
\includegraphics[width=0.95\textwidth]{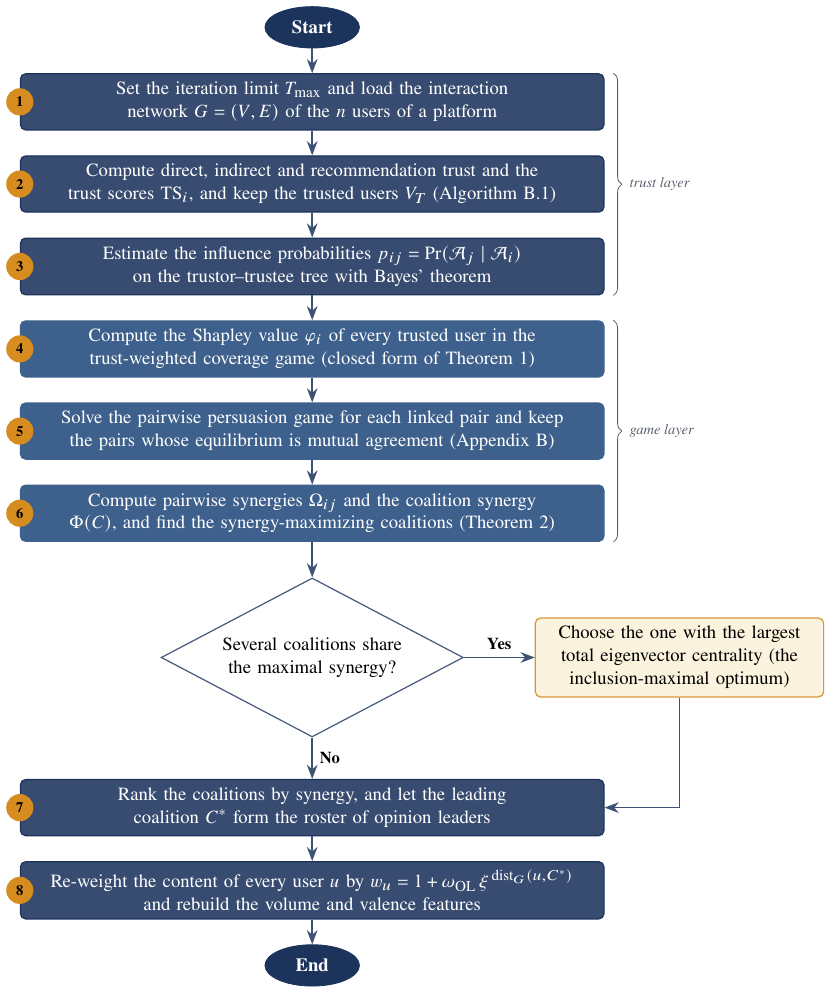}
\caption{Entire procedure of the game-theoretic opinion leader detection algorithm. Steps 1--3 form the trust layer and steps 4--6 the game layer. The closed form of Theorem~\ref{thm:shapley} and the exact algorithm of Theorem~\ref{thm:coalition} replace the enumeration of permutations and coalitions.}\label{fig:olflow}
\end{figure}

Part (iii) reconciles the tie-breaking rule of Fig.~\ref{fig:olflow} with the optimization view, because the union of tied coalitions is itself optimal and has at least their total eigenvector centrality. Algorithm~\ref{alg:coalition} in \ref{app:games} implements part (ii), which applies the parametric method of \citet{dinkelbach1967} to the ratio $S(C)/|C|^{c}$. In the empirical application the coalitions are ranked by $\Phi$, and $C^{\ast}$ yields the roster of opinion leaders. The normalization by $|C|^{c}$ is what makes exact selection tractable. If the number of leaders were fixed in advance at $s$, maximizing $S(C)$ over coalitions of $s$ users would include, for synergies equal to $0$ or $1$, the problem of deciding whether the agreement graph contains a clique of $s$ users, which is NP-complete, so no polynomial algorithm can be expected for that variant unless $\mathrm{P}=\mathrm{NP}$. Proposition~\ref{prop:stab} in \ref{app:games} adds that the selected coalition does not change under small estimation errors in the influence probabilities and the synergies when no linked pair lies close to the agreement threshold of Eq.~\eqref{eq:agree} and the optimality margin of the coalition is large enough.

\textit{Opinion-leader weighting.} Once $C^{\ast}$ is known, every user $u$ receives the weight
\begin{equation}\label{eq:olweight}
w_u=1+\omega_{\mathrm{OL}}\,\xi^{\,\mathrm{dist}_G(u,C^{\ast})},\qquad \omega_{\mathrm{OL}}>0,\ \xi\in(0,1),
\end{equation}
where $\mathrm{dist}_G(u,C^{\ast})$ is the shortest-path distance from $u$ to the coalition, equal to $0$ for leaders and to $\infty$ (so that $w_u=1$) for users not connected to it. For each platform and period $t$, let $\mathcal{E}_{t}$ be the set of posts, comments, likes, shares and reviews, $u(e)$ the user behind item $e$ and $\chi_e\in[-1,1]$ its sentiment polarity. The weighted volume and valence features of the platform are
\begin{equation}\label{eq:wfeatures}
V^{w}_{t}=\sum_{e\in\mathcal{E}_{t}}w_{u(e)},\qquad \bar{\chi}^{w}_{t}=\frac{\sum_{e\in\mathcal{E}_{t}}w_{u(e)}\,\chi_e}{\sum_{e\in\mathcal{E}_{t}}w_{u(e)}},
\end{equation}
and the other features of Table~\ref{tab:features} are weighted in the same way. Because $1\le w_u\le1+\omega_{\mathrm{OL}}$, the weighted valence remains a convex combination of item polarities in $[-1,1]$. Leaders thus tilt the indicators toward their own polarity, positive or negative, without changing their scale, which is the ``higher positive or negative weight'' of the framework, and the weight decays geometrically with the distance from the leaders.

Fig.~\ref{fig:olflow} depicts the comprehensive flowchart of the game-theoretical opinion leader detection methodology proposed in the present study. The parameters lie in the ranges $\theta_{\mathrm{D}},\theta_{\mathrm{I}},\theta_{\mathrm{R}}\in[0.5,1)$, $\alpha,\xi\in(0,1)$, $\vartheta,\delta,c\in(0,1]$, $\kappa\ge1$ and $\omega_{\mathrm{OL}}>0$, and, for the distance and the opinion dynamics of \ref{app:games}, $\lambda,\rho\in[0,0.5]$ and $\mu,\eta\in(0,0.5)$. They are set for each platform according to its structure and dynamics.

\subsubsection{Causal empirical wavelet transform}\label{sec:ewt}

The EWT of \citet{gilles2013ewt} decomposes a series $s(t)$ into $M$ modes with an adaptive filter bank. The Fourier spectrum on $[0,\pi]$ is segmented at boundaries $0=\Delta_0<\Delta_1<\dots<\Delta_{M-1}<\Delta_M=\pi$ placed midway between the $M$ largest local maxima of the spectrum magnitude, and a low-pass filter $\hat{\psi}_1$ and band-pass filters $\hat{\psi}_2,\dots,\hat{\psi}_M$ with smooth transitions of relative width $\gamma$ are built on these bands, as recalled in the proof of Theorem~\ref{thm:ewt} (\ref{app:proofs}). The modes are $s_m=s\ast\psi_m\ast\psi_m$, that is, $\hat{s}_m=\hat{s}\,|\hat{\psi}_m|^{2}$. The example in Fig.~\ref{fig:ewt}(a) has six modes. Two properties matter for forecasting.

\begin{theorem}[Tight frame and exact additivity]\label{thm:ewt}
If $\gamma<\gamma^{\ast}=\min_{1\le m\le M-1}(\Delta_{m+1}-\Delta_m)/(\Delta_{m+1}+\Delta_m)$, then \textup{(i)} $\sum_{m=1}^{M}|\hat{\psi}_m(\omega)|^{2}=1$ for all $\omega\in[-\pi,\pi]$, \textup{(ii)} $\|s\|_2^{2}=\sum_{m=1}^{M}\|s\ast\psi_m\|_2^{2}$ for every finite-energy series $s$, and \textup{(iii)} $\sum_{m=1}^{M}s_m=s$. The same identities hold for the discrete Fourier transform of a finite window.
\end{theorem}

Part (iii) guarantees that the modes passed to the network lose none of the information in the series. Without the bound on $\gamma$, three filters can overlap and the identities fail, and filtering the spectrum only once, $\hat{s}_m=\hat{s}\,\hat{\psi}_m$, would not suffice either, because the modes would then fail to add up to the series in the transition bands. The second property is a warning rather than a guarantee, and Section~\ref{sec:protocol} shows how the framework accounts for it.

\subsubsection{Stacked Bi-directional Long Short-Term Memory model}\label{sec:bilstm}

The proposed model employs a stacked version of BiLSTM and optimizes it using the Bayesian approach to fine-tune hyperparameters. The model comprises stacked LSTM layers, drop-out layer architecture, BiLSTM architecture and hyperparameter tuning through BO. A detailed exposition of the stacked LSTM and drop-out architecture is provided in \ref{app:lstm} (Fig.~\ref{fig:stacked}), together with the number of trainable weights of each layer and the comparison with LSTM layers without dropout. The architecture of the Stacked BiLSTM and its hyperparameter tuning via BO are delineated as follows.

\begin{figure}[!t]
\centering
\includegraphics[width=0.86\textwidth]{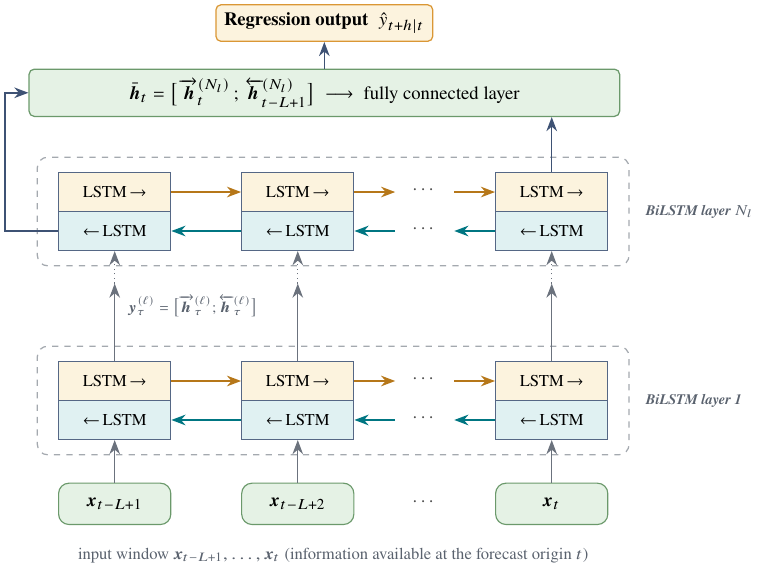}
\caption{Architecture of the proposed Stacked BiLSTM. Each block holds a forward and a backward LSTM cell that read the same input, both directions run inside the input window, which ends at the forecast origin $t$, and the readout joins the last forward state with the last backward state.}\label{fig:bilstm}
\end{figure}

The LSTM model encompasses multiple parameters that impact distinct facets of the model's architecture. Processing the data in both temporal directions captures supplementary features \citep{schuster1997bidirectional,siami2019bilstm}. Fig.~\ref{fig:bilstm} illustrates the optimized Stacked BiLSTM architecture built on this bidirectional processing. At the forecast origin $t$, the network reads the window $\bm{x}_{t-L+1},\dots,\bm{x}_t$, where each vector $\bm{x}_\tau$ stacks the scaled arrivals, their EWT modes and the weighted Internet indicators with their modes. An LSTM cell \citep{gers2000forget,hochreiter1997lstm} updates its hidden state $\bm{h}_\tau$ and memory cell $\bm{c}_\tau$ through input, forget and output gates, which we write compactly as $(\bm{h}_\tau,\bm{c}_\tau)=\mathcal{U}(\bm{x}_\tau,\bm{h}_{\tau-1},\bm{c}_{\tau-1})$. The combination of bidirectional Recurrent Neural Networks (RNNs) and LSTM yields the BiLSTM model, which enables access to the training data in both forward and backward directions \citep{schuster1997bidirectional,graves2013speech}. A BiLSTM layer runs one LSTM forward and another, with its own weights, backward over the window,
\begin{equation}\label{eq:bilstm}
\begin{gathered}
\overrightarrow{\bm{h}}_\tau=\overrightarrow{\mathcal{U}}\big(\bm{x}_\tau,\overrightarrow{\bm{h}}_{\tau-1},\overrightarrow{\bm{c}}_{\tau-1}\big),\qquad
\overleftarrow{\bm{h}}_\tau=\overleftarrow{\mathcal{U}}\big(\bm{x}_\tau,\overleftarrow{\bm{h}}_{\tau+1},\overleftarrow{\bm{c}}_{\tau+1}\big),\\
\bm{y}_\tau=\big[\overrightarrow{\bm{h}}_\tau;\overleftarrow{\bm{h}}_\tau\big],
\end{gathered}
\end{equation}
for $\tau=t-L+1,\dots,t$, with zero initial states at both ends of the window. The forward pass therefore uses $\bm{h}_{\tau-1}$ and the backward pass uses $\bm{h}_{\tau+1}$, and the output $\bm{y}_\tau$ of the layer at step $\tau$ concatenates the two hidden states. In addition, deep architectures were employed through the stacking of BiLSTM layers, where the output sequence of one layer was utilized as the input sequence for the subsequent layer \citep{graves2013speech}. The computation of hidden sequences in a stack with $N_l$ layers is performed in a sequential manner, starting from $\ell=1$ to $N_l$ and $\tau=t-L+1$ to $t$,
\begin{equation}\label{eq:stack}
\overrightarrow{\bm{h}}^{(\ell)}_\tau=\overrightarrow{\mathcal{U}}^{(\ell)}\big(\bm{y}^{(\ell-1)}_\tau,\overrightarrow{\bm{h}}^{(\ell)}_{\tau-1},\overrightarrow{\bm{c}}^{(\ell)}_{\tau-1}\big),\qquad
\overleftarrow{\bm{h}}^{(\ell)}_\tau=\overleftarrow{\mathcal{U}}^{(\ell)}\big(\bm{y}^{(\ell-1)}_\tau,\overleftarrow{\bm{h}}^{(\ell)}_{\tau+1},\overleftarrow{\bm{c}}^{(\ell)}_{\tau+1}\big),
\end{equation}
with $\bm{y}^{(0)}_\tau=\bm{x}_\tau$ and $\bm{y}^{(\ell)}_\tau=[\overrightarrow{\bm{h}}^{(\ell)}_\tau;\overleftarrow{\bm{h}}^{(\ell)}_\tau]$, so that each hidden layer receives input from both the forward and backward layers at the lower level. Dropout with rate $N_d$ is applied to the output of every layer (\ref{app:lstm}), and the network outputs are provided as follows,
\begin{equation}\label{eq:output}
\hat{y}_{t+h\mid t}=\bm{w}_y^{\top}\bar{\bm{h}}_t+b_y,\qquad \bar{\bm{h}}_t=\big[\overrightarrow{\bm{h}}^{(N_l)}_{t};\overleftarrow{\bm{h}}^{(N_l)}_{t-L+1}\big],
\end{equation}
where $\bar{\bm{h}}_t$ joins the last state of the forward pass and the last state of the backward pass of the top layer, and the fully connected layer and the regression layer of Fig.~\ref{fig:bilstm} are written as one affine map for brevity. The backward state at position $t$ has seen only $\bm{x}_t$, which is why the readout takes the backward state at the start of the window, the only one that summarizes the whole window. Both directions operate inside the window that ends at the origin $t$, so bidirectional processing never reaches beyond the information available when the forecast is made.

\subsubsection{Stacked BiLSTM hyperparameter tuning via BO}\label{sec:bo}

\begin{figure}[!t]
\centering
\includegraphics[width=0.92\textwidth]{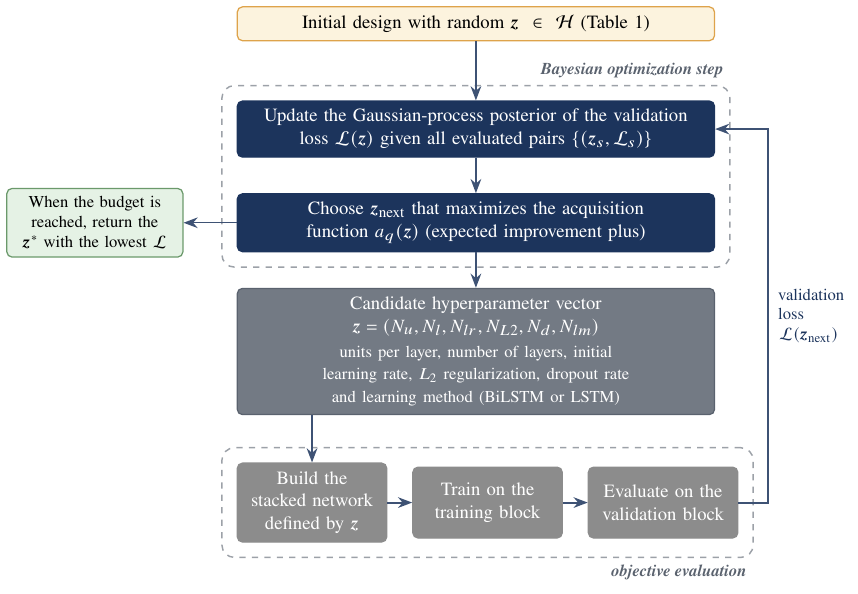}
\caption{The BO algorithm.}\label{fig:bo}
\end{figure}

Predicted outcomes in DL models are profoundly influenced by the choice of model hyperparameters, a determination that can be both labor-intensive and time-intensive when done by hand. Exhaustive grid search is susceptible to the curse of dimensionality, becoming markedly less efficient as the range and number of hyperparameters grow. Random search was introduced to alleviate some of grid search's limitations, but it can still be inadequate for more intricate model architectures. As a result, many view hyperparameter tuning as a computational optimization problem, prompting the adoption of BO \citep{snoek2012practical} to streamline this process. BO constructs a prior over the objective or fitness function, here the validation regression loss, and incrementally refines that prior by incorporating insights from previously sampled hyperparameter sets. The underlying mechanism relies on a Gaussian process to model the fitness function, while an acquisition function directs the selection of the next hyperparameter point. In this study, the ``expected-improvement-plus'' criterion serves as the acquisition function $a_{\mathrm{EI}}(\bm{z})$, guiding the identification of promising hyperparameter configurations $\bm{z}$. Table~\ref{tab:hyper} outlines the hyperparameters investigated, and Fig.~\ref{fig:bo} illustrates the flow of the BO procedure. This methodological choice reduces the protracted search times seen in random search, ensuring a more targeted exploration of hyperparameter space and leading to superior forecasting performance \citep{snoek2012practical,jones1998efficient}.

\begin{table}[!t]
\centering
\caption{BiLSTM hyperparameters and corresponding value ranges, which are optimized with the help of the suggested BO algorithm.}\label{tab:hyper}
\small
\begin{tabular}{@{}llll@{}}
\toprule
Hyperparameter functionality & Symbol & Value range & Search scale\\
\midrule
No.\ of units in each layer & $N_u$ & $\{60,\dots,250\}$ & integer\\
No.\ of layers & $N_l$ & $\{1,\dots,8\}$ & integer\\
Initial learning rate & $N_{lr}$ & $[10^{-2},\,1]$ & logarithmic\\
$L_2$ regularization & $N_{L2}$ & $[10^{-10},\,10^{-2}]$ & logarithmic\\
Drop-out rate & $N_d$ & $[0.1,\,0.9]$ & linear\\
Learning method & $N_{lm}$ & \{BiLSTM, LSTM\} & categorical\\
\bottomrule
\end{tabular}
\end{table}

In this setting, the hyperparameter search space is denoted as $\mathcal{H}$, the Cartesian product of the ranges in Table~\ref{tab:hyper}, and the objective is the validation loss
\begin{equation}\label{eq:boobj}
\mathcal{L}\colon\mathcal{H}\to\mathbb{R}_{+},\qquad \bm{z}=(N_u,N_l,N_{lr},N_{L2},N_d,N_{lm})\mapsto\mathcal{L}(\bm{z}),
\end{equation}
that is, the root mean squared error on a validation block formed by the last observations of the training sample \citep{bergmeir2012cv}, of the network trained with hyperparameters $\bm{z}$. To determine the optimal configuration of model hyperparameters, BO seeks
\begin{equation}\label{eq:boargmin}
\bm{z}^{\ast}\in\arg\min_{\bm{z}\in\mathcal{H}}\mathcal{L}(\bm{z}).
\end{equation}
In BO, two key elements underpin the minimization process, namely the Gaussian process model $\mathcal{L}(\bm{z})\sim\mathcal{GP}\big(\bar{\mathcal{L}},k(\bm{z},\bm{z}')\big)$, with constant prior mean $\bar{\mathcal{L}}$ and kernel $k$, and a Bayesian update procedure that incrementally refines this model after each new evaluation of $\mathcal{L}$. With posterior mean $\hat{\mathcal{L}}(\bm{z})$, posterior standard deviation $\varsigma(\bm{z})$ and best observed loss $\mathcal{L}^{\mathrm{best}}$, the acquisition function is the expected improvement $a_{\mathrm{EI}}(\bm{z})=\mathbb{E}\big[\max\{\mathcal{L}^{\mathrm{best}}-\mathcal{L}(\bm{z}),0\}\big]$, which has a closed form in $\hat{\mathcal{L}}(\bm{z})$ and $\varsigma(\bm{z})$ \citep{jones1998efficient}. The ``plus'' variant modifies the kernel whenever the chosen point is over-exploited, i.e., when $\varsigma(\bm{z})$ falls below a fixed fraction of the noise standard deviation, which keeps the search from stalling.

\subsubsection{Leakage-free evaluation protocol and implementation}\label{sec:protocol}

Decomposition-based forecasts are easy to evaluate incorrectly. The next lemma shows why a full-sample EWT cannot be used to produce test forecasts.

\begin{lemma}[EWT modes are two-sided]\label{lem:noncausal}
Let $M\ge2$ and $\gamma<\gamma^{\ast}$ as in Theorem~\ref{thm:ewt}, and let $g_m$ be the impulse response with $\hat{g}_m=|\hat{\psi}_m|^{2}$, so that $s_m(t)=\sum_{\tau\in\mathbb{Z}}g_m(\tau)s(t-\tau)$. Then $g_m$ is real and even, and $g_m(\tau)\ne0$ for some $\tau>0$. Hence $s_m(t)$ depends on future values $s(t+\tau)$ with weight $g_m(-\tau)=g_m(\tau)\ne0$.
\end{lemma}

If the EWT were applied once to the whole series, the modes at a test month would contain information from later months, and the reported accuracy would be optimistic, a pitfall documented for other decomposition hybrids \citep{zhang2015hybrid}. We therefore use a walk-forward (rolling-origin) protocol \citep{tashman2000outofsample}, illustrated in Fig.~\ref{fig:ewt}(b). At each origin $t$ in the test window, (a) the scaling uses the training statistics of Eq.~\eqref{eq:minmax}, (b) interpolation onto the higher-frequency grid uses observations up to $t$, (c) the EWT is applied to the segment ending at $t$, with boundaries detected once on the training sample so that the meaning of each mode stays fixed even though re-decomposition changes the modes as new observations arrive \citep{wang2016practical}, (d) the network reads the last $L$ values of every mode and indicator, and (e) the network weights, the hyperparameters $\bm{z}^{\ast}$ and the coalition $C^{\ast}$ are estimated on the training sample only. Let $\mathcal{I}_t$ be the information generated by arrivals and Internet data up to month $t$.

\begin{proposition}[No look-ahead]\label{prop:causal}
Under steps \textup{(a)--(e)}, every forecast $\hat{y}_{t+h\mid t}$ in the test window is a measurable function of $\mathcal{I}_t$. In particular, the test target $y_{t+h}$ does not enter the inputs, the scaling, the decomposition or the model selection that produce $\hat{y}_{t+h\mid t}$.
\end{proposition}

\begin{figure}[!p]
\centering
\includegraphics[width=\textwidth]{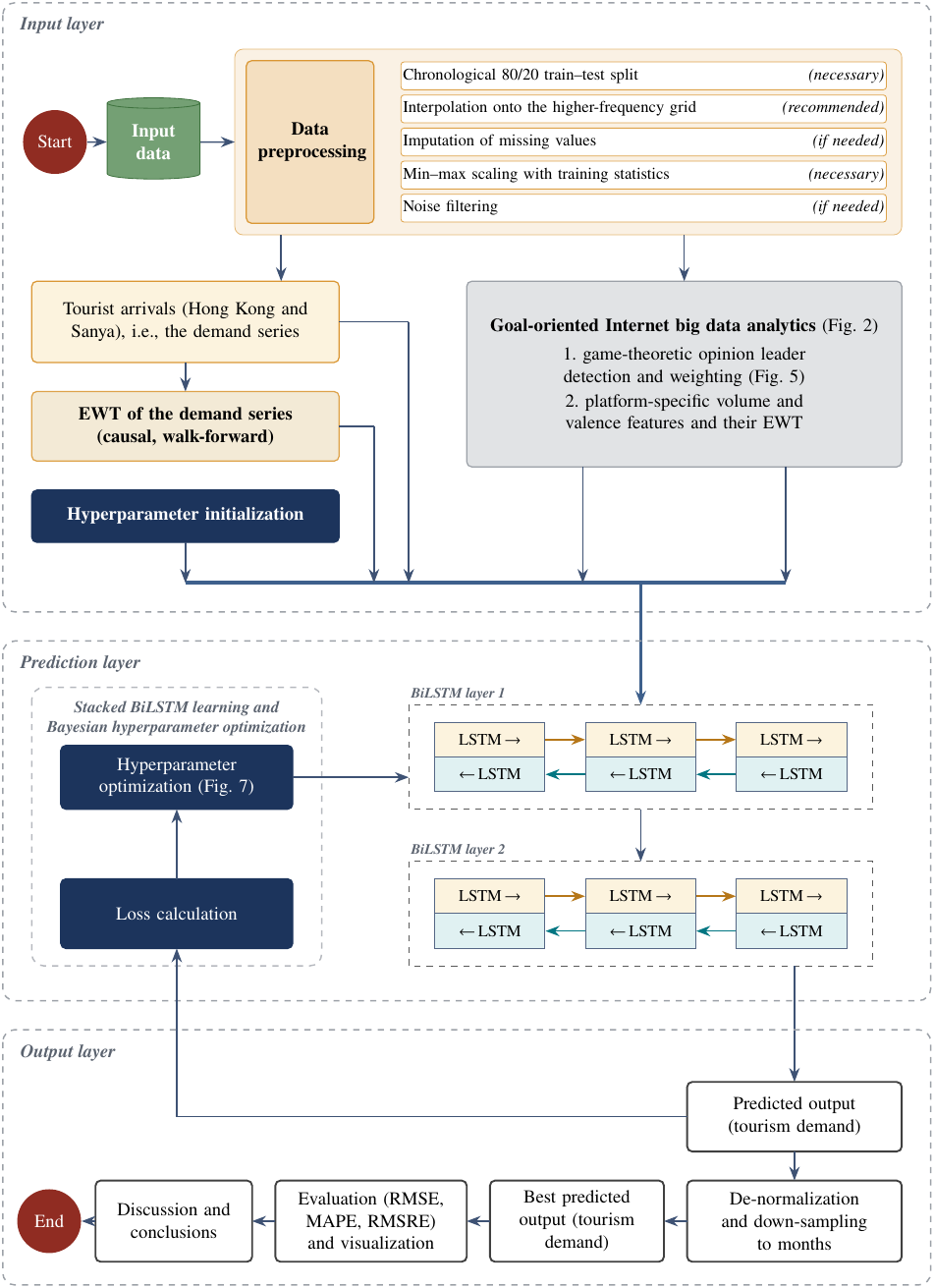}
\caption{Holistic view of the entire execution steps of the developed methodology.}\label{fig:pipeline}
\end{figure}

Proposition~\ref{prop:causal} is the formal counterpart of the statement that only out-of-sample data are used for reporting. Forecasts produced on the higher-frequency grid are de-normalized with Eq.~\eqref{eq:denorm} and aggregated to months before the error measures of Eqs.~\eqref{eq:mape}--\eqref{eq:rmsre} are computed (Fig.~\ref{fig:pipeline}). All the execution steps of the hybrid algorithm are coded in MATLAB R2021b, and the text preprocessing uses the Python NLTK library.

\section{Empirical results}\label{sec:results}

We consider various scenarios and ablation experiments to demonstrate the efficacy and robustness of our proposed hybrid approach. First, the results of the prediction of tourism demand of the two cases, i.e., Hong Kong and Sanya, are demonstrated and discussed, respectively. Next, by comparing the prediction accuracy across these scenarios and with the results of other state-of-the-art methods, we quantify the contributions of the Internet big data, game-theoretical opinion leader detection and time-frequency analysis to the improvement of tourism demand predictions, and we compare the framework built on the Stacked BiLSTM with published benchmarks. Please note that for all of the performance measurements, only the out-of-sample data (test data) are used for the reporting, in line with Proposition~\ref{prop:causal}.

\begin{figure}[!t]
\centering
\includegraphics[width=\textwidth]{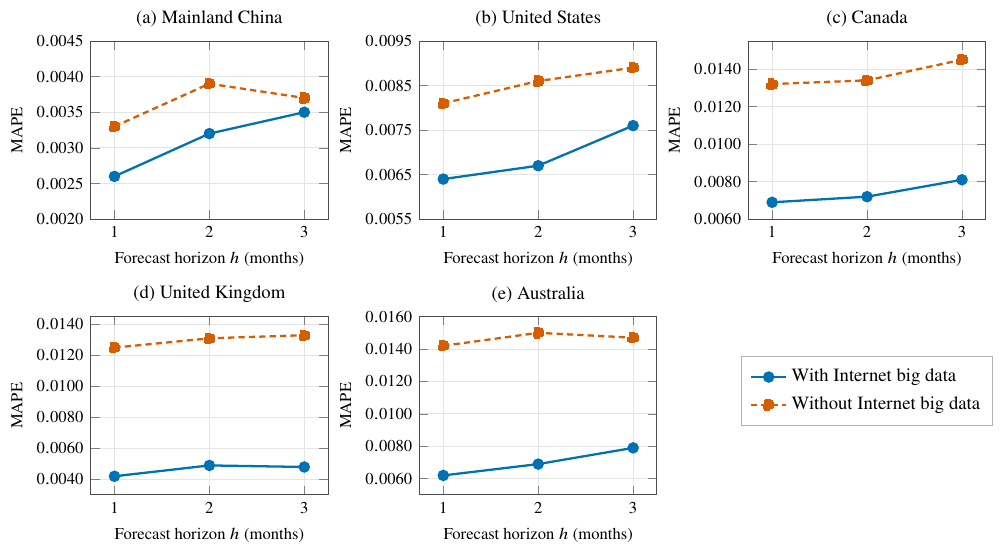}
\caption{MAPE over the prediction horizon (as a proportion). (a) Mainland China, (b) the U.S., (c) Canada, (d) the U.K.\ and (e) Australia. ``With Internet big data'' denotes the full framework, including opinion-leader weighting and the EWT, and ``Without Internet big data'' the same network fed only with past arrivals.}\label{fig:mape}
\end{figure}

\subsection{Hong Kong case}\label{sec:hk}

\subsubsection{Summary of the results}\label{sec:hk:summary}

Fig.~\ref{fig:mape} illustrates how the MAPE of the two generated predictions, i.e., considering and not considering the Internet big data, varies with the forecast lead time. Internet big data, processed with opinion-leader weighting and the EWT, lower the MAPE at every horizon and in every market, by 5.4--21.2\% for Mainland China, 14.6--22.1\% for the U.S., 44.1--47.7\% for Canada, 62.6--66.4\% for the U.K.\ and 46.3--56.3\% for Australia.

Moreover, the findings show that forecast errors generally grow as the forecasting horizon increases. They rise from $h=1$ to $h=2$ in all ten series, and at $h=3$ they rise further in seven series and level off slightly in the remaining three (Mainland China and Australia without Internet data, and the U.K.\ with Internet data). This outcome aligns with expectations that further-into-the-future predictions inevitably face greater uncertainty, as historical data become less reliable for capturing future volatility, uncertainty, complexity and ambiguity (VUCA) \citep{nikseresht2022vuca}. The gain from Internet data is largest at $h=1$ in four of the five markets and smallest at $h=3$ in four of them, which is consistent with Internet signals carrying information about imminent travel decisions. Readily accessible Internet big data therefore add the most value at short horizons, although they still reduce the MAPE by 5.4--63.9\% at $h=3$.

We do the same analysis for tourism demand by considering the various combinations of the opinion leaders' detection and EWT analysis in terms of RMSRE. The findings are shown in Table~\ref{tab:olewt}. The improvement percentage attributable to the opinion leaders' detection and EWT analysis is calculated based on
\begin{equation}\label{eq:improve}
\mathrm{Improvement}=\frac{\mathrm{EM}_0-\mathrm{EM}_1}{\mathrm{EM}_0}\times100\%,
\end{equation}
where $\mathrm{EM}_1$ is an evaluation metric (RMSRE, RMSE or MAPE) of the configuration that includes the component under study and $\mathrm{EM}_0$ is the same metric for the configuration without it. In Table~\ref{tab:olewt}, $\mathrm{EM}_1$ and $\mathrm{EM}_0$ belong to the variants with and without the EWT. The EWT reduces the RMSRE in all 30 comparisons, by 3.42--11.48\% when opinion leaders are considered and by 3.13--8.96\% when they are not. Opinion leader detection matters even more. With the EWT, it lowers the RMSRE by 7.3--16.1\% for Mainland China, 13.2--16.0\% for the U.S., 27.1--28.7\% for Canada, 48.7--50.5\% for the U.K.\ and 13.2--22.0\% for Australia, and the two components together reduce the RMSRE by 15.0--53.4\% relative to the configuration that uses neither. Valuable insights can therefore be gained from the information conveyed through opinion leaders' detection and EWT analysis across all lead times.

\begin{table}[!t]
\centering
\caption{Forecast accuracy improvements when considering and not considering opinion leaders and the EWT analysis (in terms of RMSRE). Relative improvements follow Eq.~\eqref{eq:improve}, with $\mathrm{EM}_1$ the RMSRE of the variant with the EWT and $\mathrm{EM}_0$ that of the variant without it.}\label{tab:olewt}
\small
\setlength{\tabcolsep}{5pt}
\begin{tabular}{@{}c ccc ccc@{}}
\toprule
& \multicolumn{3}{c}{Considering opinion leaders} & \multicolumn{3}{c}{Not considering opinion leaders}\\
\cmidrule(lr){2-4}\cmidrule(l){5-7}
Forecast & With & Without & Rel.\ impr. & With & Without & Rel.\ impr.\\
lead time $h$ & EWT & EWT & (\%) & EWT & EWT & (\%)\\
\midrule
\multicolumn{7}{@{}l}{\textit{Mainland China}}\\
1 & 0.0044 & 0.0048 & 8.33 & 0.0050 & 0.0052 & 3.85\\
2 & 0.0047 & 0.0052 & 9.62 & 0.0056 & 0.0061 & 8.20\\
3 & 0.0051 & 0.0055 & 7.27 & 0.0055 & 0.0060 & 8.33\\
\multicolumn{7}{@{}l}{\textit{U.S.}}\\
1 & 0.0084 & 0.0089 & 5.62 & 0.0100 & 0.0109 & 8.26\\
2 & 0.0090 & 0.0096 & 6.25 & 0.0107 & 0.0112 & 4.46\\
3 & 0.0092 & 0.0101 & 8.91 & 0.0106 & 0.0115 & 7.83\\
\multicolumn{7}{@{}l}{\textit{Canada}}\\
1 & 0.0098 & 0.0103 & 4.85 & 0.0136 & 0.0144 & 5.56\\
2 & 0.0102 & 0.0109 & 6.42 & 0.0143 & 0.0151 & 5.30\\
3 & 0.0113 & 0.0117 & 3.42 & 0.0155 & 0.0160 & 3.13\\
\multicolumn{7}{@{}l}{\textit{U.K.}}\\
1 & 0.0054 & 0.0061 & 11.48 & 0.0109 & 0.0116 & 6.03\\
2 & 0.0060 & 0.0066 & 9.09 & 0.0117 & 0.0121 & 3.31\\
3 & 0.0058 & 0.0063 & 7.94 & 0.0115 & 0.0119 & 3.36\\
\multicolumn{7}{@{}l}{\textit{Australia}}\\
1 & 0.0092 & 0.0099 & 7.07 & 0.0118 & 0.0125 & 5.60\\
2 & 0.0096 & 0.0103 & 6.80 & 0.0122 & 0.0134 & 8.96\\
3 & 0.0105 & 0.0111 & 5.41 & 0.0121 & 0.0132 & 8.33\\
\bottomrule
\end{tabular}
\end{table}

To gain further insights, we tested the algorithm's performance under different configurations of volume and valence features derived solely from social media sources (i.e., Twitter, Facebook and TripAdvisor), excluding the opinion leader detection component. This approach allows us to discern how endorsement (valence) and attention (volume) signals influence predictive accuracy. Table~\ref{tab:features_sm} compares the outcomes of our hybrid model across three settings, namely utilizing all social media variables (Full), including only endorsement (valence) factors and relying exclusively on attention (volume) elements.

In all 15 market--horizon cases, the full set gives the lowest RMSE, 1.2--6.1\% below attention features alone and 0.3--3.7\% below endorsement features alone. Endorsement features alone outperform attention features alone in every case, which agrees with the view that valence conveys perceived quality that counts cannot capture (Section~\ref{sec:lit:sma}). At $h=1$, each social-media-only configuration also beats the arrivals-only network of Table~\ref{tab:bench} in every market, so each family of features carries information on its own, and the full social media set is in turn outperformed by the complete Internet set that adds the search engine data (row b of Table~\ref{tab:bench}). Combining both feature families thus yields a consistent, if modest, improvement over either family alone, which indicates that the interplay between attention and endorsement signals contributes to forecasting success.

\begin{table}[!t]
\centering
\caption{Comparison of out-of-sample RMSE across different feature set considerations using social media data, without opinion leader detection. The last two columns give the reduction of the full set relative to each single family, computed with Eq.~\eqref{eq:improve}.}\label{tab:features_sm}
\small
\begin{tabular}{@{}c rrr rr@{}}
\toprule
& \multicolumn{3}{c}{RMSE with social media features} & \multicolumn{2}{c}{Reduction by Full (\%)}\\
\cmidrule(lr){2-4}\cmidrule(l){5-6}
Horizon & \multicolumn{1}{c}{Full} & \multicolumn{1}{c}{Attention} & \multicolumn{1}{c}{Endorsement} & \multicolumn{1}{c}{vs.} & \multicolumn{1}{c}{vs.}\\
$h$ & \multicolumn{1}{c}{set} & \multicolumn{1}{c}{features} & \multicolumn{1}{c}{features} & \multicolumn{1}{c}{Attention} & \multicolumn{1}{c}{Endorsement}\\
\midrule
\multicolumn{6}{@{}l}{\textit{Mainland China}}\\
1 & 973.502 & 986.087 & 977.196 & 1.28 & 0.38\\
2 & 985.103 & 996.745 & 989.320 & 1.17 & 0.43\\
3 & 994.902 & 1018.003 & 1007.144 & 2.27 & 1.22\\
\multicolumn{6}{@{}l}{\textit{U.S.}}\\
1 & 1013.876 & 1036.529 & 1017.221 & 2.19 & 0.33\\
2 & 1025.481 & 1051.211 & 1039.364 & 2.45 & 1.34\\
3 & 1023.258 & 1050.502 & 1037.321 & 2.59 & 1.36\\
\multicolumn{6}{@{}l}{\textit{Canada}}\\
1 & 471.503 & 502.312 & 487.547 & 6.13 & 3.29\\
2 & 479.714 & 510.109 & 495.463 & 5.96 & 3.18\\
3 & 496.228 & 516.228 & 507.011 & 3.87 & 2.13\\
\multicolumn{6}{@{}l}{\textit{U.K.}}\\
1 & 597.273 & 617.146 & 609.998 & 3.22 & 2.09\\
2 & 612.335 & 629.663 & 621.797 & 2.75 & 1.52\\
3 & 607.778 & 638.874 & 630.931 & 4.87 & 3.67\\
\multicolumn{6}{@{}l}{\textit{Australia}}\\
1 & 505.121 & 519.746 & 512.852 & 2.81 & 1.51\\
2 & 523.301 & 537.986 & 531.126 & 2.73 & 1.47\\
3 & 539.075 & 563.857 & 550.129 & 4.40 & 2.01\\
\bottomrule
\end{tabular}
\end{table}

\subsubsection{Benchmark models and comparisons}\label{sec:hk:bench}

In this subsection, the developed novel hybrid algorithm is compared against several state-of-the-art models in the context of tourism demand forecasting. Table~\ref{tab:bench} reports the one-step-ahead RMSE values for each approach, highlighting the predictive performance of our method relative to existing benchmarks. The benchmark values are the best results reported in the original studies, which use other test windows. \citet{hu2022review} evaluated 36 one-step forecasts from July 2016 to June 2019 for seven English-speaking markets that include our four long-haul markets, and \citet{park2021news} evaluated 12 one-step forecasts from December 2018 to November 2019, a window that includes the social unrest of mid-2019. In scale-free terms, the MAPE of the best model of \citet{park2021news} is 0.1623 for Mainland China and 0.1030 for the U.S., and the main mixed-frequency model of \citet{hu2022review} attains 0.0319--0.0755 across the four long-haul markets, whereas the full framework attains 0.0026--0.0069 at $h=1$ (Fig.~\ref{fig:mape}). The controlled comparison is the one among the three variants of the proposed framework, which share the series, the test window and the network. Internet big data reduce the RMSE by 4.1--41.7\%, opinion leader detection reduces it by a further 10.1--49.4\%, and the two together reduce it by 18.4--67.2\% relative to the arrivals-only network.

\begin{table}[!t]
\centering
\caption{One-step-ahead RMSE values obtained by state-of-the-art methods and the developed one to forecast tourism demand.}\label{tab:bench}
\small
\setlength{\tabcolsep}{4pt}
\begin{tabularx}{\linewidth}{@{}>{\raggedright\arraybackslash}X rrrrr@{}}
\toprule
Model & \multicolumn{1}{c}{China} & \multicolumn{1}{c}{U.S.} & \multicolumn{1}{c}{Canada} & \multicolumn{1}{c}{U.K.} & \multicolumn{1}{c}{Australia}\\
\midrule
Topic modeling + SARIMAX \citep{park2021news}$^{\dagger}$ & 655897 & 11121 & NA & NA & NA\\
SARIMA-MIDAS \citep{hu2022review}$^{\dagger}$ & NA & 4912.31 & 1267.96 & 2526.74 & 3869.27\\
\addlinespace
\multicolumn{6}{@{}l}{\textit{Proposed hybrid algorithm}}\\
\hangindent=1.8em\hspace*{0.9em}(w) without Internet big data & 1061.156 & 1038.809 & 593.470 & 780.483 & 804.962\\
\hangindent=1.8em\hspace*{0.9em}(b) with Internet big data & 948.357 & 996.411 & 409.572 & 506.418 & 468.938\\
\hangindent=1.8em\hspace*{0.9em}(l) with Internet big data and opinion leaders & \textbf{852.957} & \textbf{847.391} & \textbf{279.624} & \textbf{256.417} & \textbf{362.260}\\
\addlinespace
\hangindent=0.9em Reduction from Internet big data, w$\to$b (\%) & 10.63 & 4.08 & 30.99 & 35.11 & 41.74\\
\hangindent=0.9em Reduction from opinion leaders, b$\to$l (\%) & 10.06 & 14.96 & 31.73 & 49.37 & 22.75\\
\hangindent=0.9em Total reduction, w$\to$l (\%) & 19.62 & 18.43 & 52.88 & 67.15 & 55.00\\
\bottomrule
\end{tabularx}
\par\smallskip
\begin{minipage}{0.97\linewidth}\footnotesize
$^{\dagger}$Best output reported in the original study, on a different test window. (w) without Internet big data consideration (only using historical tourism demand time series), (b) with Internet big data consideration, (l) with Internet big data and opinion leaders' detection consideration.
\end{minipage}
\end{table}

\subsection{Sanya (COVID-19) case}\label{sec:sanya}

COVID-19 has had devastating impacts on various aspects of the tourism and hospitality industry \citep{filieri2023covid,skare2021covid}. In order to check the robustness of the proposed hybrid predictive algorithm in terms of dealing with pandemic-related VUCA and its devastating impacts, in this section, we apply the algorithm to the Sanya data, whose test window contains the onset of the pandemic.

To demonstrate the forecasting performance of the proposed novel hybrid algorithm implemented on the Sanya data, the results are shown in Table~\ref{tab:sanya}. This table compares the proposed algorithm with five models reported by \citet{hu2023sanya} for the same Sanya series and Baidu search features, in terms of the RMSE and MAPE.

\begin{table}[!t]
\centering
\caption{RMSE and MAPE of five models reported by \citet{hu2023sanya} and of the developed framework for one-step-ahead forecasts of Sanya tourism demand.}\label{tab:sanya}
\small
\begin{tabularx}{\linewidth}{@{}c>{\raggedright\arraybackslash}X rr@{}}
\toprule
No. & Model & \multicolumn{1}{c}{RMSE} & \multicolumn{1}{c}{MAPE}\\
\midrule
1 & PCA + SVM & 37789 & 0.015\\
2 & KPCA + SVM & 37789 & 0.015\\
3 & ISOMAP + SVM & 72784 & 0.024\\
4 & LLE + SVM\_PROPHET & 88537 & 0.035\\
5 & TSNE + SARIMAX\_F & 217929 & 0.083\\
\addlinespace
6 & \textit{Proposed hybrid algorithm}, (w) without Internet big data & 37058 & 0.014\\
 & \textit{Proposed hybrid algorithm}, (b) with Internet big data & 35102 & 0.013\\
 & \textit{Proposed hybrid algorithm}, (l) with Internet big data and opinion leaders & \textbf{32061} & \textbf{0.011}\\
\bottomrule
\end{tabularx}
\par\smallskip
\begin{minipage}{0.97\linewidth}\footnotesize
Models 1--5 as reported by \citet{hu2023sanya}. (w) without Internet big data consideration (only using historical tourism demand time series), (b) with Internet big data consideration, (l) with Internet big data and opinion leaders' detection consideration. MAPE is reported as a proportion.
\end{minipage}
\end{table}

From the Table~\ref{tab:sanya} results, the proposed hybrid algorithm attains the lowest RMSE and MAPE. Internet big data reduce the RMSE by 5.28\% and the MAPE by 7.14\%, and opinion leader weighting brings further reductions of 8.66\% and 15.38\%. The full framework lowers the RMSE by 15.16\% relative to the best benchmarks (PCA + SVM and KPCA + SVM, 37,789). Since the test window covers the collapse and partial recovery of arrivals in 2020 and 2021, these results indicate that the framework is robust enough to deal with VUCA conditions such as those of COVID-19.

\section{Discussion and implications}\label{sec:discussion}

This study utilized various tourism-related Internet big data sources to develop a hybrid destination tourism demand forecasting approach. The approach was applied to two cases to predict the number of tourist arrivals. The first case concerns Hong Kong from five key source countries before COVID-19, and the second is Sanya, over a period that includes the pandemic. The present study utilizes internet big data sourced from prominent platforms such as Baidu, Google, Twitter, Facebook and TripAdvisor. The TripAdvisor platform incorporated evaluations of prominent tourist destinations, diverse lodging categories and significant commercial markets for tourism in Hong Kong. In the context of Facebook, Twitter and TripAdvisor, the analysis of data encompasses both quantitative measures of volume and qualitative assessments of valence. Furthermore, before integrating the Internet's vast data into our hybrid forecasting methodology, we employ a game-theory-based algorithm for detecting opinion leaders from diverse social media data. This algorithm identifies a cohort of users with the highest degree of synergy, which is referred to as the coalition of opinion leaders. We offer a robust Stacked BiLSTM model to perform the final forecasting, which captures long and short memory in the data. The results suggest that (a) tourism demand forecasting based upon multiple Internet big data sources can improve forecasting accuracy and precision by a large margin, (b) the utilized combination makes it possible for the developed method to be sufficiently dynamic to treat the VUCA of the data, and (c) the proposed hybrid approach outperformed the benchmark models considered when multiple-frequency data from various Internet big data sources, along with a game-theoretical opinion leader detection method and the hybrid memory-based algorithm, were incorporated into the forecasting exercise. In what follows, significant theoretical and managerial insights and implications are given.

\subsection{Theoretical contributions}\label{sec:theory}

\begin{itemize}[leftmargin=*,itemsep=2pt]
\item Our research extends multisource forecasting \citep{li2020multisource} by combining diverse Internet big data sources, including electronic word-of-mouth, to forecast tourism demand. Previous studies primarily focused on singular or dual data sources for destination-level demand prediction. Relying exclusively on attention-based or Internet traffic data, such as search queries and web traffic, is inadequate as it may fail to account for unfavorable conditions despite increasing tourist attention. Augmenting imperfect data with relevant information is essential for improving forecasting precision, and Table~\ref{tab:features_sm} shows that valence features are the more informative of the two social media families.
\item This paper captures the seasonal, cyclical, trend and other dynamics of tourism demand through the memory-based algorithm (robust Stacked BiLSTM model) and the EWT time-frequency algorithm.
\item This research extends mixed-frequency modeling in tourism demand forecasting \citep{bangwayo2015midas,hu2022review}. The utilization of the mixed-frequency approach is a proficient approach to circumvent the loss of information that may occur when higher-frequency data are aggregated \citep{bangwayo2015midas}. The findings of this research demonstrate that combining interpolation-based mixed-frequency inputs with the EWT and a stacked BiLSTM produced more precise predictions than the published results of alternative techniques such as MIDAS \citep{hu2022review}, or models that incorporate aggregated online review data as explanatory factors and benchmark forecasting algorithms. Hence, the present study makes a valuable contribution to the advancement of model development in the domain of tourism demand forecasting.
\item Building on the strong predictive capacity of our Stacked BiLSTM model, there is significant scope for incorporating additional predictor variables and features into the suggested algorithm to further refine tourism demand forecasts. The model's architecture affords considerable flexibility, enabling analysts to expand its input dimensions, with the BO stage adjusting the network to the new inputs.
\item Furthermore, our framework supports short-term prediction strategies, one to three months ahead, while retaining the potential for mid- and long-term forecasting improvements, for example through network architectures designed for long-range dependencies, such as the dilated causal convolutions of \citet{vandenoord2016wavenet}.
\item According to the results, our suggested algorithm is robust across markets with very different levels and dynamics and through the COVID-19 shock.
\item The BiLSTM architecture exhibits a high degree of flexibility and robustness when applied to various data sets, and BO is capable of estimating the necessary parameter set for each of them. The superior performance of the proposed design can be attributed to a search space that includes the depth of the stack and the dropout rate, so that BO can select deep, regularized architectures when the data support them. All of the proposed architectures feature base LSTM layers that retain prior information and account for temporal dependencies.
\item The study used game theory to identify opinion leaders from Internet big data, which brings social influence into the forecasting model in a principled way. A trustor--trustee tree was created to represent trust dynamics among users. Four alternative methods for acquiring payoffs, based on different parameters and conditions, were proposed in \ref{app:games}, and their equilibria were characterized (Propositions~\ref{prop:sol4}, \ref{prop:sol12} and \ref{prop:sol3}). In extensive games with imperfect information, individual users may not know other players' strategies. Similarly, in social networks, users may lack awareness of each other and rely on shared experiences and recommendations to build trust, and the persuasion dynamics of Proposition~\ref{prop:dyn} show how two such users converge to a common opinion that lies between their initial ones.
\item The concept of Shapley value (Definition~\ref{def:game}) was employed in our study to detect opinion leaders in Internet big data. This approach allowed us to determine the marginal contribution of a user within a coalition. The Shapley value is utilized to ascertain the requisite expectations for establishing synergy. Theorems~\ref{thm:shapley} and \ref{thm:coalition} show that both the Shapley values and the coalition of maximal synergy can be computed exactly at scale, which removes the main computational objection to game-theoretic opinion leader detection on large platforms, and Proposition~\ref{prop:stab} shows that the selected coalition is robust to small estimation errors when its optimality margin is large enough.
\item The opinion leader detection approach has the advantage of improving its reliability as the volume of observed interactions increases. Proposition~\ref{prop:hoeffding} shows that the estimation error of the influence probabilities shrinks at the rate $1/\sqrt{m}$ in the number $m$ of observed endorsements, so busier networks yield more accurate trust and influence estimates and, in turn, a more reliable coalition.
\end{itemize}

\subsection{Managerial insights}\label{sec:managerial}

\begin{itemize}[leftmargin=*,itemsep=2pt]
\item From a managerial perspective, accurate tourism demand forecasting is critical for businesses in the travel and tourism industry. It enables businesses to optimize their operations, make informed decisions and improve their profitability. By using Internet big data, time-frequency analysis and a hybrid memory-aware prediction approach, businesses can develop more accurate and reliable forecasting models, which can help them stay ahead of the competition and adapt to changing market conditions.
\item Additionally, since we combined various Internet big data via the data mining procedure shown in Fig.~\ref{fig:framework}, we can have better forecasting results and demonstrate Internet big data's true power and value in the tourism industry.
\item By incorporating opinion leader detection into the analysis of Internet big data, this study avoids a blanket examination of all social media posts. Rather, we employ a game-theoretic approach to identify highly influential communities and opinion leaders, assigning them proportional weights to reflect their importance. This method not only yields more targeted and effective social media insights for the forecasting process, but also aligns with the Pareto principle (80--20 rule), wherein the majority of influential outcomes can be traced back to a relatively small subset of individuals and discussions. Destination marketers can monitor the roster of leaders directly, because a shift in their sentiment moves the weighted valence indicators before it shows up in arrivals.
\item Accurate and timely forecasting results should be utilized by policymakers in destinations like Hong Kong and Sanya for the purpose of planning, for example to schedule staff, transport capacity and crowd management one to three months ahead.
\item Our new method allows policymakers to identify factors influencing variations in projected tourist arrivals. By employing an algorithm to detect opinion leaders among tourists on social media platforms, decision-makers can pinpoint the sectors with the most significant impact on destination demand fluctuations, for example by comparing the weighted review indicators of shopping, attractions and hotels. The algorithm identifies a coalition of opinion leaders, representing users with the highest level of synergy. The use of multisource Internet big data further enhances the advantages of this approach.
\end{itemize}

\section{Conclusions and future work}\label{sec:conclusion}

This study demonstrated that harnessing multisource Internet big data can substantially enhance the precision of tourism demand forecasts. By incorporating data from multiple online channels, we achieved high accuracy in predicting monthly tourist arrivals for Hong Kong (from five key source countries) and Sanya, with lower errors than those reported for several advanced forecasting methods in the literature. Central to our approach is a game-theory-based algorithm, which identifies highly influential user groups, referred to as the coalition of opinion leaders, within social media platforms. By detecting these opinion leaders, we gain deeper insight into real-time tourist sentiments and preferences, thereby refining the predictive power of our model. Relative to the same network fed only with past arrivals, the full framework reduces the one-step-ahead RMSE by 18.4--67.2\% across the Hong Kong markets and by 13.5\% for Sanya, and opinion leader detection alone accounts for reductions of 10.1--49.4\% and 8.7\%, respectively.

Our framework also features a robust Stacked BiLSTM model, which is particularly effective in handling non-stationary time-series data with complex seasonal, cyclical and trend components. The adoption of a memory-based algorithm, in tandem with the EWT time-frequency method, helps the proposed forecasting system cope with high levels of VUCA. Experiments show that this hybrid algorithm remains robust in various scenarios, including pre-COVID-19 conditions and the pandemic shock, and can efficiently manage diverse data sets laden with noise and sudden shifts in tourist behavior. On the theoretical side, we showed that the Shapley values of the trust-weighted coverage game have a closed form, that the coalition of maximal synergy is computable exactly in polynomial time and, given a sufficient margin, robust to small estimation errors, that the pairwise persuasion games generically have unique equilibria that decide which users cooperate, and that a causal application of the EWT is necessary for out-of-sample evaluation. As far as current knowledge permits, the algorithmic constituents in this study have not been employed together in prior tourism demand research. This work on multisource Internet big data analytics allows destination authorities, management businesses and stakeholders to obtain timely and accurate monthly forecasts, while also facilitating crowd management and long-term strategic planning. Moreover, the insights gleaned from social media enable the continuous tracking of tourists' interests, thereby further refining policymaking and enhancing the competitiveness of destinations.

Looking to the future, several directions can push this line of research forward. First, researchers may explore alternative optimization approaches for tuning hyperparameters to potentially improve performance and reduce computational overhead. Additionally, expanding data collection to multilingual social media platforms could broaden the model's applicability and augment the generalizability of findings across diverse tourism markets. Another promising avenue involves segmenting online communities by employing techniques such as dynamic topic modeling, testing the resulting network associations with the quadratic assignment procedure (QAP), followed by targeted application of the proposed framework to each identified subgroup. Re-estimating the benchmark models on identical test windows, adding formal tests of predictive accuracy and analyzing the sensitivity of the results to the parameters of the opinion-leader module would further strengthen the evidence. Finally, adopting influential factor analysis methods, such as SHapley Additive exPlanations (SHAP) \citep{lundberg2017shap} and partial dependence plots (PDP), may offer deeper interpretations of the forecasting outputs by clarifying the contributions of various predictors, as Shapley-based explanations have recently done for deep demand forecasts \citep{shajalal2024forecastexplainer} and for news-based forecasts of economic cycles \citep{sun2025cycle}. Such explanations would also connect naturally with the Shapley values that already drive the selection of opinion leaders. Because none of its components is specific to tourism, the framework can also be tested on demand planning in logistics, supply chains and production.

It is anticipated that these advancements will further underscore the utility of Internet big data within TM and beyond. As innovative uses of large-scale online data continue to emerge, this work can inspire researchers to integrate both structured and unstructured data from diverse Internet sources, ultimately refining models of demand forecasting and offering actionable, data-driven insights for better tourism planning and policy formulation.

\newpage
\FloatBarrier
\appendix
\counterwithin*{figure}{section}
\counterwithin*{table}{section}
\counterwithin*{proposition}{section}
\counterwithin*{algorithm}{section}
\renewcommand{\theproposition}{\Alph{section}.\arabic{proposition}}
\renewcommand{\thealgorithm}{\Alph{section}.\arabic{algorithm}}
\setcounter{figure}{0}
\setcounter{table}{0}

\section{Internet big data features}\label{app:features}

Table~\ref{tab:features} summarizes the variables extracted from each platform, and Table~\ref{tab:keywords} lists the main keywords and hashtags whose combinations were used to collect the social media data.

\begin{table}[!ht]
\centering
\caption{Summary of the used Internet big data features.}\label{tab:features}
\small
\begin{tabularx}{\linewidth}{@{}>{\raggedright\arraybackslash}p{0.17\linewidth}>{\raggedright\arraybackslash}p{0.14\linewidth}>{\raggedright\arraybackslash}p{0.2\linewidth}>{\raggedright\arraybackslash}X@{}}
\toprule
Category & Platform & Type of data & Variables\\
\midrule
Search engine data & Google & Search & Search queries, trends and web traffic\\
 & Baidu & Search & Search queries, trends and web traffic\\
\addlinespace
Social media data & TripAdvisor & Volume and valence & Review volume and rating for attractions, hotels and shopping markets, as in \citet{hu2022review}\\
\addlinespace
 & Facebook & Volume (attention) & Numbers of comments, posts, likes and shares\\
 & & Valence (endorsement) & Average length, sentiment and polarity of comments\\
\addlinespace
 & Twitter & Volume (attention) & Numbers of comments, posts, likes and shares\\
 & & Valence (endorsement) & Average length, sentiment and polarity of comments\\
\bottomrule
\end{tabularx}
\end{table}

\begin{table}[!ht]
\centering
\caption{Main keywords and hashtags whose combinations were used for social media data extraction.}\label{tab:keywords}
\small
\begin{tabular}{@{}>{\raggedright\arraybackslash}p{0.47\textwidth}>{\raggedright\arraybackslash}p{0.47\textwidth}@{}}
\toprule
Main keywords and hashtags & Main keywords and hashtags\\
\midrule
Tourism-related words (e.g., travel) & Economy-related words (such as growth, rise)\\
Online tourism-market-related words (e.g., tourism platform) & Customer-related words (such as anger, satisfaction)\\
Hotel-related words (e.g., The Ritz-Carlton, hotel, room) & Industry-issue-related words (such as brand, market)\\
Tourist-attraction-related words (e.g., Hong Kong Skyline) & Public-rights-related words (such as transparency, safety, privacy)\\
General Sanya- and Hong Kong-related words (e.g., hong\_kong, government) & Societal-issue-related words (such as family, woman)\\
Shopping-market-related words (e.g., Stanley Market) & Sanya and Hong Kong political and regulation-related words (such as political, community, authority, policy)\\
Celebrity-related words (e.g., Wong, Cheung) & International event-related words (such as event, exhibition)\\
\bottomrule
\end{tabular}
\end{table}

\subsection*{Text preprocessing}

It is imperative to perform data cleaning before analyzing a text corpus to enhance the outcomes of text mining. Prior to conducting data mining, the research conducted a sequence of text preprocessing procedures utilizing the Python NLTK library. These procedures encompassed tokenization, elimination of non-alphanumeric characters and conversion of all text to lowercase. Part-of-speech tagging was employed to select nouns, verbs, adjectives and adverbs, as well as to recognize bigram and trigram words, and lemmatization was applied to all terms. The study incorporated custom stop words for each dataset, alongside the commonly used NLTK stop words, which are words that occur frequently but do not hold significant meaning. Examples of such custom stop words include ``also'', ``could'' and ``get''. Bigrams and trigrams, which consist of a sequence of two or three tokens, were employed to capture commonly occurring multiword expressions.
\FloatBarrier
\section{Opinion-leader module, persuasion games and synergy calculation}\label{app:games}

\subsection*{Criteria used to select opinion leaders}

The identification of an opinion leader can be achieved through both qualitative and quantitative methods, which take into account the individual's area of expertise and the specific demand for their opinions. The criteria for selecting opinion leaders may differ across various approaches. Numerous scholars have conducted analyses and investigations on diverse methodologies utilizing a range of parameters to categorize opinion leaders. Their study presents a compilation of significant papers pertaining to the subject matter, highlighting their respective advantages and drawbacks. Diverse scholars employed a range of parameters and centrality measures to select the opinion leaders' roster. In the realm of opinion leadership, the process of selecting individuals to serve as influential figures is typically categorized into two distinct groups. These groups are differentiated by the inherent characteristics of the network in question and the behavior exhibited by users within said network. The initial classification is contingent upon the network's configuration, size, compactness and nodal degree. Conversely, the second approach is predicated upon the actions exhibited by the user, including but not limited to retweets, response time, trustworthiness and text mining. In addition, it is worth noting that various scholars hold distinct perspectives and employ diverse approaches in identifying opinion leaders within a network. According to \citet{walter2020opportunity}, diverse circumstances and distinct opportunities in human existence contribute to the emergence of opinion leaders. The proposition posits that if a user consistently obtains primary information from reputable sources, there is no necessity for them to depend on alternative social media platforms. The information that has been collected is adequate to establish them as opinion leaders. \citet{wang2020travelfrog} established the functions and investigated the impacts of opinion leaders on the dissemination of popular mobile games. The researchers conducted an experiment to determine the impact of opinion leaders on the diffusion rate of game-related topics. The results indicated that when an opinion leader promotes such topics, the diffusion rate is significantly higher than when promoted through user-generated content, which supports giving the content of opinion leaders a larger weight. In summary, each theory has presented a distinct approach to identifying opinion leaders, and each captures a different facet of opinion leadership.

\subsection*{Computation of the trust scores}

Algorithm~\ref{alg:trust} summarizes how the trust layer of Fig.~\ref{fig:olflow} computes the metrics of Eqs.~\eqref{eq:dt}--\eqref{eq:rt}, the trust scores of Eq.~\eqref{eq:ts} and the trust graph $G_T$.

\begin{algorithm}[!ht]
\caption{Trust scores (trust layer of Fig.~\ref{fig:olflow})}\label{alg:trust}
\begin{algorithmic}[1]
\Require interaction network $G=(V,E)$ with counts $n_{ij}$, recommendation data $b_i$ and $\chi_{ji}$, weight $\alpha\in(0,1)$, thresholds $\theta_{\mathrm{D}},\theta_{\mathrm{I}},\theta_{\mathrm{R}}\in[0.5,1)$, iteration limit $T_{\max}$, tolerance $\varepsilon_{\mathrm{tol}}>0$
\Ensure trust scores $\mathrm{TS}_i$, trusted set $V_T$ and trust graph $G_T$
\State compute $\bar{n}_{ij}$, $N(i)$, $d_i$, $d_i^{+}$ and $\mathrm{CL}_i$ for all users
\For{$i\in V$}
  \State $\mathrm{DT}_i\gets\sum_{j\in N(i)}\bar{n}_{ij}$ \Comment{Eq.~\eqref{eq:dt}}
  \State $\mathrm{IDT}_i\gets\sum_{j\in N(i)}\big[\alpha\bar{n}_{ij}+(1-\alpha)\sum_{k\in N(j)\setminus\{i\}}\mathrm{CL}_k\big]$ \Comment{Eq.~\eqref{eq:idt}}
\EndFor
\State $\mathbf{R}\gets\mathbf{0}$, \; $\ell\gets0$
\Repeat \Comment{Eq.~\eqref{eq:rt}, Proposition~\ref{prop:rt}}
  \State $\mathbf{R}^{\mathrm{new}}\gets\alpha\mathbf{b}+(1-\alpha)\mathbf{P}^{\top}\mathbf{R}$, \; $\Delta\gets\|\mathbf{R}^{\mathrm{new}}-\mathbf{R}\|_1$, \; $\mathbf{R}\gets\mathbf{R}^{\mathrm{new}}$, \; $\ell\gets\ell+1$
\Until{$\Delta\le\varepsilon_{\mathrm{tol}}$ \textbf{or} $\ell=T_{\max}$}
\State rescale $\mathrm{DT}$, $\mathrm{IDT}$ and $\mathrm{RT}$ (the entries of $\mathbf{R}$) to $[0,1]$ and compute $\mathrm{TS}_i$ by Eq.~\eqref{eq:ts}
\State $V_T\gets\{i\mid\mathrm{TS}_i>0\}$, \; $G_T\gets$ undirected subgraph of $G$ induced by $V_T$
\end{algorithmic}
\end{algorithm}

\subsection*{The four persuasion games}

Throughout this appendix, $A$ and $B$ are two linked users, each holding an original plan, and the strategies are keeping the original plan ($\mathsf{o}$), adopting the other's plan, i.e., the altered plan ($\mathsf{a}$), and, from Solution~2 on, an intermediate agreement ($\mathsf{g}$). Rows of every payoff matrix correspond to the strategies of $A$ and columns to those of $B$, in the order $\mathsf{o},\mathsf{a},\mathsf{g}$. The parameters $\pi_{\mathsf{o}}$, $\pi_{\mathsf{a}}$ and $\pi_{\mathsf{g}}$ are positive.

\textbf{Solution 1.} In this scenario, the two players engage in persuasive communication with the aim of convincing the other to adopt their proposed plan or adjust their strategy in alignment with the other player's plan. The acceptance of a player's strategy by another player is not a straightforward matter. As a result, a fixed payoff is linked to each choice. Let $g(\mathsf{o})=\pi_{\mathsf{o}}$ and $g(\mathsf{a})=-\pi_{\mathsf{a}}$, so that a player gains $\pi_{\mathsf{o}}$ when it maintains its original plan and loses $\pi_{\mathsf{a}}$ when it alters it, and the choice of the opponent affects the player in the opposite way. The payoffs are $M_A(s_A,s_B)=g(s_A)-g(s_B)$ and $M_B=-M_A$, so that
\begin{equation}\label{eq:C1sol1}
M_A=\begin{bmatrix}0 & \pi_{\mathsf{o}}+\pi_{\mathsf{a}}\\ -\pi_{\mathsf{a}}-\pi_{\mathsf{o}} & 0\end{bmatrix},\qquad
M_B=\begin{bmatrix}0 & -\pi_{\mathsf{o}}-\pi_{\mathsf{a}}\\ \pi_{\mathsf{a}}+\pi_{\mathsf{o}} & 0\end{bmatrix}.
\end{equation}
The aforementioned scenario demonstrates that in the event that player A maintains its strategy, player B is presented with two alternatives, namely to modify its plan or adhere to it. In both scenarios, the resulting payoff would be $-\pi_{\mathsf{o}}-\pi_{\mathsf{a}}$ and 0, correspondingly. Opting for 0 is the optimal choice as opposed to selecting $-\pi_{\mathsf{o}}-\pi_{\mathsf{a}}$, so B does not alter its plan either. Based on the preceding discourse, it can be inferred that neither player would receive a payoff, resulting in a failure to reach a mutually beneficial agreement. Consequently, an alternative resolution has been presented whereby both parties reach a mutually acceptable compromise.

\textbf{Solution 2.} The proposed solution involves the introduction of an intermediate strategy known as ``agreement'', wherein both players agree to a compromise without completely altering their existing strategies. Moving to the agreement costs a player the concession $\pi_{\mathsf{g}}$, i.e., $g(\mathsf{g})=-\pi_{\mathsf{g}}$, and again $M_A(s_A,s_B)=g(s_A)-g(s_B)$ and $M_B=-M_A$,
\begin{equation}\label{eq:C2sol2}
M_A=\begin{bmatrix}0 & \pi_{\mathsf{o}}+\pi_{\mathsf{a}} & \pi_{\mathsf{o}}+\pi_{\mathsf{g}}\\ -\pi_{\mathsf{a}}-\pi_{\mathsf{o}} & 0 & \pi_{\mathsf{g}}-\pi_{\mathsf{a}}\\ -\pi_{\mathsf{g}}-\pi_{\mathsf{o}} & \pi_{\mathsf{a}}-\pi_{\mathsf{g}} & 0\end{bmatrix},\qquad M_B=-M_A .
\end{equation}
In this instance, it was observed that in the event that a player maintains its initial plan, the opposing player has three options available to them. These options include altering their own plan, maintaining their original strategy or reaching a mutually agreed upon compromise. Across all three alternatives, the outcome for the opposing player would be $-\pi_{\mathsf{o}}-\pi_{\mathsf{a}}$, 0 and $-\pi_{\mathsf{o}}-\pi_{\mathsf{g}}$, respectively. Therefore, it can be inferred that the optimal course of action for the player is to select a payoff of zero and maintain their current strategy. The following result makes the deadlock of Solutions~1 and 2 precise.

\begin{proposition}[Deadlock in Solutions 1 and 2]\label{prop:sol12}
In Solutions~1 and 2, keeping the original plan strictly dominates every other strategy for both players. Hence $(\mathsf{o},\mathsf{o})$ is the unique Nash equilibrium, in pure or mixed strategies, and both players receive 0. Both games are symmetric zero-sum games with value 0.
\end{proposition}

It was demonstrated that the optimal course of action for both players is to maintain their current strategy, resulting in no payoff. Therefore, it can be inferred that neither player in this scenario is able to reach a satisfactory outcome, indicating a need for an improved strategy to achieve the maximum possible reward. Therefore, an additional optimal solution is presented, utilizing centrality in conjunction with the intermediate solution.

\textbf{Solution 3.} The present approach incorporates a novel parameter denoted as ``distance'', represented by $d$, that denotes the average centrality among the users. Centrality is a significant factor in determining the significance of a node within a social network. The degree of centrality exhibited by a user is contingent upon the underlying dynamics and structure of the network. The temporal evolution of network dynamics results in fluctuations in the centrality of individual users. The degree to which a user is central within a network significantly impacts their accessibility and potential to influence the behavior and knowledge of other users.

The present study employs the Closeness Centrality (CC), Betweenness Centrality (BC), Degree Centrality (DC) \citep{freeman1978centrality} and the local clustering coefficient $\mathrm{CL}_i$ as metrics to establish the distance $d_{ij}$ between users $i$ and $j$ within the network, as defined by the following equation,
\begin{equation}\label{eq:dist}
d_{ij}=\epsilon_d+\tfrac{1}{2}\big(\zeta_i+\zeta_j\big)+\lambda\max\{\mathrm{CL}_i,\mathrm{CL}_j\}+\rho\min\{\mathrm{CL}_i,\mathrm{CL}_j\},\qquad \zeta_u=\sqrt{\frac{\mathrm{BC}_u\,\mathrm{CC}_u}{\mathrm{DC}_u}},
\end{equation}
where $\zeta_u$ is a composite centrality computed from normalized measures and $\epsilon_d>0$ is a small constant that keeps $1/d_{ij}$ finite when both users have zero betweenness and zero clustering. The distance therefore combines the average composite centrality of the two users with their clustering coefficients. The equation incorporates two weighted coefficients, namely $\lambda$ and $\rho$, which serve the purpose of balancing the distance metric $d$. The variables $\lambda$ and $\rho$ are constrained to the interval $[0,0.5]$. The coefficient $\lambda$ weights the larger and $\rho$ the smaller of the two clustering coefficients, so that $d_{ij}=d_{ji}$ and the outcome of a game does not depend on which user is labeled $A$. It was also observed that a gradual increase in the clustering coefficient of a user results in a corresponding increase in the distance between users. Each player that moves to the agreement now adds the closeness bonus $1/d$ to the payoffs of both players, where $d=d_{ij}$ for the two users $A=i$ and $B=j$, so that
\begin{equation}\label{eq:C3sol3}
\begin{gathered}
M_A=\begin{bmatrix}0 & \pi_{\mathsf{o}}+\pi_{\mathsf{a}} & \pi_{\mathsf{o}}+\pi_{\mathsf{g}}+\frac{1}{d}\\[3pt] -\pi_{\mathsf{a}}-\pi_{\mathsf{o}} & 0 & \pi_{\mathsf{g}}-\pi_{\mathsf{a}}+\frac{1}{d}\\[3pt] -\pi_{\mathsf{g}}-\pi_{\mathsf{o}}+\frac{1}{d} & \pi_{\mathsf{a}}-\pi_{\mathsf{g}}+\frac{1}{d} & \frac{2}{d}\end{bmatrix},\\[6pt]
M_B=\begin{bmatrix}0 & -\pi_{\mathsf{o}}-\pi_{\mathsf{a}} & -\pi_{\mathsf{o}}-\pi_{\mathsf{g}}+\frac{1}{d}\\[3pt] \pi_{\mathsf{a}}+\pi_{\mathsf{o}} & 0 & \pi_{\mathsf{a}}-\pi_{\mathsf{g}}+\frac{1}{d}\\[3pt] \pi_{\mathsf{g}}+\pi_{\mathsf{o}}+\frac{1}{d} & \pi_{\mathsf{g}}-\pi_{\mathsf{a}}+\frac{1}{d} & \frac{2}{d}\end{bmatrix}.
\end{gathered}
\end{equation}
In this instance, an analysis was conducted to determine that in the event of mutual agreement between the players regarding the intermediate solution, the payoff for player A is equivalent to $2/d$. Despite yielding superior outcomes relative to Solution~2, the aforementioned approach is still susceptible to a recurring issue whereby a player's adherence to their original strategy may result in another player having three options, namely $-\pi_{\mathsf{o}}-\pi_{\mathsf{a}}$, 0 and $-\pi_{\mathsf{o}}-\pi_{\mathsf{g}}+1/d$, to select from in terms of their payoff. Therefore, selecting payoff 0 would be the optimal choice for the other player whenever $\pi_{\mathsf{o}}+\pi_{\mathsf{g}}>1/d$, as Proposition~\ref{prop:sol3} shows. Once more, in order to address this issue, we propose a novel resolution incorporating the variable $d$.

\begin{proposition}[Threshold and social dilemma in Solution 3]\label{prop:sol3}
In Solution~3, altering is strictly dominated by keeping the original plan. If $1/d>\pi_{\mathsf{o}}+\pi_{\mathsf{g}}$, agreement strictly dominates for both players and $(\mathsf{g},\mathsf{g})$ is the unique Nash equilibrium, with payoffs $2/d$ each. If $1/d<\pi_{\mathsf{o}}+\pi_{\mathsf{g}}$, keeping the original plan strictly dominates and $(\mathsf{o},\mathsf{o})$ is the unique Nash equilibrium, with payoffs 0, although $(\mathsf{g},\mathsf{g})$ would give each player $2/d>0$. In the latter case the game is a prisoner's dilemma. If $1/d=\pi_{\mathsf{o}}+\pi_{\mathsf{g}}$, both players are indifferent between $\mathsf{o}$ and $\mathsf{g}$ against every strategy of the opponent, and every profile in which each player mixes over $\mathsf{o}$ and $\mathsf{g}$ is a Nash equilibrium.
\end{proposition}

\textbf{Solution 4.} Empirical evidence suggests that in practical settings, it is not always the case that a user will fully concur or dissent with the strategic choices of another player. There is a probability regarding how much a player can persuade another player with their proposed strategy. The present solution incorporates a novel parameter denoted as $p$ that connotes the level of incentivization in the game. The probabilities $p_A$ and $p_B$ are defined to represent the extent to which player A influences player B, and vice versa, and in the network they are the influence probabilities $p_A=p_{ij}$ and $p_B=p_{ji}$ of Eq.~\eqref{eq:bayes}. Two parameters determine the game's outcome, $p_A$ and $p_B$, which have an impact on the payoff of each player. In the event that player A lacks the ability to exert influence over player B, which happens with probability $1-p_A$, B keeps its plan. The payoff of A is the separable function of Eq.~\eqref{eq:sep}, which gives
\begin{equation}\label{eq:C4MA}
M_A=\mbox{\small\setlength{\arraycolsep}{3pt}$\begin{bmatrix}
\pi_{\mathsf{o}}(p_A-p_B) & \pi_{\mathsf{o}}(1-p_B)+\pi_{\mathsf{a}}p_A & \pi_{\mathsf{o}}(1-p_B)+\pi_{\mathsf{g}}+\frac{1}{d}\\[3pt]
-\pi_{\mathsf{a}}p_B-\pi_{\mathsf{o}}(1-p_A) & \pi_{\mathsf{a}}(p_A-p_B) & -\pi_{\mathsf{a}}p_B+\pi_{\mathsf{g}}+\frac{1}{d}\\[3pt]
-\pi_{\mathsf{o}}(1-p_A)-\pi_{\mathsf{g}}+\frac{1}{d} & \pi_{\mathsf{a}}p_A-\pi_{\mathsf{g}}+\frac{1}{d} & \frac{2}{d}
\end{bmatrix}$},
\end{equation}
and the payoff of B is obtained by exchanging the roles of the two players, including $p_A\leftrightarrow p_B$, i.e., $M_B(s_A,s_B)=M_A(s_B,s_A)\big|_{p_A\leftrightarrow p_B}$,
\begin{equation}\label{eq:C5MB}
M_B=\mbox{\small\setlength{\arraycolsep}{3pt}$\begin{bmatrix}
\pi_{\mathsf{o}}(p_B-p_A) & -\pi_{\mathsf{a}}p_A-\pi_{\mathsf{o}}(1-p_B) & -\pi_{\mathsf{o}}(1-p_B)-\pi_{\mathsf{g}}+\frac{1}{d}\\[3pt]
\pi_{\mathsf{o}}(1-p_A)+\pi_{\mathsf{a}}p_B & \pi_{\mathsf{a}}(p_B-p_A) & \pi_{\mathsf{a}}p_B-\pi_{\mathsf{g}}+\frac{1}{d}\\[3pt]
\pi_{\mathsf{o}}(1-p_A)+\pi_{\mathsf{g}}+\frac{1}{d} & -\pi_{\mathsf{a}}p_A+\pi_{\mathsf{g}}+\frac{1}{d} & \frac{2}{d}
\end{bmatrix}$}.
\end{equation}
Note that $M_B$ is not simply the transpose of $M_A$. Without the exchange of $p_A$ and $p_B$, both players would earn $\pi_{\mathsf{o}}(p_A-p_B)$ when neither moves, so the advantage of the more persuasive player would be shared by its opponent, which contradicts the meaning of $p_A$ and $p_B$. In this instance, it was determined that in the event that neither player alters its strategy, they may still receive a payoff contingent upon the values of $p_A$ and $p_B$, subject to the conditions $0<p_A<1$ and $0<p_B<1$. Likewise, in the event that both parties concur on the intermediary resolution, the resulting benefits remain equal for each of them. Consequently, the proposed solution serves as a source of motivation for players to engage in strategic planning and make informed decisions regarding their next move. Proposition~\ref{prop:sol4} shows that player A opts for the agreement regardless of the strategy of B exactly when $\pi_{\mathsf{o}}(1-p_B)<1/d-\pi_{\mathsf{g}}$, and player B does so exactly when $\pi_{\mathsf{o}}(1-p_A)<1/d-\pi_{\mathsf{g}}$.

Assuming discrete rounds $r=0,1,2,\dots$, it can be posited that both players make their decisions based on their prior progress. Regarding this matter, the temporal advancement required for one player to persuade another player can be expressed through the opinions $E_A(r)$ and $E_B(r)$ of the two players,
\begin{align}
E_A(r+1)&=E_A(r)+\mu\big(E_B(r)-E_A(r)\big), \label{eq:C6a}\\
E_B(r+1)&=E_B(r)+\eta\big(E_A(r)-E_B(r)\big), \label{eq:C6b}
\end{align}
where the two configurable variables $\mu$ and $\eta$ are constrained to the interval $(0,0.5)$. The sign in Eq.~\eqref{eq:C6b} matters. If the second equation used $E_B(r)-E_A(r)$ instead, player B would move away from A, the gap would evolve as $E_B(r+1)-E_A(r+1)=(1+\eta-\mu)\big(E_B(r)-E_A(r)\big)$ and it would diverge whenever $\eta>\mu$. Eqs.~\eqref{eq:C6a}--\eqref{eq:C6b} instead form a two-agent DeGroot process \citep{degroot1974consensus}.

\begin{proposition}[Convergence of the persuasion dynamics]\label{prop:dyn}
Let $\mu,\eta\in(0,0.5)$ and $\Delta E(r)=E_B(r)-E_A(r)$. Then $\Delta E(r)=(1-\mu-\eta)^{r}\Delta E(0)$, so the gap shrinks geometrically and never changes sign. The weighted mean $\eta E_A(r)+\mu E_B(r)$ is invariant, and both opinions converge to $E^{\ast}=\big(\eta E_A(0)+\mu E_B(0)\big)/(\mu+\eta)$ from opposite sides, so that they remain between $E_A(0)$ and $E_B(0)$ for every $r$. If $|\Delta E(0)|>\varepsilon_E>0$, the gap is at most $\varepsilon_E$ after $r_E=\big\lceil\ln(\varepsilon_E/|\Delta E(0)|)/\ln(1-\mu-\eta)\big\rceil$ rounds.
\end{proposition}

The number of rounds $r_E$ is the time one player needs to persuade the other, and the limit $E^{\ast}$ lies closer to the opinion of the player who adjusts more slowly, i.e., the more persuasive one. The upper bound $0.5$ on $\mu$ and $\eta$ rules out overshooting, because it guarantees $0<1-\mu-\eta<1$, so that neither player jumps past the other. Table~\ref{tab:games} summarizes the four solutions.

\begin{table}[!ht]
\centering
\caption{Equilibria of the four pairwise persuasion games.}\label{tab:games}
\small
\begin{tabularx}{\linewidth}{@{}l>{\raggedright\arraybackslash}X>{\raggedright\arraybackslash}X>{\raggedright\arraybackslash}X@{}}
\toprule
Solution & Payoff structure & Equilibrium & Result\\
\midrule
1 & zero-sum, strategies $\{\mathsf{o},\mathsf{a}\}$ & $(\mathsf{o},\mathsf{o})$, payoffs $(0,0)$ & deadlock (Prop.~\ref{prop:sol12})\\
2 & zero-sum, adds agreement $\mathsf{g}$ & $(\mathsf{o},\mathsf{o})$, payoffs $(0,0)$ & deadlock (Prop.~\ref{prop:sol12})\\
3 & zero-sum part plus common bonus $1/d$ per agreeing player & $(\mathsf{g},\mathsf{g})$ if $1/d>\pi_{\mathsf{o}}+\pi_{\mathsf{g}}$, $(\mathsf{o},\mathsf{o})$ if $1/d<\pi_{\mathsf{o}}+\pi_{\mathsf{g}}$, any mixture of $\mathsf{o}$ and $\mathsf{g}$ at equality & cooperation or prisoner's dilemma (Prop.~\ref{prop:sol3})\\
4 & separable, with influence probabilities $p_A,p_B$ & dominant strategies, mutual agreement iff Eq.~\eqref{eq:agree} holds & agreement graph $G^{\mathrm{ag}}$ (Prop.~\ref{prop:sol4})\\
\bottomrule
\end{tabularx}
\end{table}

\subsection*{Synergy calculation}

\begin{figure}[!ht]
\centering
\includegraphics[width=0.66\textwidth]{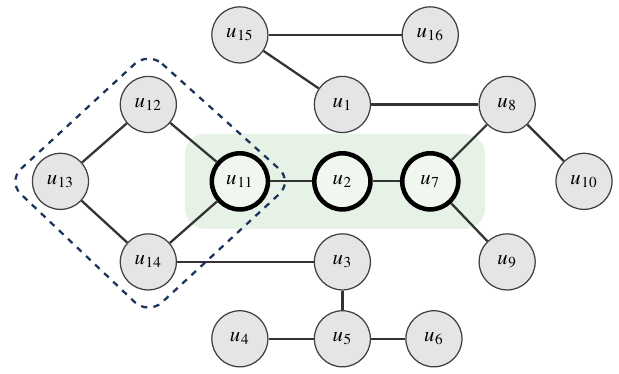}
\caption{The brokerage coalition of the $u_2$, $u_7$ and $u_{11}$ nodes (shaded) and the coalition of maximal synergy $\{u_{11},u_{12},u_{13},u_{14}\}$ for unit trust scores and $c=1$ (dashed outline).}\label{fig:netC1}
\end{figure}

The concept of synergy in a network refers to the collaborative capacity of specific nodes, which, when combined, generate more significant value than the sum of their individual contributions. The coalition game involves players who possess individual resources that hold significant value. Upon forming an alliance, the players may generate a collective synergy that surpasses or falls short of the total value of their resources at a given time. Consider a hypothetical scenario in a game where the individual payoffs of two players, denoted as $i$ and $j$, are 3 and 2, respectively. In the event that both players collaborate during gameplay, the resultant collective value may be six, as determined by a multiplicative factor. This value surpasses the sum of their individual payoffs \citep{hernandez2019synergy,jain2020coalitions}, and the excess of the joint value over that sum, here $6-(3+2)=1$, is the synergy of the pair. As a second illustration, let us contemplate a network comprising 16 nodes, as depicted in Fig.~\ref{fig:netC1}. Table~\ref{tab:toy} reports exact centralities for this network. The nodes with the highest degree are $u_5$, $u_7$, $u_8$, $u_{11}$ and $u_{14}$, each with three neighbors. Node $u_7$ has the largest betweenness centrality (59), followed closely by $u_{11}$ (58.5) and $u_2$ (56), whereas $u_2$ and $u_{11}$ share the largest closeness centrality (1/3) and $u_{11}$ has the largest eigenvector centrality. However, it is not a suitable methodology to evaluate the significance of a node solely based on a singular attribute, and the measures in Table~\ref{tab:toy} indeed rank the nodes differently. This is because the dissemination of information within the network is not solely reliant on one or two nodes but rather a group of nodes with similar or dissimilar attributes. The nodes $u_2$, $u_7$ and $u_{11}$ form a brokerage coalition, which exhibited the ability to either facilitate or impede the dissemination of information, contingent upon the prevailing circumstances. The three nodes with the largest betweenness are cut vertices that together form a connected spine. Of the 78 pairs among the remaining 13 nodes, the spine carries 49.5 in terms of shortest-path flow, i.e., its group betweenness \citep{everett1999groups} equals 63.5\% of the maximum, so it controls most of the traffic between the left and right parts of the network. Synergy in the sense of Definition~\ref{def:synergy} rewards a different property, namely cohesion. With unit trust scores, $\kappa=1$, $\vartheta=\delta=c=1$ and every link in the agreement graph, Algorithm~\ref{alg:coalition} returns the four-cycle $\{u_{11},u_{12},u_{13},u_{14}\}$, outlined in Fig.~\ref{fig:netC1}, with $\Phi=0.423$, whereas the brokerage coalition attains $\Phi=0.230$. Lowering $c$ to $0.5$ penalizes size less and enlarges the optimal coalition to 15 of the 16 users. The two notions are complementary, since brokers control the flow of information between groups and cohesive coalitions amplify it within them. In the event that erroneous data are disseminated by a node, a group of such nodes has the potential to disrupt the spread of that information.

\begin{table}[!ht]
\centering
\caption{Exact centralities of the 16-node network of Fig.~\ref{fig:netC1}. Deg.\ is the degree, BC the unnormalized betweenness, CC the closeness, EC the eigenvector centrality (unit Euclidean norm) and $\varphi_i$ the Shapley value of the game of Eq.~\eqref{eq:game} with unit trust scores and $\kappa=1$ (Corollary~\ref{cor:shapley}(ii)).}\label{tab:toy}
\small
\setlength{\tabcolsep}{4pt}
\begin{tabular}{@{}l rrrrr@{\hspace{16pt}}l rrrrr@{}}
\toprule
Node & Deg. & BC & CC & EC & $\varphi_i$ & Node & Deg. & BC & CC & EC & $\varphi_i$\\
\cmidrule(r{8pt}){1-6}\cmidrule(l){7-12}
$u_{1}$ & 2 & 26 & .238 & .098 & .917 & $u_{9}$ & 1 & 0 & .246 & .107 & .750\\
$u_{2}$ & 2 & 56 & .333 & .306 & .833 & $u_{10}$ & 1 & 0 & .224 & .077 & .750\\
$u_{3}$ & 2 & 36 & .263 & .270 & .833 & $u_{11}$ & 3 & 58.5 & .333 & .470 & 1.167\\
$u_{4}$ & 1 & 0 & .185 & .076 & .750 & $u_{12}$ & 2 & 4.5 & .263 & .343 & .917\\
$u_{5}$ & 3 & 27 & .224 & .179 & 1.583 & $u_{13}$ & 2 & 2.5 & .246 & .340 & .917\\
$u_{6}$ & 1 & 0 & .185 & .076 & .750 & $u_{14}$ & 3 & 48.5 & .306 & .458 & 1.167\\
$u_{7}$ & 3 & 59 & .319 & .252 & 1.333 & $u_{15}$ & 2 & 14 & .200 & .051 & 1.167\\
$u_{8}$ & 3 & 47 & .283 & .181 & 1.333 & $u_{16}$ & 1 & 0 & .169 & .022 & .833\\
\bottomrule
\end{tabular}
\end{table}

If all the different parts of a machine operate in a coordinated manner with mutual reasoning, it is possible for one machine to exhibit higher productivity compared to another. Similarly, within the realm of social media, a smaller, more cohesive coalition may yield more effective and inventive results in contrast to a larger, less organized group, which is exactly the behavior that Proposition~\ref{prop:admission} establishes for the synergy of Eq.~\eqref{eq:Phi}. The primary concept of synergy between two networks is founded on the existence of structural holes within a network \citep{burt2004holes}. A structural hole is formed when a solitary node establishes connections with an extensive network and simultaneously links with another node that is associated with a distinct extensive network. These two nodes serve as the sole means of transmitting information between groups. The aforementioned nodes have the responsibility of disseminating knowledge and information \citep{burt2004holes}. These nodes significantly influence network operations due to their potential to impact multiple links concurrently. Such brokers create synergy between groups, whereas Definition~\ref{def:synergy} measures the cohesion within a coalition, which is the complementarity illustrated in Fig.~\ref{fig:netC1}.

\begin{figure}[!ht]
\centering
\includegraphics[width=\textwidth]{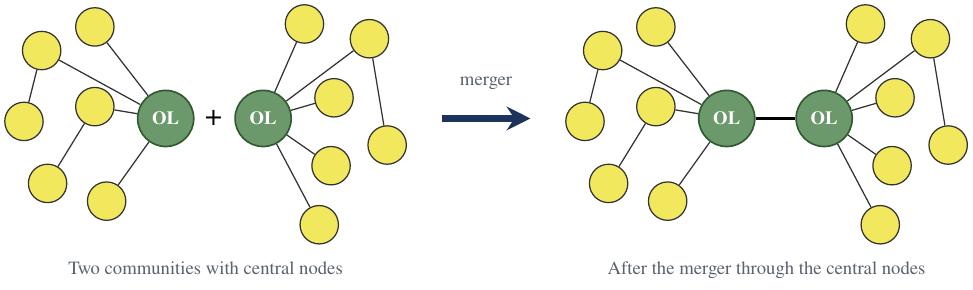}
\caption{A sample of two networks' synergetic coalition, formed by linking their central nodes (OL).}\label{fig:netC2}
\end{figure}

Regarding Fig.~\ref{fig:netC2}, it is worth noting that two networks exist wherein a node with a greater centrality degree (the green node labeled OL) is present, and most of the remaining nodes are linked to that node. In the event that a node within the network desires to communicate information with another network, it is imperative that all data are routed through the green node. As such, this particular node assumes a pivotal role in facilitating the synergistic collaboration between the two networks. The integration of two or more nodes is probable when the resultant merging outcome significantly exceeds the anticipated outcomes. Eigenvector centrality (EC) is utilized as the primary component for evaluating the synergy of the merger. The source node seeks to merge preferably with target nodes that possess a higher Shapley value, because $\Omega_{ij}$ increases with $\varphi_j$. The measurement of the synergy $\Omega_{ij}$ resulting from the merger of two nodes and the overall synergy $\Phi$ of a coalition are given in Definition~\ref{def:synergy}. The user's behavior is conditioned by the specific factor denoted as $c$. Additionally, the coordination rates that characterize the level of coordination among the users are represented by the symbols $\delta$ and $\vartheta$. The variables $c$, $\delta$ and $\vartheta$ take on values in $(0,1]$, since a zero value would make the synergy vanish or, for $c$, would count every pair with equal weight. Algorithm~\ref{alg:coalition} computes the coalition of maximal synergy as stated in Theorem~\ref{thm:coalition}.

\begin{algorithm}[!ht]
\caption{Exact coalition selection by Dinkelbach's method}\label{alg:coalition}
\begin{algorithmic}[1]
\Require trusted users $V_T$, pairwise synergies $\Omega_{ij}\ge0$, exponent $c\in(0,1]$, coordination rate $\delta\in(0,1]$
\Ensure inclusion-maximal coalition $C^{\ast}$ of maximal synergy $\Phi(C^{\ast})=\delta q^{\ast}$
\State $C\gets V_T$, \; $q\gets S(C)/|C|^{c}$ \Comment{$S(C)=\sum_{i<j\in C}\Omega_{ij}^{c}$}
\Loop
  \State $C'\gets$ inclusion-maximal maximizer of $F_q(D)=S(D)-q|D|^{c}$ over $D\subseteq V_T$ \Comment{minimize $-F_q$}
  \If{$F_q(C')=0$}
     \State \Return $C^{\ast}\gets C'$ and $q^{\ast}\gets q$ \Comment{$C'\supseteq C$ is optimal, Theorem~\ref{thm:coalition}(iii)}
  \EndIf
  \State $C\gets C'$, \; $q\gets S(C)/|C|^{c}$ \Comment{$|C|$ strictly decreases}
\EndLoop
\end{algorithmic}
\end{algorithm}

\subsection*{Robustness of the selected coalition}

The coalition $C^{\ast}$ is computed from estimated quantities. The influence probabilities are estimated from endorsement counts (Proposition~\ref{prop:hoeffding}) and decide through Eq.~\eqref{eq:agree} which links enter the agreement graph, while the synergies $\Omega_{ij}$ inherit errors from the trust scores and the centralities, for example when the observation window changes. Let $\tilde{\Omega}_{ij}\ge0$ denote estimated synergies, $\tilde{\Phi}$ the synergy of Eq.~\eqref{eq:Phi} computed from them and $\Delta^{\mathrm{ag}}$ the maximum degree of the agreement graph. The next result bounds the effect of both kinds of error.

\begin{proposition}[Stability of the coalition of opinion leaders]\label{prop:stab}
\textup{(i)} Suppose that $\big|\pi_{\mathsf{o}}\big(1-\min\{p_{ij},p_{ji}\}\big)-1/d_{ij}+\pi_{\mathsf{g}}\big|>\pi_{\mathsf{o}}\varepsilon$ for every $\{i,j\}\in E_T$. Then, on the event of Proposition~\ref{prop:hoeffding} that $|\hat{p}_{ij}-p_{ij}|<\varepsilon$ for all ordered pairs of linked users, the graph of the links $\{i,j\}\in E_T$ that satisfy Eq.~\eqref{eq:agree} with $\hat{p}_{ij}$ and $\hat{p}_{ji}$ in place of $p_{ij}$ and $p_{ji}$ equals $G^{\mathrm{ag}}$. \textup{(ii)} Suppose that the agreement graph is fixed and that $|\tilde{\Omega}_{ij}-\Omega_{ij}|\le\varepsilon_\Omega$ on its edges, with $\tilde{\Omega}_{ij}=\Omega_{ij}=0$ elsewhere. Then $|\tilde{\Phi}(C)-\Phi(C)|\le\tfrac{\delta}{2}\Delta^{\mathrm{ag}}|C|^{1-c}\varepsilon_\Omega^{c}$ for every nonempty $C\subseteq V_T$. \textup{(iii)} Let $\varepsilon_\Phi=\tfrac{\delta}{2}\Delta^{\mathrm{ag}}n_T^{1-c}\varepsilon_\Omega^{c}$. Under the assumptions of (ii), every maximizer $\tilde{C}$ of $\tilde{\Phi}$ satisfies $\Phi(\tilde{C})\ge\Phi(C^{\ast})-2\varepsilon_\Phi$, and if $\Phi(C^{\ast})-\Phi(C)>2\varepsilon_\Phi$ for every nonempty $C\ne C^{\ast}$, then $C^{\ast}$ is the unique maximizer of $\tilde{\Phi}$.
\end{proposition}

Part (i) turns the sample-size condition of Proposition~\ref{prop:hoeffding} into a guarantee for the agreement graph, provided that no linked pair lies too close to the agreement threshold of Eq.~\eqref{eq:agree}. Parts (ii) and (iii) show that small errors in the synergies can only replace the coalition by one that is almost as good, and that a coalition whose optimality margin exceeds $2\varepsilon_\Phi$ is selected exactly. For $c=1$ the bound $\varepsilon_\Phi=\tfrac{\delta}{2}\Delta^{\mathrm{ag}}\varepsilon_\Omega$ does not depend on the number of users $n_T$. Changes in the topology of the agreement graph are covered by part (i), not by part (ii).
\FloatBarrier
\section{Stacked LSTM and dropout architecture}\label{app:lstm}

Stacked LSTM and dropout architecture are explained in this appendix, which also counts the trainable weights of the network.

\subsection*{Stacked LSTM}

\begin{figure}[!ht]
\centering
\includegraphics[width=\textwidth]{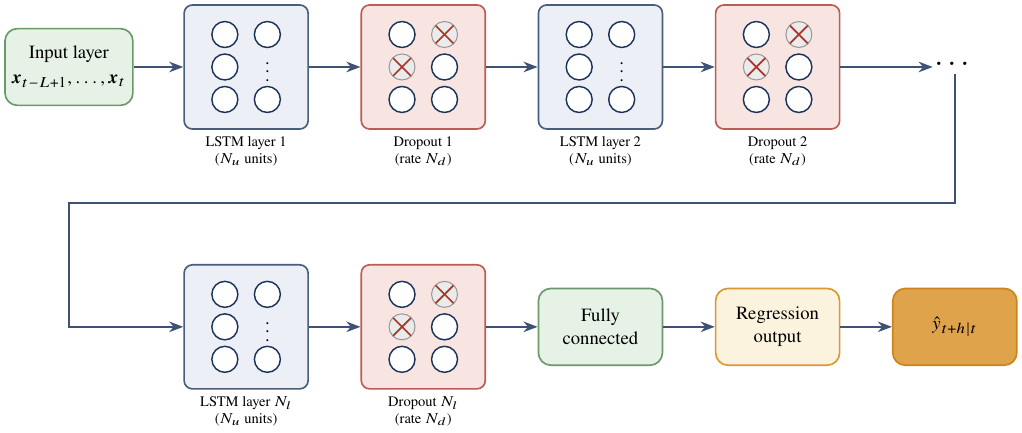}
\caption{The architecture of the stacked LSTM with dropout layers.}\label{fig:stacked}
\end{figure}

Customized neural networks with greater depth are frequently employed to address complex prediction tasks and to enhance accuracy. Employing a stacked LSTM architecture makes it possible to depict increasingly intricate input patterns at each layer \citep{graves2013speech}. The proposed approach adheres to a fundamental stacked LSTM architecture, as depicted in Fig.~\ref{fig:stacked}. The present investigation involves the utilization of a stacked LSTM model, wherein multiple LSTM layers are employed in conjunction with dropout layers and an output regression layer at the last output. The output sequences generated by the lower LSTM layers are utilized as inputs for the subsequent LSTM layer. The stacked LSTM model can enhance the non-linear transformation of raw data and expedite convergence because its parameters are distributed over several layers. Consequently, the present investigation incorporates a stacked LSTM architecture within our proposed model (stacked BiLSTM).

\subsection*{Number of trainable weights}

An LSTM cell \citep{gers2000forget,hochreiter1997lstm} has an input, a forget and an output gate and a candidate update of its memory cell, each with an input weight matrix, a recurrent weight matrix and a bias. A bidirectional layer with $N_u$ units that reads inputs of dimension $N_x$ therefore has $8N_u(N_x+N_u+1)$ trainable weights, and every further layer of the stack, whose inputs are the $2N_u$ outputs of the layer below, has $8N_u(3N_u+1)$. With the smallest layer of Table~\ref{tab:hyper}, $N_u=60$, a single input series already gives 29,760 weights in the first layer and 86,880 in each further layer, which is why dropout, $L_2$ regularization and validation-based tuning matter for training samples as short as ours.

\subsection*{Dropout architecture}

The regularization technique known as dropout reduces overfitting in feedforward and recurrent networks alike \citep{gal2016dropout,srivastava2014dropout}, and the reduction of overfitting leads to increased accuracy in the model's outcomes. The rate $N_d$ is a hyperparameter of BO with range $[0.1,0.9]$ (Table~\ref{tab:hyper}), and a dropout rate of 0.1 was employed to mitigate the risk of overfitting. Consequently, the present study added a dropout layer and compared its performance to that of conventional LSTM models lacking such a layer. In these experiments, dropout LSTMs performed better than conventional LSTMs, and the dropout architecture has therefore been integrated into this study.

\FloatBarrier
\section{Proofs of the formal results}\label{app:proofs}

\begin{proof}[Proof of Proposition~\ref{prop:rt}]
In vector form, Eq.~\eqref{eq:rt} reads $\mathbf{R}=\Gamma(\mathbf{R})$ with $\Gamma(\mathbf{R})=\alpha\mathbf{b}+(1-\alpha)\mathbf{P}^{\top}\mathbf{R}$. The induced $\ell_1$ norm of $\mathbf{P}^{\top}$ is its largest absolute column sum, $\max_{j}\sum_{i}|P_{ji}|=\max_{j}\sum_{i:(j,i)\in E}|\chi_{ji}|/d_j^{+}\le1$, because $|\chi_{ji}|\le1$ and $j$ has exactly $d_j^{+}$ out-neighbors, while a user without out-neighbors contributes a zero column. Hence $\|\Gamma(\mathbf{R})-\Gamma(\mathbf{R}')\|_1\le(1-\alpha)\|\mathbf{R}-\mathbf{R}'\|_1$, and $\Gamma$ is a contraction of $(\mathbb{R}^{n},\|\cdot\|_1)$ with modulus $1-\alpha<1$. By the Banach fixed-point theorem, $\Gamma$ has a unique fixed point $\mathbf{R}^{\ast}$ and $\|\mathbf{R}^{(\ell)}-\mathbf{R}^{\ast}\|_1\le(1-\alpha)^{\ell}\|\mathbf{R}^{(0)}-\mathbf{R}^{\ast}\|_1$. Because the spectral radius of $(1-\alpha)\mathbf{P}^{\top}$ is at most $1-\alpha<1$, the matrix $\mathbf{I}-(1-\alpha)\mathbf{P}^{\top}$ is invertible with Neumann series $\sum_{\ell\ge0}\big((1-\alpha)\mathbf{P}^{\top}\big)^{\ell}$, so $\mathbf{R}^{\ast}=\alpha\big(\mathbf{I}-(1-\alpha)\mathbf{P}^{\top}\big)^{-1}\mathbf{b}$ and $\|\mathbf{R}^{\ast}\|_1\le\alpha\|\mathbf{b}\|_1\sum_{\ell\ge0}(1-\alpha)^{\ell}=\|\mathbf{b}\|_1$. With $\mathbf{R}^{(0)}=\mathbf{0}$ the error after $T_{\max}$ iterations is therefore at most $(1-\alpha)^{T_{\max}}\|\mathbf{b}\|_1$.
\end{proof}

\begin{proof}[Proof of Proposition~\ref{prop:hoeffding}]
Conditionally on $m_i$, $\hat{p}_{ij}=m_i^{-1}\sum_{e=1}^{m_i}J_e$, where $J_e$ indicates that $j$ endorsed the $e$th item after $i$, is the mean of $m_i$ independent $\{0,1\}$-valued variables with mean $p_{ij}$. Hoeffding's inequality \citep{hoeffding1963} gives $\Pr(|\hat{p}_{ij}-p_{ij}|\ge\varepsilon\mid m_i)\le2\exp(-2m_i\varepsilon^{2})$. If $m_i\ge m_{\min}$ for every trusted user with a neighbor in $G_T$, the union bound over the $2|E_T|$ ordered pairs of linked users gives $\Pr\big(\max_{(i,j)}|\hat{p}_{ij}-p_{ij}|\ge\varepsilon\big)\le4|E_T|\exp(-2m_{\min}\varepsilon^{2})$, which is at most $\nu$ when $m_{\min}\ge\ln(4|E_T|/\nu)/(2\varepsilon^{2})$.
\end{proof}

\begin{proof}[Proof of Theorem~\ref{thm:shapley}]
For $u\in V_T$ let $v_u(C)=\1\{u\in C\ \text{or}\ |N_T(u)\cap C|\ge\kappa\}$, so that $v=\sum_u\mathrm{TS}_u v_u$. The Shapley value is linear in the game \citep{shapley1953value}, hence $\varphi_i(v)=\sum_u\mathrm{TS}_u\,\varphi_i(v_u)$. Recall that $\varphi_i(v_u)=\mathbb{E}_\sigma\big[v_u(\mathrm{Pre}_\sigma(i)\cup\{i\})-v_u(\mathrm{Pre}_\sigma(i))\big]$, where $\sigma$ is a uniformly random order of $V_T$ and $\mathrm{Pre}_\sigma(i)$ is the set of users preceding $i$ in $\sigma$. Fix $u$, write $N[u]=\{u\}\cup N_T(u)$ and $d_u=|N_T(u)|$. The game $v_u$ depends only on membership in $N[u]$, so users outside $N[u]$ are null players and $\varphi_i(v_u)=0$ for $i\notin N[u]$. For $i\in N[u]$ only the relative order of the $d_u+1$ members of $N[u]$ matters, and it is uniform over the $(d_u+1)!$ orders. Since $v_u$ is monotone and $\{0,1\}$-valued, the marginal contribution of $i$ is $1$ exactly when $i$ switches $v_u$ from $0$ to $1$.

\emph{Case $i=u$.} Adding $u$ always yields $v_u=1$, and $v_u(\mathrm{Pre}_\sigma(u))=0$ exactly when fewer than $\kappa$ neighbors of $u$ precede $u$. The number of neighbors preceding $u$ is uniform on $\{0,1,\dots,d_u\}$, because the position of $u$ among $N[u]$ is uniform. Hence $\varphi_u(v_u)=\min\{\kappa,d_u+1\}/(d_u+1)=\min\{1,\kappa/(1+d_u)\}$.

\emph{Case $i\in N_T(u)$.} User $i$ is pivotal exactly when $u$ has not yet arrived and exactly $\kappa-1$ of the other $d_u-1$ neighbors precede $i$. If $\kappa>d_u$ this is impossible and $\varphi_i(v_u)=0$. Otherwise, the favorable relative orders place $i$ in position $\kappa$ among $N[u]$, fill the first $\kappa-1$ positions with neighbors other than $i$, which can be done in $\binom{d_u-1}{\kappa-1}(\kappa-1)!$ ways, and arrange the remaining $d_u-\kappa+1$ members, $u$ included, after $i$ in $(d_u-\kappa+1)!$ ways. Therefore
\[
\varphi_i(v_u)=\frac{\binom{d_u-1}{\kappa-1}(\kappa-1)!\,(d_u-\kappa+1)!}{(d_u+1)!}=\frac{(d_u-1)!\,(d_u-\kappa+1)}{(d_u+1)!}=\frac{d_u-\kappa+1}{d_u(d_u+1)}.
\]
Both subcases together give $\varphi_i(v_u)=(d_u-\kappa+1)^{+}/\big(d_u(d_u+1)\big)$ for $u\in N_T(i)$, where we used that $i\in N_T(u)$ if and only if $u\in N_T(i)$. Summing $\mathrm{TS}_u\varphi_i(v_u)$ over $u\in\{i\}\cup N_T(i)$ yields Eq.~\eqref{eq:shapleyclosed}. Given the degrees, the sum for user $i$ has $1+d_i$ terms, so all values cost $O(n_T+\sum_i d_i)=O(n_T+|E_T|)$ operations.
\end{proof}

\begin{proof}[Proof of Corollary~\ref{cor:shapley}]
(i) For each $u$, each of the $d_u$ neighbors of $u$ receives $(d_u-\kappa+1)^{+}/\big(d_u(d_u+1)\big)$ from $v_u$, so $\sum_{i\in N[u]}\varphi_i(v_u)=\min\{1,\kappa/(1+d_u)\}+(d_u-\kappa+1)^{+}/(1+d_u)$, also when $d_u=0$. This sum equals $\kappa/(1+d_u)+(d_u+1-\kappa)/(1+d_u)=1$ if $\kappa\le d_u$ and $1+0=1$ if $\kappa>d_u$. Multiplying by $\mathrm{TS}_u$ and summing over $u$ gives efficiency, which is also a general property of the Shapley value since $v(V_T)=\sum_u\mathrm{TS}_u$. (ii) Substitute $\kappa=1$ in Eq.~\eqref{eq:shapleyclosed}. (iii) If $\kappa>\max_u d_u$, no user can have $\kappa$ neighbors in a coalition, so $v_u(C)=\1\{u\in C\}$, $v(C)=\sum_{u\in C}\mathrm{TS}_u$ is additive, and Eq.~\eqref{eq:shapleyclosed} reduces to $\varphi_i=\mathrm{TS}_i$ because $\kappa\ge1+d_i$ and $(d_u-\kappa+1)^{+}=0$. An additive game satisfies $v(C\cup D)=v(C)+v(D)$ for disjoint $C$ and $D$, so no coalition creates synergy.
\end{proof}

\begin{proof}[Proof of Proposition~\ref{prop:sol4}]
By Eq.~\eqref{eq:sep}, $M_A(s_A,s_B)-M_A(s_A',s_B)=a_A(s_A)-a_A(s_A')$ does not depend on $s_B$, so comparisons between strategies of $A$ hold against every strategy of $B$. (i) $a_A(\mathsf{o})-a_A(\mathsf{a})=\pi_{\mathsf{o}}(1-p_B)+\pi_{\mathsf{a}}p_B>0$, so $\mathsf{o}$ strictly dominates $\mathsf{a}$. (ii) $a_A(\mathsf{g})-a_A(\mathsf{o})=1/d_{ij}-\pi_{\mathsf{g}}-\pi_{\mathsf{o}}(1-p_B)$, whose sign decides whether $\mathsf{g}$ or $\mathsf{o}$ is strictly dominant. The same computation for $B$, whose payoff is obtained by exchanging the roles of the players, involves $p_A$. (iii) When the inequalities are strict, each player has a strictly dominant strategy. A strictly dominated strategy receives zero probability in every Nash equilibrium, so the profile of dominant strategies is the unique equilibrium, in pure or mixed strategies. Both players choose $\mathsf{g}$ if and only if $\pi_{\mathsf{o}}(1-p_B)<1/d_{ij}-\pi_{\mathsf{g}}$ and $\pi_{\mathsf{o}}(1-p_A)<1/d_{ij}-\pi_{\mathsf{g}}$, which is equivalent to Eq.~\eqref{eq:agree} because $\max\{1-p_A,1-p_B\}=1-\min\{p_A,p_B\}$.
\end{proof}

\begin{proof}[Proof of Proposition~\ref{prop:admission}]
The pairs of $C\cup\{k\}$ are the pairs of $C$ and the pairs $\{k,j\}$ with $j\in C$. By Eq.~\eqref{eq:Phi}, $\Phi(C)=\delta s^{-c}S(C)$ and $\Phi(C\cup\{k\})=\delta(s+1)^{-c}\big(S(C)+S(k,C)\big)$. Hence $\Phi(C\cup\{k\})>\Phi(C)$ if and only if $S(C)+S(k,C)>\big((s+1)/s\big)^{c}S(C)$, i.e., $S(k,C)>\big[(1+1/s)^{c}-1\big]S(C)$. For $c\in(0,1]$ and $x>0$, Bernoulli's inequality gives $(1+x)^{c}\le1+cx$, while $e^{y}\ge1+y$ and $\ln(1+x)\ge x/(1+x)$ give $(1+x)^{c}=e^{c\ln(1+x)}\ge1+c\ln(1+x)\ge1+cx/(1+x)$. With $x=1/s$ these bounds read $c/(s+1)\le(1+1/s)^{c}-1\le c/s$, which yields the two sufficient conditions.
\end{proof}

\begin{proof}[Proof of Theorem~\ref{thm:coalition}]
(i) For $C\subseteq C'\subseteq V_T$ and $k\notin C'$, $S(C\cup\{k\})-S(C)=\sum_{j\in C}\Omega_{kj}^{c}\le\sum_{j\in C'}\Omega_{kj}^{c}=S(C'\cup\{k\})-S(C')$ because $\Omega_{kj}\ge0$, so $S$ is supermodular. The map $s\mapsto s^{c}$ is concave on $\{0,1,\dots,n_T\}$ for $c\in(0,1]$, and a concave function of the cardinality is submodular \citep{fujishige2005submodular}. Hence $-q|C|^{c}$ is supermodular for $q\ge0$, and so is $F_q$.

(ii) For nonempty $C^{(\ell)}$, $q^{(\ell)}=S(C^{(\ell)})/|C^{(\ell)}|^{c}$ gives $F_{q^{(\ell)}}(C^{(\ell)})=0$, so $\max_C F_{q^{(\ell)}}(C)\ge0$. If the maximum equals $0$, then $S(C)\le q^{(\ell)}|C|^{c}$ for every nonempty $C$, i.e., $C^{(\ell)}$ attains $q^{\ast}$. Otherwise the maximizer $C^{(\ell+1)}$ satisfies $F_{q^{(\ell)}}(C^{(\ell+1)})>0=F_{q^{(\ell)}}(\emptyset)$, so $C^{(\ell+1)}\neq\emptyset$ and $q^{(\ell+1)}>q^{(\ell)}$. Suppose that the iteration has not stopped at steps $\ell$ and $\ell+1$. Optimality of $C^{(\ell+1)}$ for $q^{(\ell)}$ and of $C^{(\ell+2)}$ for $q^{(\ell+1)}$ gives
\begin{align*}
S(C^{(\ell+1)})-q^{(\ell)}|C^{(\ell+1)}|^{c}&\ge S(C^{(\ell+2)})-q^{(\ell)}|C^{(\ell+2)}|^{c},\\
S(C^{(\ell+2)})-q^{(\ell+1)}|C^{(\ell+2)}|^{c}&\ge S(C^{(\ell+1)})-q^{(\ell+1)}|C^{(\ell+1)}|^{c}.
\end{align*}
Adding the two inequalities yields $(q^{(\ell+1)}-q^{(\ell)})\big(|C^{(\ell+1)}|^{c}-|C^{(\ell+2)}|^{c}\big)\ge0$, hence $|C^{(\ell+2)}|\le|C^{(\ell+1)}|$. If equality held, the two inequalities would force $S(C^{(\ell+2)})=S(C^{(\ell+1)})$ and therefore $\max_CF_{q^{(\ell+1)}}(C)=F_{q^{(\ell+1)}}(C^{(\ell+2)})=F_{q^{(\ell+1)}}(C^{(\ell+1)})=0$, i.e., the iteration would stop at step $\ell+1$. Thus $|C^{(\ell)}|$ strictly decreases for $\ell\ge1$ as long as the iteration continues. Since $|C^{(1)}|\le n_T$, there are at most $n_T$ non-terminal steps, and at most $n_T+1$ maximizations in total. Maximizing the supermodular $F_q$ is minimizing the submodular $-F_q$, which takes strongly polynomial time \citep{schrijver2000submodular}. For $c=1$, $\max_C\{S(C)-q|C|\}$ is the parametric subproblem of the densest-subgraph problem with nonnegative edge weights $\Omega_{ij}$, which reduces to a minimum $s$--$t$ cut \citep{goldberg1984density}.

(iii) Let $C'$ and $C''$ be optimal coalitions. Then $F_{q^{\ast}}(C')=F_{q^{\ast}}(C'')=0=\max_CF_{q^{\ast}}(C)$. Supermodularity gives $F_{q^{\ast}}(C'\cup C'')+F_{q^{\ast}}(C'\cap C'')\ge F_{q^{\ast}}(C')+F_{q^{\ast}}(C'')=0$, and since neither term can exceed the maximum $0$, both equal $0$. Thus $C'\cup C''$ is a nonempty maximizer with ratio $q^{\ast}$, i.e., optimal, and $C'\cap C''$ is either optimal or empty. The union $C^{\ast}$ of all optimal coalitions is therefore optimal and contains every optimal coalition. Because $\mathrm{EC}_i\ge0$ by the Perron--Frobenius theorem, $\sum_{i\in C^{\ast}}\mathrm{EC}_i\ge\sum_{i\in C}\mathrm{EC}_i$ for every optimal $C$, with strict inequality for $C\ne C^{\ast}$ when all $\mathrm{EC}_i>0$, which the same theorem guarantees for a connected $G_T$. Finally, $k\in C^{\ast}$ if and only if $\max\{F_{q^{\ast}}(C)\mid k\in C\}=0$, so $C^{\ast}$ can also be computed with $n_T$ further submodular minimizations over sets that contain a given element.
\end{proof}

\begin{proof}[Proof of Theorem~\ref{thm:ewt}]
We recall the filters of \citet{gilles2013ewt}. Let $\beta$ satisfy $\beta(\theta)=0$ for $\theta\le0$, $\beta(\theta)=1$ for $\theta\ge1$ and $\beta(\theta)+\beta(1-\theta)=1$ for $0\le\theta\le1$, for example $\beta(\theta)=\theta^{4}(35-84\theta+70\theta^{2}-20\theta^{3})$ on $[0,1]$, which is the choice used here, and let $\beta_m(\omega)=\beta\big((|\omega|-(1-\gamma)\Delta_m)/(2\gamma\Delta_m)\big)$ rescale the $m$th transition zone to $[0,1]$. Then
\begin{align*}
\hat{\psi}_1(\omega)&=\begin{cases}1, & |\omega|\le(1-\gamma)\Delta_1,\\[2pt]
\cos\!\big[\tfrac{\pi}{2}\,\beta_1(\omega)\big], & (1-\gamma)\Delta_1\le|\omega|\le(1+\gamma)\Delta_1,\\[2pt]
0, & \text{otherwise},\end{cases}\\[4pt]
\hat{\psi}_m(\omega)&=\begin{cases}\sin\!\big[\tfrac{\pi}{2}\,\beta_{m-1}(\omega)\big], & (1-\gamma)\Delta_{m-1}\le|\omega|\le(1+\gamma)\Delta_{m-1},\\[2pt]
1, & (1+\gamma)\Delta_{m-1}\le|\omega|\le(1-\gamma)\Delta_{m},\\[2pt]
\cos\!\big[\tfrac{\pi}{2}\,\beta_{m}(\omega)\big], & (1-\gamma)\Delta_{m}\le|\omega|\le(1+\gamma)\Delta_{m},\\[2pt]
0, & \text{otherwise},\end{cases}
\end{align*}
for $m=2,\dots,M-1$, and $\hat{\psi}_M$ has the same lower transition at $\Delta_{M-1}$ and equals 1 from $(1+\gamma)\Delta_{M-1}$ up to $\pi$. All filters are even in $\omega$, so it suffices to consider $\omega\in[0,\pi]$. Let $I_m=[(1-\gamma)\Delta_m,(1+\gamma)\Delta_m]$ for $m=1,\dots,M-1$. The condition $\gamma<\gamma^{\ast}$ is equivalent to $(1+\gamma)\Delta_m<(1-\gamma)\Delta_{m+1}$ for $m=1,\dots,M-1$, with $\Delta_M=\pi$, so the transition zones are pairwise disjoint and lie inside $(0,\pi)$. Outside $\bigcup_mI_m$, the definitions above give exactly one filter equal to $1$ and all others equal to $0$. Inside $I_m$, only $\hat{\psi}_m$ (upper transition) and $\hat{\psi}_{m+1}$ (lower transition) are nonzero, and they equal $\cos(\frac{\pi}{2}\beta_m(\omega))$ and $\sin(\frac{\pi}{2}\beta_m(\omega))$, whose squares add to $1$. This proves (i). By Parseval's identity for the discrete-time Fourier transform and (i), $\sum_m\|s\ast\psi_m\|_2^{2}=\frac{1}{2\pi}\int_{-\pi}^{\pi}|\hat{s}(\omega)|^{2}\sum_m|\hat{\psi}_m(\omega)|^{2}\,\mathrm{d}\omega=\|s\|_2^{2}$, which is (ii). For (iii), $\sum_m\hat{s}_m=\hat{s}\sum_m|\hat{\psi}_m|^{2}=\hat{s}$, and the inverse transform is unique. For a window of $K$ samples, the discrete Fourier transform samples the same filters on the grid $\omega_k=2\pi k/K$, the partition of unity holds at every grid point, and the same arguments apply with sums in place of integrals.
\end{proof}

\begin{proof}[Proof of Lemma~\ref{lem:noncausal}]
Since $\hat{g}_m=|\hat{\psi}_m|^{2}$ is real and even,
\[
g_m(\tau)=\frac{1}{2\pi}\int_{-\pi}^{\pi}\hat{g}_m(\omega)\cos(\omega\tau)\,\mathrm{d}\omega
\]
is real and $g_m(-\tau)=g_m(\tau)$. Suppose that $g_m(\tau)=0$ for all $\tau>0$. By evenness $g_m(\tau)=0$ for all $\tau\ne0$, so $g_m$ is $g_m(0)$ times the unit impulse and $\hat{g}_m\equiv g_m(0)$ is constant on $[-\pi,\pi]$. For $M\ge2$ and $\gamma<\gamma^{\ast}$, however, every filter equals $1$ on its plateau and $0$ outside its support, two sets of positive measure, a contradiction. Hence $g_m(\tau)\ne0$ for some $\tau>0$, and the term $g_m(-\tau)s(t+\tau)$ of the convolution makes $s_m(t)$ depend on the future value $s(t+\tau)$.
\end{proof}

\begin{proof}[Proof of Proposition~\ref{prop:causal}]
Under (e), the network weights, $\bm{z}^{\ast}$, $C^{\ast}$, the opinion-leader weights of Eq.~\eqref{eq:olweight} and the scaling constants are fixed functions of training data, which precede every test origin and therefore belong to $\mathcal{I}_t$. Under (b) and (c), the interpolated series and the EWT modes at the origin $t$ are deterministic, hence measurable, functions of observations dated $t$ or earlier, and by (d) the input window collects values dated $t-L+1,\dots,t$. The forecast $\hat{y}_{t+h\mid t}$ is a composition of these measurable maps with the network map of Eqs.~\eqref{eq:bilstm}--\eqref{eq:output}, which is continuous because its LSTM cells compose affine maps with the logistic and hyperbolic tangent functions \citep{hochreiter1997lstm}, and is therefore $\mathcal{I}_t$-measurable. Since $h\ge1$, $y_{t+h}$ is not among the arguments of any of these maps.
\end{proof}

\begin{proof}[Proof of Proposition~\ref{prop:sol12}]
In both solutions $M_A(s_A,s_B)=g(s_A)-g(s_B)$ with $g(\mathsf{o})=\pi_{\mathsf{o}}>0$ and $g(\mathsf{a})=-\pi_{\mathsf{a}}<0$, plus $g(\mathsf{g})=-\pi_{\mathsf{g}}<0$ in Solution~2. Hence $M_A(\mathsf{o},s_B)-M_A(s,s_B)=g(\mathsf{o})-g(s)>0$ for every $s\ne\mathsf{o}$ and every $s_B$, and the same holds for $B$ because $M_B(s_A,s_B)=g(s_B)-g(s_A)$. Strictly dominant strategies form the unique Nash equilibrium, in pure or mixed strategies, with payoffs $g(\mathsf{o})-g(\mathsf{o})=0$. The games are zero-sum by construction and symmetric because $M_B(s_A,s_B)=M_A(s_B,s_A)$, so their value is $0$.
\end{proof}

\begin{proof}[Proof of Proposition~\ref{prop:sol3}]
From Eq.~\eqref{eq:C3sol3}, $M_A(s_A,s_B)=g(s_A)-g(s_B)+\frac{1}{d}\big(\1\{s_A=\mathsf{g}\}+\1\{s_B=\mathsf{g}\}\big)$ with $g$ as in Solution~2, and $M_B$ is obtained by exchanging $s_A$ and $s_B$. Thus $M_A(\mathsf{g},s_B)-M_A(\mathsf{o},s_B)=1/d-\pi_{\mathsf{g}}-\pi_{\mathsf{o}}$ and $M_A(\mathsf{o},s_B)-M_A(\mathsf{a},s_B)=\pi_{\mathsf{o}}+\pi_{\mathsf{a}}>0$ for every $s_B$, and the same differences hold for $B$. The sign of $1/d-(\pi_{\mathsf{o}}+\pi_{\mathsf{g}})$ therefore determines which strategy is strictly dominant, and the dominant profile is the unique equilibrium. Its payoffs are $2/d$ each at $(\mathsf{g},\mathsf{g})$ and $0$ each at $(\mathsf{o},\mathsf{o})$. When $1/d<\pi_{\mathsf{o}}+\pi_{\mathsf{g}}$, the unique equilibrium $(\mathsf{o},\mathsf{o})$ is Pareto-dominated by $(\mathsf{g},\mathsf{g})$, which is the defining feature of a prisoner's dilemma. If $1/d=\pi_{\mathsf{o}}+\pi_{\mathsf{g}}$, the difference $M_A(\mathsf{g},s_B)-M_A(\mathsf{o},s_B)$ vanishes for every $s_B$, and the same holds for $B$, while $\mathsf{a}$ remains strictly dominated. By linearity of expected payoffs, every mixture over $\mathsf{o}$ and $\mathsf{g}$ is then a best response to every strategy of the opponent, so every profile of such mixtures is a Nash equilibrium.
\end{proof}

\begin{proof}[Proof of Proposition~\ref{prop:dyn}]
Subtracting Eq.~\eqref{eq:C6a} from Eq.~\eqref{eq:C6b} gives $\Delta E(r+1)=\Delta E(r)-\eta\Delta E(r)-\mu\Delta E(r)=(1-\mu-\eta)\Delta E(r)$, hence $\Delta E(r)=(1-\mu-\eta)^{r}\Delta E(0)$ with $0<1-\mu-\eta<1$, so the gap decreases geometrically without changing sign. Moreover, $\eta E_A(r+1)+\mu E_B(r+1)=\eta E_A(r)+\eta\mu\Delta E(r)+\mu E_B(r)-\mu\eta\Delta E(r)=\eta E_A(r)+\mu E_B(r)$. Writing $E_A(r)=E^{\ast}-\frac{\mu}{\mu+\eta}\Delta E(r)$ and $E_B(r)=E^{\ast}+\frac{\eta}{\mu+\eta}\Delta E(r)$, which is consistent with the invariant, shows that both opinions converge to $E^{\ast}$. Because $\Delta E(r)$ keeps the sign of $\Delta E(0)$ and $|\Delta E(r)|$ decreases, $E_A(r)$ moves monotonically from $E_A(0)$ towards $E^{\ast}$ and $E_B(r)$ moves monotonically from $E_B(0)$ towards $E^{\ast}$, from the other side. Since $E^{\ast}$ is a weighted mean of $E_A(0)$ and $E_B(0)$ with positive weights, both opinions remain between $E_A(0)$ and $E_B(0)$. Finally, $|\Delta E(r)|\le\varepsilon_E$ if and only if $r\ln(1-\mu-\eta)\le\ln(\varepsilon_E/|\Delta E(0)|)$, i.e., $r\ge\ln(\varepsilon_E/|\Delta E(0)|)/\ln(1-\mu-\eta)$ because $\ln(1-\mu-\eta)<0$. The smallest such integer is $r_E$.
\end{proof}

\begin{proof}[Proof of Proposition~\ref{prop:stab}]
(i) For $\{i,j\}\in E_T$ let $\Lambda_{ij}=1/d_{ij}-\pi_{\mathsf{g}}-\pi_{\mathsf{o}}\big(1-\min\{p_{ij},p_{ji}\}\big)$, so that by Proposition~\ref{prop:sol4} the link belongs to $G^{\mathrm{ag}}$ if and only if $\Lambda_{ij}>0$, and let $\hat{\Lambda}_{ij}$ be the same expression with the estimates. Because $|\min\{a,b\}-\min\{a',b'\}|\le\max\{|a-a'|,|b-b'|\}$, on the event of Proposition~\ref{prop:hoeffding} we have $|\hat{\Lambda}_{ij}-\Lambda_{ij}|\le\pi_{\mathsf{o}}\max\{|\hat{p}_{ij}-p_{ij}|,|\hat{p}_{ji}-p_{ji}|\}<\pi_{\mathsf{o}}\varepsilon<|\Lambda_{ij}|$. Hence $\hat{\Lambda}_{ij}$ and $\Lambda_{ij}$ have the same sign for every linked pair, and unlinked pairs never enter either graph. (ii) For $a,b\ge0$ and $c\in(0,1]$, $|a^{c}-b^{c}|\le|a-b|^{c}$, because $x\mapsto x^{c}$ is concave with value $0$ at $0$ and therefore subadditive, which gives $\max\{a,b\}^{c}\le\min\{a,b\}^{c}+|a-b|^{c}$. By Eq.~\eqref{eq:Phi}, $\Phi(C)=\delta|C|^{-c}\sum_{i<j\in C}\Omega_{ij}^{c}$, so $|\tilde{\Phi}(C)-\Phi(C)|\le\delta|C|^{-c}\,e(C)\,\varepsilon_\Omega^{c}$, where $e(C)$ is the number of edges of the agreement graph with both ends in $C$. Every user has at most $\Delta^{\mathrm{ag}}$ such edges, so $2e(C)\le\Delta^{\mathrm{ag}}|C|$, and the bound follows. (iii) Since $|C|\le n_T$ and $1-c\ge0$, part (ii) gives $|\tilde{\Phi}(C)-\Phi(C)|\le\varepsilon_\Phi$ for every nonempty $C$. For a maximizer $\tilde{C}$ of $\tilde{\Phi}$, $\Phi(\tilde{C})\ge\tilde{\Phi}(\tilde{C})-\varepsilon_\Phi\ge\tilde{\Phi}(C^{\ast})-\varepsilon_\Phi\ge\Phi(C^{\ast})-2\varepsilon_\Phi$. Under the margin condition, $\tilde{\Phi}(C)\le\Phi(C)+\varepsilon_\Phi<\Phi(C^{\ast})-\varepsilon_\Phi\le\tilde{\Phi}(C^{\ast})$ for every nonempty $C\ne C^{\ast}$, so $C^{\ast}$ is the unique maximizer of $\tilde{\Phi}$.
\end{proof}

\FloatBarrier
\section{Notation}\label{app:notation}

Tables~\ref{tab:notation} and \ref{tab:notation2} collect the main symbols. Throughout, $i$, $j$, $k$ and $u$ index users, $t$ indexes months, $\tau$ indexes steps within the input window and lags in Lemma~\ref{lem:noncausal}, $m$ indexes EWT modes and $\ell$ indexes iterations or network layers. A symbol with one user index describes a user and one with two user indices describes a pair, so that $d_i$ is a degree and $d_{ij}$ a distance.

\begin{table}[!htbp]
\centering
\caption{Main notation for forecasting, networks, trust and influence.}\label{tab:notation}
\footnotesize
\begin{tabularx}{\linewidth}{@{}>{\raggedright\arraybackslash}p{0.37\linewidth}>{\raggedright\arraybackslash}X@{}}
\toprule
Symbol & Meaning\\
\midrule
\multicolumn{2}{@{}l}{\textit{Forecasting and evaluation}}\\
$y_t$, $\hat{y}_{t+h\mid t}$ & arrivals in month $t$ and their $h$-step-ahead forecast made at origin $t$\\
$h$, $\mathcal{T}_h$ & forecast horizon (lead time) and set of evaluated test months\\
$x_{\min},x_{\max}$, $x_{\mathrm{low}},x_{\mathrm{high}}$ & training range of a series and target range of the scaling\\
$\mathrm{EM}_1$, $\mathrm{EM}_0$ & evaluation metric with and without the component under study\\
$\mathcal{I}_t$ & information available at origin $t$\\
\addlinespace
\multicolumn{2}{@{}l}{\textit{Network, trust and influence}}\\
$G=(V,E)$, $n$ & interaction network of a platform and number of users\\
$N(i)$, $N^{-}(i)$, $d_i$, $d_i^{+}$, $\mathrm{CL}_i$ & neighbors, in-neighbors, degree (in $G_T$ from Eq.~\eqref{eq:game} on), out-degree and local clustering coefficient of user $i$\\
$T_{ij}$, $\tau_{ij}$, $\tau^{\,k}_{ij}$, $r_i$ & trust degree, direct and third-party trust levels, and reputation\\
$\mathrm{DT}_i$, $\mathrm{IDT}_i$, $\mathrm{RT}_i$, $\mathrm{TS}_i$ & direct, indirect and recommendation trust, and trust score\\
$\alpha$, $b_i$, $\chi_{ji}$ & weight of direct evidence, net recommendation feedback and recommendation polarity\\
$\theta_{\mathrm{D}},\theta_{\mathrm{I}},\theta_{\mathrm{R}}$, $T_{\max}$, $\varepsilon_{\mathrm{tol}}$ & trust thresholds, and iteration limit and tolerance for recommendation trust\\
$V_T$, $n_T$, $G_T=(V_T,E_T)$, $N_T(i)$ & trusted users, their number, trust graph and neighborhoods in it\\
$p_{ij}$, $\hat{p}_{ij}$, $m_i$, $m_{ij}$ & influence probability of $i$ on $j$, its estimate, number of endorsements by $i$ and number of them that $j$ followed\\
$\varepsilon$, $\nu$, $m_{\min}$ & accuracy, failure probability and minimum number of endorsements in Proposition~\ref{prop:hoeffding}\\
\bottomrule
\end{tabularx}
\end{table}

\begin{table}[!htbp]
\centering
\caption{Main notation for games, coalitions, weighting, decomposition and learning.}\label{tab:notation2}
\footnotesize
\begin{tabularx}{\linewidth}{@{}>{\raggedright\arraybackslash}p{0.37\linewidth}>{\raggedright\arraybackslash}X@{}}
\toprule
Symbol & Meaning\\
\midrule
\multicolumn{2}{@{}l}{\textit{Games, coalitions and weighting}}\\
$v$, $\kappa$, $\varphi_i$ & characteristic function, coverage threshold and Shapley value\\
$\mathsf{o},\mathsf{a},\mathsf{g}$, $M_A$, $M_B$ & strategies (original plan, altered plan, agreement) and payoff matrices\\
$\pi_{\mathsf{o}},\pi_{\mathsf{a}},\pi_{\mathsf{g}}$, $p_A$, $p_B$ & payoff parameters and influence probabilities of the two players\\
$d_{ij}$, $\zeta_i$, $\lambda$, $\rho$, $\epsilon_d$ & symmetric distance between users, composite centrality and distance parameters\\
$E^{\mathrm{ag}}$, $G^{\mathrm{ag}}$ & agreement edges and agreement graph\\
$\mathrm{EC}_i$, $\Omega_{ij}$, $S(C)$, $\Phi(C)$ & eigenvector centrality, pairwise synergy, synergy mass and coalition synergy\\
$\vartheta$, $\delta$, $c$ & coordination rates and size exponent of the synergy\\
$q^{\ast}$, $C^{\ast}$ & maximal synergy ratio and coalition of opinion leaders\\
$w_u$, $\omega_{\mathrm{OL}}$, $\xi$ & content weight of user $u$, leader bonus and decay factor\\
$V^{w}_{t}$, $\bar{\chi}^{w}_{t}$, $\chi_e$ & weighted volume, weighted valence and polarity of item $e$\\
$\mu$, $\eta$, $E_A(r)$, $E_B(r)$ & adjustment rates and opinions in the persuasion dynamics\\
$E^{\ast}$, $\varepsilon_E$, $r_E$ & consensus opinion, gap tolerance and number of rounds to reach it\\
$\Delta^{\mathrm{ag}}$, $\varepsilon_\Omega$, $\tilde{\Omega}_{ij}$, $\tilde{\Phi}$, $\varepsilon_\Phi$ & maximum degree of the agreement graph, bound on synergy errors, estimated synergies, coalition synergy computed from them and bound on its error\\
\addlinespace
\multicolumn{2}{@{}l}{\textit{Decomposition and learning}}\\
$s(t)$, $s_m(t)$, $M$ & series, its EWT modes and number of modes\\
$\Delta_m$, $\gamma$, $\hat{\psi}_m$, $\beta$ & spectral boundaries, transition ratio, filters and transition function\\
$L$, $\bm{x}_\tau$, $N_x$ & input window length, input vector at step $\tau$ and its dimension\\
$\bm{h}_\tau$, $\bm{c}_\tau$, $\bar{\bm{h}}_t$ & hidden state and memory cell of an LSTM, and read-out state of the stack\\
$N_u$, $N_l$, $N_{lr}$, $N_{L2}$, $N_d$, $N_{lm}$ & hyperparameters of Table~\ref{tab:hyper}\\
$\bm{z}$, $\mathcal{H}$ & hyperparameter vector and search space\\
$\mathcal{L}(\bm{z})$, $\hat{\mathcal{L}}(\bm{z})$, $\varsigma(\bm{z})$ & validation loss and its posterior mean and standard deviation\\
$a_{\mathrm{EI}}(\bm{z})$ & acquisition function of BO (expected improvement)\\
\bottomrule
\end{tabularx}
\end{table}

\FloatBarrier

\bibliographystyle{elsarticle-harv}
\bibliography{references}
\end{document}